\documentclass[10pt,twocolumn,letterpaper]{article}
\newif\ifarxiv % defines the version switch below -- keep this line as it is

\arxivtrue

\ifarxiv
  \usepackage[pagenumbers]{cvpr} % To force page numbers, e.g. for an arXiv version
\else
  \usepackage[review]{cvpr}      % To produce the REVIEW version
\fi

\usepackage{tabularx}
\makeatletter\TX@delta=0.1pt\makeatother
\usepackage{multirow}
\usepackage{makecell}
\usepackage[most]{tcolorbox}
\usepackage{colortbl}
\usepackage{array}

\definecolor{dualsaebg}{RGB}{235,242,255}
\definecolor{myyellow}{RGB}{230, 230, 50}
\definecolor{residual_fig2}{RGB}{255, 178, 110}
\definecolor{recon_fig2}{RGB}{121, 173, 210}

\definecolor{x_fig4}{RGB}{2, 114, 178}
\definecolor{xhat_fig4}{RGB}{230, 159, 0}

\definecolor{dualrow}{RGB}{242,246,255}
\newcommand{\dc}[1]{\cellcolor{dualrow}#1}

\newcommand{\stdval}[2]{#1\textsubscript{$\pm$#2}}
\newcommand{\stdbf}[2]{\textbf{#1}\textsubscript{$\pm$#2}}

\newcommand{\dualhl}[1]{%
  \begingroup
  \setlength{\fboxsep}{1pt}%
  \colorbox{orange!22}{\texttt{#1}}%
  \endgroup
}

\newcommand{\basehl}[1]{%
  \begingroup
  \setlength{\fboxsep}{1pt}%
  \colorbox{orange!25}{\texttt{#1}}%
  \endgroup
}

\newcommand{\missbl}[1]{%
  \begingroup
  \setlength{\fboxsep}{1pt}%
  \colorbox{gray!25}{\texttt{#1}}%
  \endgroup
}

\newcommand{\actscore}[1]{%
  \textcolor{gray!75}{\texttt{\textit{#1}}}%
}

\newcommand{\dualhlcap}[1]{%
  \begingroup
  \setlength{\fboxsep}{1pt}%
  \colorbox{orange!22}{#1}%
  \endgroup
}

\newcommand{\missblcap}[1]{%
  \begingroup
  \setlength{\fboxsep}{1pt}%
  \colorbox{gray!25}{#1}%
  \endgroup
}

\newcommand{\steerhl}[1]{%
    \colorbox{orange!20}{\texttt{#1}}%
}

\usepackage{placeins}
\makeatletter
\let\@fb@origFloatBarrier\FloatBarrier
\renewcommand\FloatBarrier{\@fb@origFloatBarrier
  \ifdim\pagegoal=\maxdimen\ifx\@toplist\@empty\ifx\@botlist\@empty
    \global\@colroom\@colht\global\vsize\@colht\fi\fi\fi}
\makeatother

\makeatletter
\newcommand\supptableofcontents{%
  \addtocontents{toc}{\protect\setcounter{tocdepth}{2}}%
  \begingroup
    \setcounter{tocdepth}{-1}%
    \def\numberline##1##2.~{\hb@xt@\@tempdima{##1.\hfil}}%
    \def\@tocrmarg{2.55em plus 1fil}%
    \hyphenpenalty\@M \exhyphenpenalty\@M
    \renewcommand*\l@section[2]{%
      \ifnum\c@tocdepth>\z@
        \addpenalty\@secpenalty \addvspace{0.4em \@plus\p@}%
        \setlength\@tempdima{1.5em}%
        \begingroup
          \parindent\z@ \rightskip\@pnumwidth \@plus 1fil \parfillskip -\@pnumwidth
          \leavevmode\bfseries \advance\leftskip\@tempdima \hskip -\leftskip
          ##1\nobreak\hfill \nobreak\hb@xt@\@pnumwidth{\hss ##2\kern-\p@\kern\p@}\par
        \endgroup
      \fi}%
    \tableofcontents
  \endgroup
  \setcounter{tocdepth}{2}%
}
\makeatother

\definecolor{cvprblue}{rgb}{0.21,0.49,0.74}
\usepackage[pagebackref,breaklinks,colorlinks,allcolors=cvprblue]{hyperref}

\def\confName{CVPR}
\def\confYear{2027}

\title{Separating Interpretability and Reconstruction in Sparse Autoencoders}

\ifarxiv
\author{%
Anh Tien Nguyen$^{1}$ \qquad Dimitris Samaras$^{1,\dagger}$ \qquad Mahdi S. Hosseini$^{2,3,\dagger}$\\
$^{1}$Stony Brook University \qquad $^{2}$Concordia University \qquad $^{3}$Mila--Quebec AI Institute\\
{\tt\small \{atinguyen,samaras\}@cs.stonybrook.edu \qquad mahdi.hosseini@concordia.ca}
}
\else
\author{Anonymous}
\fi

\usepackage{algorithm}
\usepackage{algpseudocode}

\begin{document}
\maketitle
\ifarxiv
  % unnumbered footnote for the dagger in the author list (\fnsymbol{2} is the dagger)
  {\renewcommand{\thefootnote}{\fnsymbol{footnote}}\footnotetext[2]{Equal senior authorship.}}
  % unmarked footnote line with the code link
  {\renewcommand{\thefootnote}{}\footnotetext{Code: \href{https://github.com/AtlasAnalyticsLab/DualSAE}{\nolinkurl{github.com/AtlasAnalyticsLab/DualSAE}}}}
\fi
\begin{abstract}
Sparse Autoencoders (SAEs) are effective tools for mechanistic interpretability by decomposing neural representations into sparse and interpretable features. However, a single sparse representation in traditional SAEs is responsible for both reconstruction fidelity and interpretability, which are not guaranteed to align. To separate these objectives, we propose \textbf{DualSAE}, a modular framework that augments standard SAEs with a plug-in Residual Branch. A standard SAE serves as the \textit{Sparse Branch}, learning a sparse code that contains the dominant semantics of the input. The added \textit{Residual Branch} models the non-semantic information omitted by the Sparse Branch. 
We ensure that the contributions of the two branches towards the overall reconstruction have minimal overlap through a decorrelation loss. We also adopt a two-stage training strategy that first stabilizes sparse representation learning before coupling with the Residual Branch. Across diverse encoder--SAE configurations on two vision datasets and two language corpora, DualSAE achieves mean relative improvements of approximately 29\% in reconstruction fidelity and 34\% in interpretability (measured by monosemanticity score). Moreover, with more interpretable representations, DualSAE yields consistent performance gains in the downstream applications of topic modeling, hypothesis generation, and LLM steering. Qualitative analysis further suggests that the Sparse Branch indeed preserves dominant semantic content, whereas the Residual Branch captures non-semantic information to improve reconstruction fidelity. 
\end{abstract}

%To decorrelate the two branches, we introduce a {decorrelation loss} that discourages the Residual Branch from producing corrections aligned with the Sparse Branch.

% Across various encoder-SAE settings on two vision datasets and two corpora, DualSAE improves reconstruction fidelity by about 30\%, interpretability (through the monosemanticity score) by about 31\%, more than doubles class purity.    
\section{Introduction}
\label{sec:intro}

\begin{figure}[t]
    \centering
    % \vspace{-1em}
    \includegraphics[trim={0 0 1.9cm 0},clip,width=\linewidth]{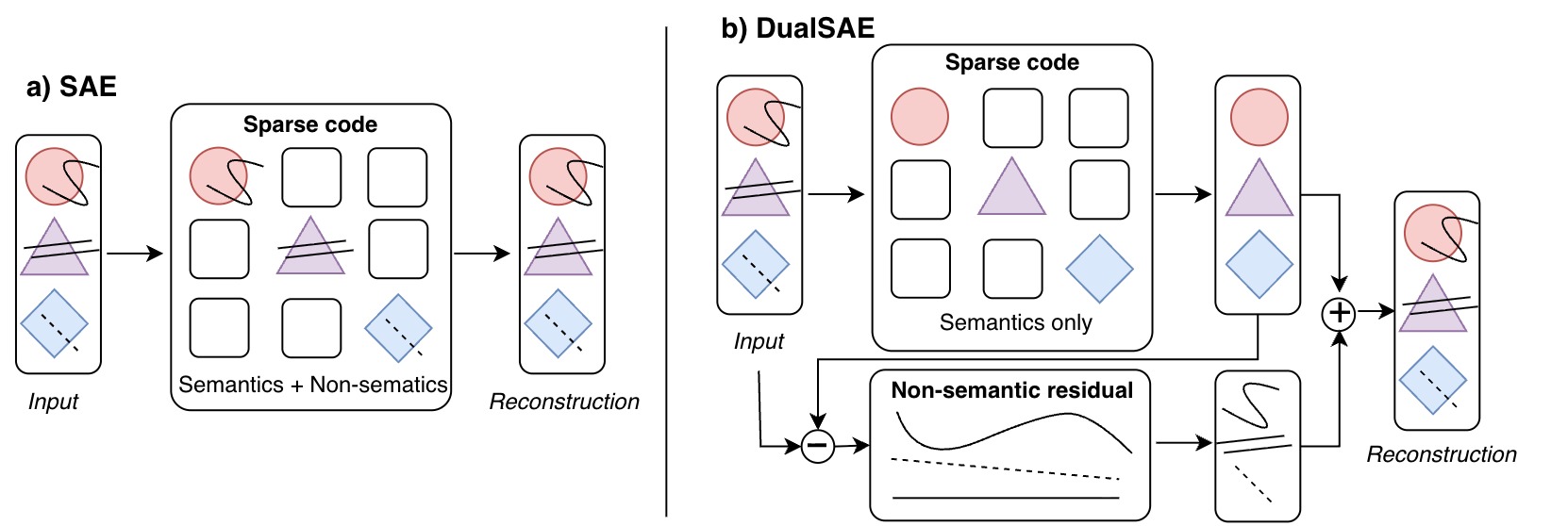}
    \caption{\textbf{Standard SAE and DualSAE comparison}. (a) A standard SAE encodes both dominant semantic concepts and non-semantic information into a single sparse code, entangling interpretability with reconstruction. (b) Our DualSAE disentangles these objectives by adding a non-semantic Residual Branch, which separates interpretability from the overall reconstruction objective.}
    \vspace{-1em}
    \label{fig:compare}
\end{figure}

Sparse autoencoders (SAEs) decompose high-dimensional neural activations into sparse, interpretable latent features while preserving enough information for reconstruction \citep{bricken2023monosemanticity, cunningham2023, scalingsae, templeton2024scaling}. Ideally, each latent captures a coherent and interpretable concept \citep{olshausen1996emergence,lee2006efficient,mairal2009online,topksae}. However, in standard SAEs (Figure \ref{fig:compare}a), the same sparse code is responsible for both interpretability and reconstruction \citep{scalingsae,batchtopksae,jumprelu,cui2026on}. These two objectives are not guaranteed to align: reconstruction favors any useful information, including non-semantic information that may not correspond to interpretable concepts. As a result, using a reconstruction objective to learn interpretable features can force latents to mix multiple factors, creating a trade-off between interpretability and reconstruction fidelity. We argue that this tension stems from requiring one sparse representation to fulfill two distinct roles.

To address the above problem, we propose \textbf{DualSAE}, a modular framework that separates semantic information from non-semantic residual information through a dual-branch architecture (Figure~\ref{fig:compare}b). Architecturally, DualSAE extends a standard SAE by adding a Residual Branch. The \emph{Sparse Branch}, modeled as a standard SAE, encodes the input into a sparse code and is responsible for capturing the dominant semantic information. {In contrast, the \emph{Residual Branch} is {not designed for interpretability}. Instead, its role is to model the remaining non-semantic information omitted by the Sparse Branch that is still useful for accurate reconstruction.} This decomposition separates the representational responsibilities of the two branches: the Sparse Branch focuses on learning sparse, semantically coherent features, while the Residual Branch captures non-semantic residual information required for faithful reconstruction.

Empirically, by plugging the Residual Branch into standard SAEs, DualSAE significantly improves interpretability by 34\% and reconstruction fidelity by 29\% on various encoder–SAE configurations as shown by experiments on two vision datasets (ImageNet-1K \citep{imagenet15russakovsky} and OpenImagesV7 \citep{OpenImages}) and two language corpora (OpenWebText \citep{Gokaslan2019OpenWeb} and GitHub code \citep{codeparrot_github_code}). The more interpretable representation from DualSAE also improves performance on two vision (topic modeling and hypothesis generation) and one language (LLM steering) downstream applications, with qualitative results supporting enhanced interpretability. In summary, our main contributions are:
    \begin{enumerate}
        \item \textbf{DualSAE}, a modular framework that augments standard SAEs with a plug-in Residual Branch to separate interpretability from the overall reconstruction objective.
        \item A \textbf{decorrelation loss} to ensure that the contributions of the two branches towards the overall reconstruction have minimal overlap. This is done by first stabilizing the Sparse Branch during the warm-up stage and then training the Residual Branch with the decorrelation loss.
        \item \textbf{Quantitative and qualitative evidence}, in both vision and natural language domains, showing that DualSAE significantly improves reconstruction fidelity and interpretability. 
    \end{enumerate}

\section{Related work}
\label{sec:related}

\textbf{Sparse autoencoders.}
    Sparse Autoencoders (SAEs) decompose dense activations into sparse, inspectable features~\citep{bricken2023monosemanticity,scalingsae,jumprelu,nasiri-sarvi2026sparc}, which is motivated by the superposition hypothesis~\citep{elhage2022toy}. Existing SAEs employ different sparsification mechanisms, such as TopK SAE~\citep{scalingsae,topksae}, BatchTopK SAE~\citep{batchtopksae}, ReLU SAE~\citep{reluSAE}, and JumpReLU SAE~\citep{jumprelu}. Despite architectural differences, these methods reconstruct activations from a single sparse code, coupling interpretability with reconstruction.

\noindent \textbf{Structure and regularization in SAEs.} Various methods introduce structure or additional regularization into SAE representations. GatedSAE~\citep{gatedsae} separates feature detection from magnitude estimation, applying the sparsity penalty to the gating pathway to mitigate shrinkage and improve the sparsity–reconstruction trade-off. MatryoshkaSAE~\citep{bussmann2025matryoshka} jointly trains nested dictionaries with reconstruction objectives at multiple sizes, encouraging the learning of both general and specific features. OrtSAE~\citep{korznikov2026ortsae} penalizes high pairwise cosine similarities between decoder vectors to reduce feature absorption and composition. AlignSAE~\citep{yang2026alignsae} uses supervised post-training to bind predefined concepts to dedicated latent slots while preserving free slots for general reconstruction. CB-SAE~\citep{cbsae} adopts a post-hoc approach that prunes low-utility SAE features and introduces a lightweight concept bottleneck aligned with user-defined concepts. More closely related to our work, Deng \etal~\citep{deng2026decompose} combine a low-rank linear bottleneck with a standard SAE to separate dense, high-variance information from sparse features. However, those branches could overlap each other. In DualSAE, we introduce an explicit decorrelation loss that penalizes residual corrections aligned with active sparse decoder directions, directly discouraging information overlap between branches. We further provide a theoretical analysis showing that, without decorrelation, the Residual Branch can minimize reconstruction error without depending on sparse representations. 
% Our analysis demonstrates how the decorrelation loss discourages this shortcut by penalizing residual corrections that overlap with active sparse directions. Combined with two-stage training, DualSAE explicitly promotes complementary branch specialization while decoupling interpretability-oriented representation learning from reconstruction fidelity.
\section{Motivation}

Standard SAEs use a single sparse representation for both interpretability and reconstruction fidelity~\cite{reluSAE,topksae,batchtopksae,jumprelu}. There is often a trade-off between these two objectives. To study this tension, let \(x\) denote an image embedding, \(\hat{x}\) its SAE reconstruction, and \(r=x-\hat{x}\) the residual. We examine whether $\hat{x}$ mostly captures the semantic information, while non-semantic information remains in $r$. \textit{Semantic information} is the part of an embedding that is shared by inputs of the same concept and distinguishes it from other concepts, \ie, it is predictive of the concept label. \textit{Non\-semantic information} is the remainder that is still needed to reconstruct the embedding but is only weakly predictive of the concept label.

\begin{figure}[ht]
    % \vspace{-1em}
    \includegraphics[trim={0cm 0 10.2cm 0},clip,width=\linewidth]{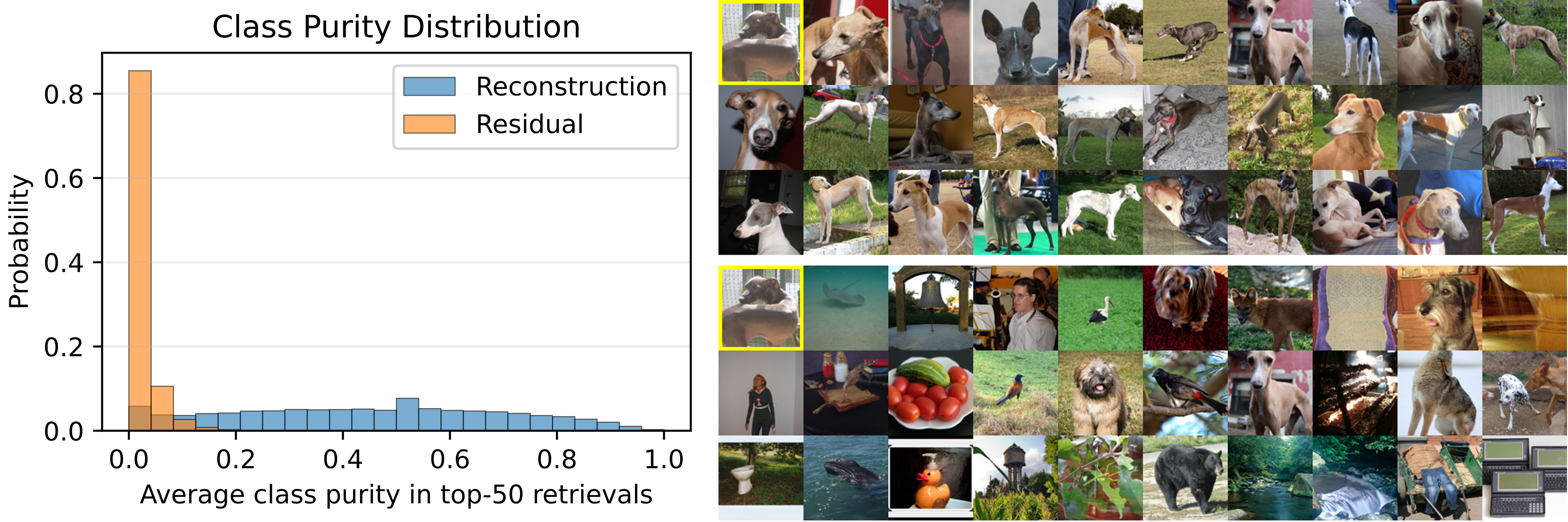}
    % \vspace{-0.5em}
    \caption{\textbf{Left:} Distribution of class purity for nearest neighbors retrieved using \textcolor{recon_fig2}{reconstruction \(\hat{x}\)} and \textcolor{residual_fig2}{residual \(r\)} from BatchTopK SAE. Retrievals from \textcolor{recon_fig2}{\(\hat{x}\)} are substantially purer than those from \textcolor{residual_fig2}{\(r\)}. \textbf{Right:} Example retrievals for the same \textcolor{myyellow}{query image}  using \(\hat{x}\) (top) and \(r\) (bottom). The reconstruction $\hat{x}$ shows consistent semantics, while the residual is non-semantic.}
    \label{fig:motivation_exp}
    % \vspace{-1em}
\end{figure}

To test the above hypothesis, we adopt BatchTopK SAE on CLIP embeddings from ImageNet-1K and evaluate semantic consistency through nearest-neighbor retrieval. For each validation image, we use its \(\hat{x}\) (or \(r\)) as the query and retrieve the 50 most cosine-similar \(\hat{x}\) (or \(r\)) of the other images. We define \textit{class purity} as the fraction of these retrieved images that belong to the query's class. A higher class purity indicates that the representation groups images of the same semantic class more consistently. As shown in Figure~\ref{fig:motivation_exp} (left), the \textcolor{recon_fig2}{reconstructed \(\hat{x}\)} achieves substantially higher class purity than the \textcolor{residual_fig2}{residual \(r\)}, indicating that the sparse reconstruction better preserves class-specific semantic information. Qualitatively, given a \textit{dog} query image (yellow border), \(\hat{x}\) retrieves semantically consistent \textit{dog} samples (top-right), whereas \(r\) retrieves unrelated samples (bottom-right). Together, these observations suggest that the sparse reconstruction captures dominant semantic structure, while the residual retains information less aligned with semantic class identity. This motivates our design to separate these two types of information rather than forcing them into a single sparse representation.

\section{Method}

\subsection{DualSAE}
\textbf{Sparse Branch}. The branch is designed similarly to standard SAEs. 
Given an input feature \(x \in \mathbb{R}^d\), it produces an overcomplete sparse latent code \(z \in \mathbb{R}^m\) as \(z = S(W_E x + b_E)\), and reconstructs the input as \(x^M = W_D z + b_D\), where \(S(\cdot)\) denotes the sparsification rule, \(W_E \in \mathbb{R}^{m \times d}\) is the encoder, and \(W_D \in \mathbb{R}^{d \times m}\) is the decoder (\(m \gg d\)).
As a result, \(x^M\) is the Sparse Branch reconstruction, containing the dominant semantic information. 

\begin{figure}[ht]
    %\vspace{-1em}
    \centering
    \includegraphics[width=\linewidth]{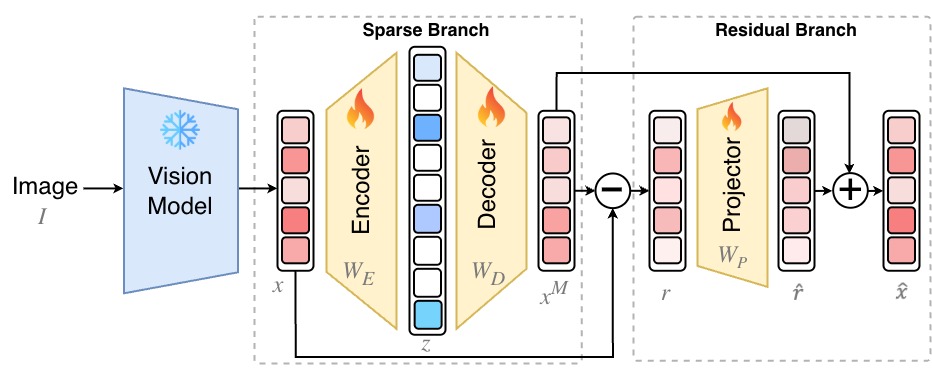}
    %\vspace{-0.3em}
    \caption{\textbf{DualSAE architecture}. The \textit{Sparse Branch}, modeled as a standard SAE, captures semantic information of the input feature $x$ in the sparse latent $z$, then projects it back to the original dimension as $x^M$. The augmented \textit{Residual Branch} captures the non-semantic information $\hat{r}$. The final reconstructed feature combines the outputs of both branches: $\hat{x} = x^M + \hat{r}$.}
    %\vspace{-1em}
    \label{fig:dual_sae}
\end{figure}

\textbf{Residual Branch}. We do not require \(x^M\) to reconstruct all information in \(x\), as \(x\) may contain non-semantic information that should not be learned by sparse interpretable features. 
We therefore introduce a Residual Branch that models the non-semantic information. Specifically, we define the residual as \(r = x - x^M\), and let the Residual Branch predict \(\hat{r} = Q(W_P r)\), where $W_P$ is a projector and $Q(\cdot)$ is a nonlinear operator. 
The final reconstruction combines the two branches as \(\hat{x} = x^M + \hat{r}\).

\textbf{Role of each branch.} The Residual Branch is not designed to be interpretable. Instead, it provides a separate branch for non-semantic information that is useful for reconstruction. In a standard SAE, both dominant semantic and non-semantic information must be modeled by the single sparse code (Figure \ref{fig:compare}a), forcing the Sparse Branch to serve both interpretability and reconstruction objectives. In DualSAE, the non-semantic information of the Residual Branch is separated from semantic information in the Sparse Branch (Figure \ref{fig:compare}b), allowing the sparse representations to have higher interpretability without being directly constrained by the reconstruction objective.
    
\textbf{Reconstruction objective.} We use the standard L2 error to control the reconstruction quality $\mathcal{L}_{\text{rec}} = \|x - \hat x\|_2^2$. As mentioned above, we do not force \(x^M\) alone to match \(x\) perfectly. Instead, we combine the outputs of both branches to obtain the reconstructed feature $\hat x$. 
In other words, the reconstruction constraint now applies to the combined sparse reconstruction and residual, rather than to the sparse reconstruction alone. 
% This is the key design difference from traditional SAEs.

\subsection{Decorrelation loss}
    \textbf{Problem formulation.}
    Let the Sparse Branch produce \(x^M=W_Dz^M(x)\), define the residual as
    \(r=x-x^M\), and let the Residual Branch produce
    \(\hat r = Q(W_P r):=P_{\theta_P}(r)\), where \(P_{\theta_P}\) is an arbitrary nonlinear
    function. The final reconstruction is \(\hat x=x^M+\hat r\). The reconstruction error is
    \begin{equation}
        \hat x-x
        =
        P_{\theta_P}(r)-r,
        \quad
        \mathcal{L}_{\mathrm{rec}}
        =
        \mathbb{E}_x
        \left[
            \left\|P_{\theta_P}(r)-r\right\|_2^2
        \right].
        \label{eq:nonlinear-shortcut}
    \end{equation}
    This exposes a shortcut:
    regardless of the quality of \(x^M\), the Residual Branch can minimize the
    reconstruction loss by learning the \textit{identity map}
    \(\hat{r}= r\) on the residual distribution. Consequently,
    reconstruction alone does not require the Sparse Branch to capture meaningful
    information.
    
    \textbf{Information overlap.}
    Let \(\mathcal{A}(x)=\{j:z_j(x) > 0\}\) denote the active sparse latents and
    let \(D_x=W_D[:,\mathcal{A}(x)]\) contain the active decoder
    directions. These directions define the sample-dependent sparse subspace
    \(\mathcal{S}_M(x)=\operatorname{span}(D_x)\). If the residual \(\hat{r}\) aligns strongly with $\mathcal{S}_M$, the Residual Branch can overwrite information already represented by the active sparse features. To discourage this overlap, we penalize the correlations between $\hat{r}$ and the active decoder directions:
    \begin{equation}
        \mathcal{L}_{\mathrm{dcor}} = \|D_x^\top \hat{r}\|_2^2 .
        \label{eq:span_penalty}
    \end{equation}

    % \textbf{Information overlap.}
    % Let
    % \(\mathcal{A}(x)=\{j:z_j^M(x)\neq0\}\)
    % denote the active sparse latents, and let
    % \[
    %     D_x=[\bar{w}_j]_{j\in\mathcal{A}(x)},
    %     \qquad
    %     \bar{w}_j=\frac{w_j}{\|w_j\|_2},
    % \]
    % contain their normalized decoder directions. These directions span the
    % sample-dependent sparse subspace
    % \(S_M(x)=\operatorname{span}(D_x)\).
    % If the residual correction \(\hat{r}\) aligns strongly with these directions,
    % the Residual Branch can overwrite information already represented by the
    % active sparse features. To discourage this redundancy, we penalize the
    % correlation between \(\hat{r}\) and the active sparse decoder directions:
    % \[
    %     \mathcal{L}_{\mathrm{dcor}}
    %     = \|D_x^\top \hat{r}\|_2^2.
    % \]

    % \textbf{Effect on the nonlinear shortcut.}
    % For a fixed residual, let \(y=P_{\theta_P}(r)\). The output 
    % objective is
    % \(\ell_r(y)=\|y-r\|_2^2+\gamma\|D_x^\top y\|_2^2\).
    % At the identity shortcut \(y=r\), the reconstruction gradient vanishes,
    % whereas $\left.\nabla_y\ell_r(y)\right|_{y=r} = 2\gamma D_xD_x^\top r.$   
    % This gradient is nonzero whenever \(D_x^\top r\neq0\), meaning that the
    % copied residual overlaps with the active mono directions. Thus, the identity
    % map is not stationary unless the residual is already orthogonal to the active mono subspace.

    \textbf{Effect on the identity map.}
    We consider the objective with respect to the Residual Branch output: $\ell_r(\hat{r})
        = \|\hat{r}-r\|_2^2
        + \gamma\|D_x^\top \hat{r}\|_2^2$.
    At the identity solution \(\hat{r}=r\), the reconstruction gradient vanishes,
    whereas $\left.\nabla_{\hat{r}} \ell_r(\hat{r})\right|_{\hat{r}=r}
        = 2\gamma D_xD_x^\top {r}.$
    This gradient is nonzero whenever \(D_x^\top {r}\neq0\), \ie, whenever the  residual contains components aligned with the active sparse decoder
    directions. Thus, the identity map \(\hat{r}=r\) is not stationary in output
    space.
    
    \textbf{Training objective.} With the decorrelation loss, the final training loss of DualSAE is
    \begin{equation}
        \mathcal{L}
        =
        \mathcal{L}_{\mathrm{rec}}
        +
        \lambda\mathcal{L}_{\mathrm{sparse}}
        +
        \gamma\mathcal{L}_{\mathrm{dcor}} ,
    \end{equation}

    % Here, \(\mathcal{L}_{\mathrm{sparse}}\) depends on the mono-branch SAE variant. For vanilla SAEs, sparsity is encouraged through the L1 constraint  $(\mathcal{L}_{\mathrm{sparse}}=\|x\|_1)$. For TopK- or BatchTopK SAEs, sparsity is enforced directly by the latent selection rule, so no additional sparsity loss is required and we set \(\mathcal{L}_{\mathrm{sparse}}=0\). 
    where \(\lambda\) and \(\gamma\) control the strengths of the sparsity and the decorrelation constraints, respectively. Note that for hard-sparsity methods, \eg, TopK and BatchTopK, the sparsity loss $\mathcal{L}_{\mathrm{sparse}}$ is not used.

    % \textbf{Clarification.} Our decorrelation loss is not intended as an contrastive objective. Contrastive methods primarily organize relationships across samples or views. By contrast, our decorrelation loss operates within a single input and across two branches with different functional roles.  Its purpose is therefore not to separate samples or modalities, but to prevent the Residual Branch from redundantly encoding information already represented by the Sparse Branch.

\subsection{Two-stage training}

    \begin{algorithm}[t]
    \caption{Two-stage training}
    \label{alg:dualsae_training}
    \small
    \algtext*{EndFor}
    \algrenewcommand{\algorithmiccomment}[1]{\hfill{\color{gray}\(\triangleright\) #1}}
    \begin{algorithmic}[1]
    \State \textbf{Stage 1: Train Sparse Branch}
    \For{\textit{stage-1 steps}} \Comment{warm-up as a standard SAE}
        \State $z \gets S(W_E x+b_E)$ \Comment{sparse latent code}
        \State $x^M \gets W_Dz+b_D$ \Comment{semantic reconstruction}
        \State $\mathcal{L} \gets
        \mathcal{L}_{\mathrm{rec}}(x,x^M)
        + \lambda\mathcal{L}_{\mathrm{sparse}}$ \Comment{no decorrelation loss}
        \State Update $W_E,W_D$ \Comment{no Residual Branch yet}
    \EndFor
    \State \textbf{Stage 2: Train DualSAE}
    \For{\textit{stage-2 steps}} \Comment{add the Residual Branch}
        \State $z \gets S(W_E x+b_E)$ \Comment{same Sparse Branch as Stage 1}
        \State $x^M \gets W_Dz+b_D$
        \State $\hat r \gets Q(W_P(x-x^M))$ \Comment{predict residual}
        \State $\hat x \gets x^M+\hat r$ \Comment{combine both branches}
        \State $\mathcal{L} \gets
        \mathcal{L}_{\mathrm{rec}}
        + \lambda\mathcal{L}_{\mathrm{sparse}}
        + \gamma\mathcal{L}_{\mathrm{dcor}}$ \Comment{all losses}
        \State Update $W_E,W_D,W_P$ \Comment{train jointly}
        % \vspace{-0.3em}
    \EndFor
    \end{algorithmic}
    \end{algorithm}
    % \vspace{-2em}

    We train DualSAE with a two-stage schedule to stabilize the sparse dictionary
before introducing the Residual Branch (Algorithm~\ref{alg:dualsae_training}).
During the warm-up stage, the Sparse Branch is optimized as a standard SAE,
allowing it to capture the dominant semantic information of the
input without interference from the Residual Branch. After warm-up,
we activate the Residual Branch and the decorrelation loss
\(\mathcal{L}_{\mathrm{dcor}}\), encouraging the Residual Branch to model
non-semantic information while discouraging it from
reusing active sparse decoder directions.

    \begin{figure}[h]
        \vspace{-0.7em}
        \includegraphics[width=\linewidth]{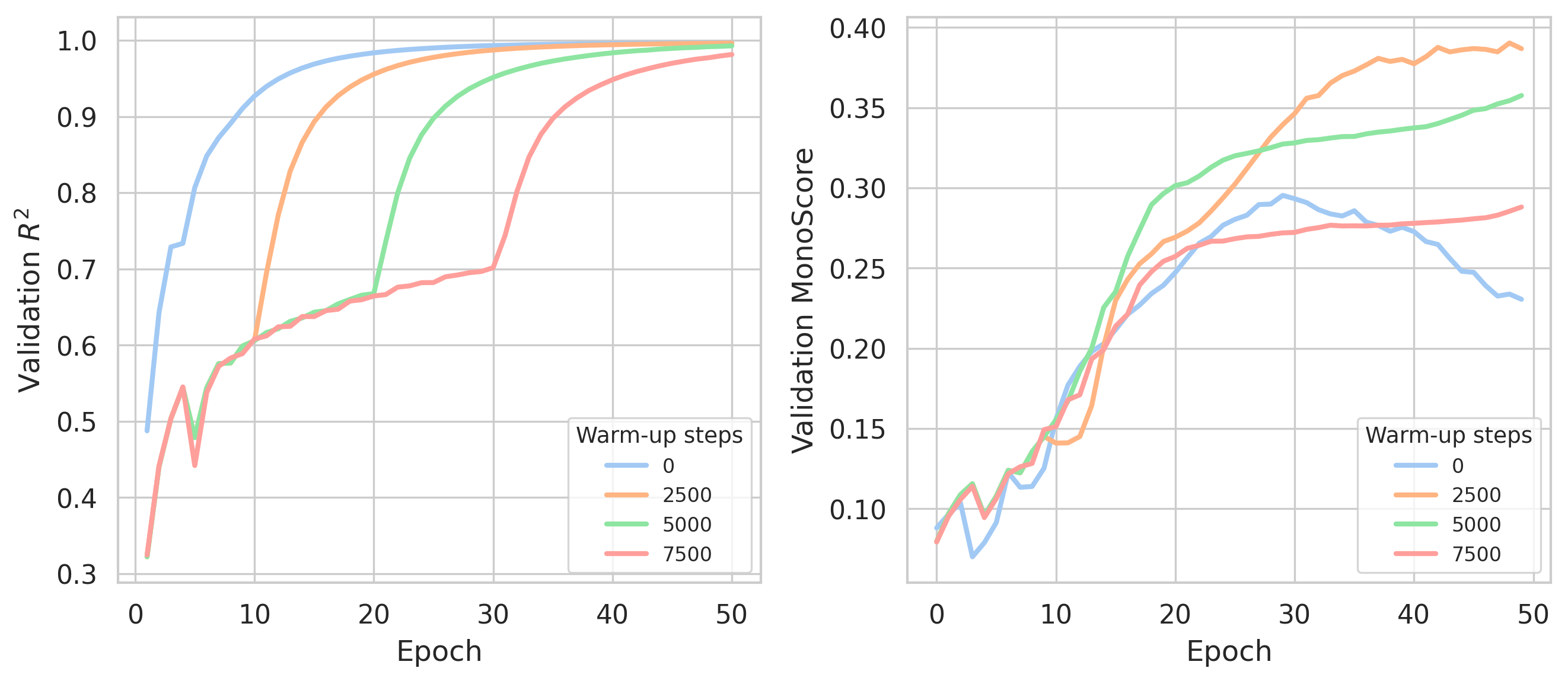}
        \vspace{-2em}
        \caption{\textbf{Effect of warm-up length}, showing reconstruction \(R^2\) (left) and MonoScore (right) during ImageNet-1K training. Shorter warm-up accelerates reconstruction gains; moderate warm-up improves monosemanticity. Excessive warm-up delays reconstruction gains without further improving MonoScore.}
        \label{fig:warmup_ablation}
        \vspace{-0.7em}
    \end{figure}
    
    We also measure the reconstruction quality ($R^2$) and interpretability (MonoScore~\cite{monoscore}) under different warm-up lengths of Dual-BatchTopK on ImageNet-1K, illustrated in Figure~\ref{fig:warmup_ablation}. With little or no warm-up, reconstruction fidelity improves rapidly, but MonoScore remains lower.
    A moderate warm-up provides a better balance: the model reaches better reconstruction while maintaining stronger monosemanticity.
    In contrast, an excessively long warm-up delays reconstruction improvement and does not further improve MonoScore, suggesting that the Residual Branch should be introduced after the Sparse Branch has sufficiently stabilized, while avoiding an unnecessarily long warm-up.
\section{Experimental analysis}
\label{sec:experiment}

    % \textcolor{blue}{This section aims to address five questions: (i) Does DualSAE improve both reconstruction fidelity and interpretability over standard SAEs? (ii) Do these gains come from joint training of the two branches rather than extra parameters or residual capacity? (iii) Do the two branches capture semantic and non-semantic information, respectively? (iv) How robust is DualSAE to the expansion ratio? (v) Do the advantages transfer to language and downstream applications?}

    This section comprehensively evaluates DualSAE across diverse settings. We first compare DualSAE with standard SAEs in reconstruction fidelity and interpretability (Sec.~\ref{sec:exp_main}). We then analyze where these gains come from (Sec.~\ref{exp:why}), branch decomposition (Sec.~\ref{sec:exp_branch}), and expansion robustness (Sec.~\ref{sec:exp_expansion}). Finally, we evaluate DualSAE on language (Sec.~\ref{sec:exp_language}) and downstream tasks (Sec.~\ref{sec:exp_topic}).

    \textbf{Datasets and models.} We adopt features of ImageNet-1K~\citep{imagenet15russakovsky} and OpenImagesV7~\citep{OpenImages} from three encoders: CLIP~\citep{clip}, SigLIP2~\citep{SigLIP2}, and ViT-L/16~\citep{vit}. Besides vision domain, we also evaluate DualSAE in the natural language domain. Specifically, we obtain activations from Pythia~\citep{biderman2023pythia} on OpenWebText~\citep{Gokaslan2019OpenWeb} and GitHub code~\citep{codeparrot_github_code}.
    We evaluate six different SAE variants: TopK~\citep{topksae}, BatchTopK~\citep{batchtopksae}, JumpReLU~\citep{jumprelu}, GatedSAE~\citep{gatedsae}, MatryoshkaSAE~\citep{bussmann2025matryoshka}, and CB-SAE~\citep{cbsae}, together with their DualSAE counterparts.
    All quantitative results are averaged across three random seeds.
    Please refer to Supplementary~\ref{app:exp} and~\ref{app:llm} for detailed experimental settings and metric definitions.

    % Sec.~\ref{sec:exp_main} compares DualSAE with standard SAEs, Secs.~\ref{sec:exp_branch}--\ref{sec:exp_expansion} analyze its branches, design, and robustness, and Secs.~\ref{sec:exp_topic}--\ref{sec:exp_language} cover topic modeling and language models. Supplementary~\ref{sec:hypo_sae} reports hypothesis generation.

\subsection{Performance between DualSAE and SAE}
\label{sec:exp_main}

\begin{figure}[!h]
    \centering
    % \vspace{-1em}
    \includegraphics[trim={0.5cm 1.38cm 0.8cm 0},clip,width=\linewidth]{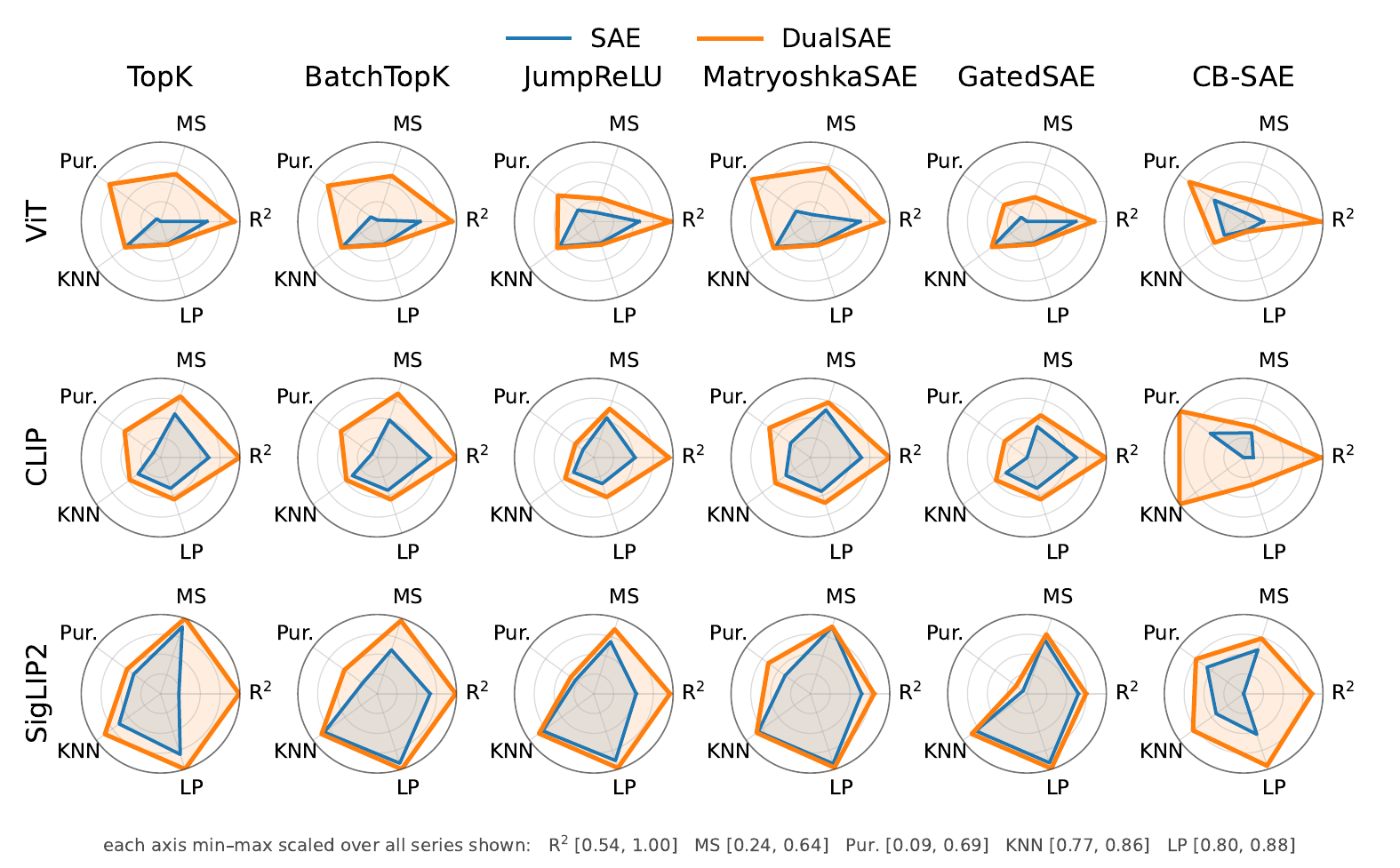}
    \vspace{-1em}
    \caption{\textbf{Comparison of SAE and DualSAE on ImageNet-1K.}
    Radar plots compare each SAE variant with its DualSAE counterpart across three encoders and six SAE variants using reconstruction \(R^2\), MonoScore, class purity, KNN accuracy, and LP accuracy. Across 18 settings, DualSAE generally improves reconstruction fidelity and interpretability. Similar results on OpenImagesV7 are presented in Figure~\ref{fig:radar_openimagesv7}.}
    \label{fig:radar_imagenet}
    %\vspace{-0.5em}
    \end{figure}

    We compare the six SAE variants with their DualSAE counterparts on ImageNet-1K across three encoders.
    % We report five metrics: reconstruction quality \(R^2\), MonoScore (MS)~\citep{monoscore}, class purity~\citep{nasiri-sarvi2026sparc}, KNN accuracy, and linear probing (LP) accuracy.
    For reconstruction quality, we adopt three metrics: \(R^2\), KNN accuracy, and linear probing (LP) accuracy. 
    For interpretability, we measure MonoScore (MS)~\citep{monoscore} and class purity~\citep{nasiri-sarvi2026sparc}.
    In Figure~\ref{fig:radar_imagenet}, across the 18 settings, DualSAE generally expands the radar profile relative to the corresponding SAE baselines. The gains are consistent in reconstruction fidelity \(R^2\), KNN accuracy, and LP accuracy (by 27.1\%, 1.8\%, and 1.2\% on average, respectively), indicating that adding the Residual Branch improves reconstruction fidelity.
    DualSAE also consistently improves MonoScore and class purity (by 31.2\% and 113.8\% on average), showing that the additional reconstruction pathway does not compromise interpretability and can instead improve the semantic coherence of the sparse representation.
    Similar advantages are also observed on OpenImagesV7 (Supplementary~\ref{app:detail_results}). 
    % The results show that our design is effective across datasets, SAE variants, and encoders.
    % on natural language representations of two corpora (OpenWebText and GitHub code), presented in Sec.~\ref{sec:exp_language}.
    % DualSAE also substantially enhances the reconstruction quality and interpretability on OpenImagesV7, shown in Supplementary~\ref{app:detail_results} and~\ref{app:recon_alignment}. 
    % Moreover, we also observe the advantage of DualSAE on natural language representations of two corpora (OpenWebText and GitHub code), presented in Sec.~\ref{sec:exp_language}.

    % \subsection{Why higher interpretability?}
    \subsection{Performance improvement analysis}
    \label{exp:why}
\begin{figure}[!h]
        \centering
        \vspace{-1em}
        \includegraphics[width=1\linewidth]{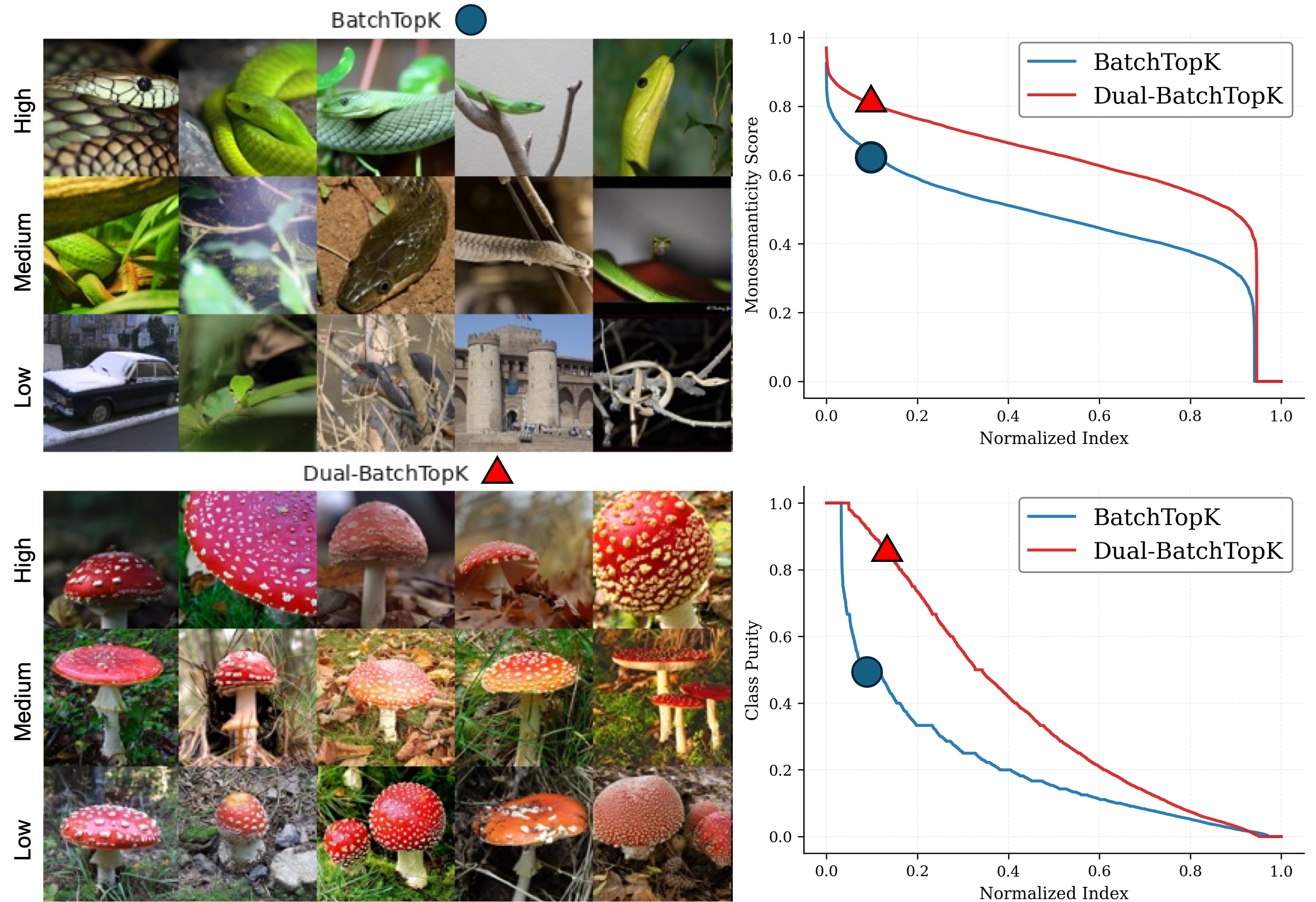}
        \caption{\textbf{Improved interpretability of sparse latents with DualSAE.}
        Left: For two latent dimensions selected at the same MonoScore rank, Dual-BatchTopK shows more coherent activating samples from high to low activations compared to BatchTopK. Right: MonoScore (top) and class purity (bottom) across all latents, showing consistently higher interpretability for DualSAE.}
        \label{fig:matched_latents}
        \vspace{-0.7em}
    \end{figure}

    % \subsubsection{Interpretability}
    % \textbf{Interpretability of sparse latents.} 
    \textbf{Interpretability analysis.} We examine whether the higher MonoScore and class purity reflect more interpretable latents. Figure~\ref{fig:matched_latents} compares BatchTopK and Dual-BatchTopK on SigLIP2 features of ImageNet-1K.
    For one latent per model at the same MonoScore rank, we show samples with high, medium, and low activations (left). The BatchTopK latent is only partially coherent: its strongest activations are mostly \textit{snakes}, but weaker ones include unrelated objects. In contrast, the Dual-BatchTopK latent activates on \textit{mushrooms} across the full activation range, \ie, its semantics remain stable beyond the top activations. The MonoScore and class-purity curves over all latents (right) quantitatively confirm that this is not limited to a single latent.

\begin{figure}[t]
            \centering
            \includegraphics[width=\linewidth]{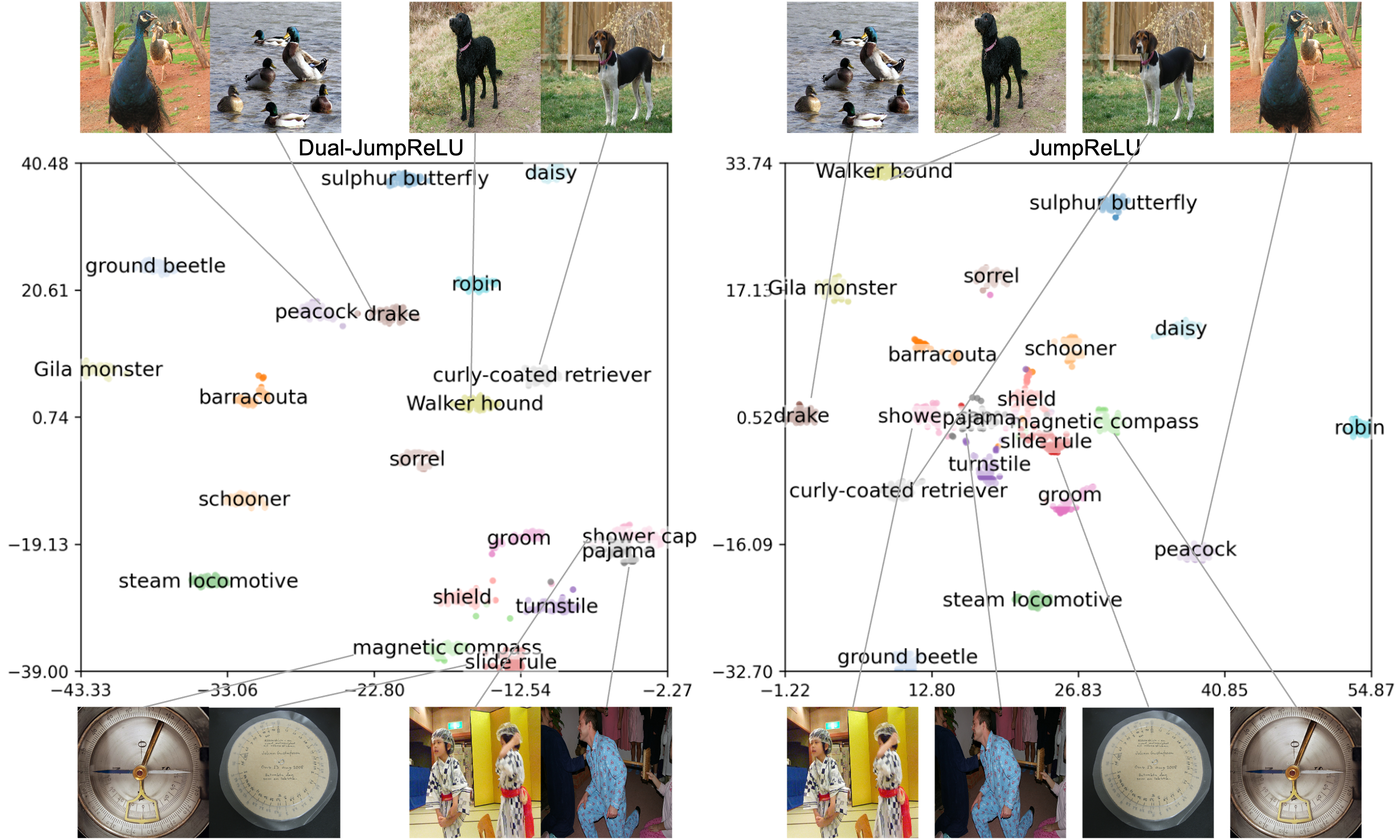}
            \caption{\textbf{t-SNE visualization of sparse representations.}
            Sparse codes of 20 randomly selected ImageNet-1K classes show that Dual-JumpReLU (left) forms better-separated local semantic neighborhoods than JumpReLU (right).}
            \label{fig:tsne_examples}
            \vspace{-1em}
        \end{figure}

    % \textbf{Semantic neighborhoods in sparse latent space.} 
    We further use t-SNE~\citep{tsne} to compare the sparse codes \(z\) of Dual-JumpReLU and JumpReLU for 20 random ImageNet-1K classes: if \(z\) is semantically coherent, related classes should lie nearby and unrelated ones apart. DualSAE forms clearer and more coherent local neighborhoods (Figure~\ref{fig:tsne_examples}), suggesting that the Sparse Branch captures dominant semantics more cleanly. In contrast, the baseline shows central crowding that mixes semantically different classes, suggesting that a single sparse bottleneck may entangle semantic and non-semantic information. 

    \textit{Why does an additional reconstruction pathway make the sparse code more interpretable?} A standard SAE must explain both semantic and non-semantic information, so part of its dictionary is spent on reconstruction-driven directions that mix unrelated inputs. This appears as the unrelated samples in Figure~\ref{fig:matched_latents} and the central crowding in Figure~\ref{fig:tsne_examples}. Once the Residual Branch absorbs this information, the sparse latents can specialize in single concepts, which enhance interpretability.

%     \subsection{Residual Branch vs. more parameters}
% \label{sec:exp_params}
\textbf{Capacity analysis.} DualSAE contains additional trainable parameters through its Residual Branch. To determine
whether its improvements arise solely from this increased capacity, we
construct parameter-matched SAE baselines by increasing their sparse-code
dimension until their total number of trainable parameters matches that of
DualSAE. All other training and evaluation settings are kept unchanged.
In Table~\ref{tab:parameter-matched-sae}, across all settings,
performance decreases when replacing DualSAE with SAEs.
Therefore, simply allocating the same parameter budget to a wider sparse
dictionary does not reproduce DualSAE's advantages. These results suggest that \textcolor{black}{{the benefit of the Residual Branch is not merely to provide more parameters, but to provide a separate representational pathway for non-semantic information.}}
% These results suggest that \textcolor{black}{\textit{the benefit of the Residual Branch is not merely to provide more parameters, but to provide a separate representational pathway for non-semantic information.}}
% These results suggest that \textcolor{black}{\textit{the benefit of the Residual Branch is not merely to provide more parameters, but to provide a separate representational pathway for non-semantic information.}}

\begin{table}[!h]
    \centering
    % %\vspace{-0.6em}
    \caption{
        \textbf{Parameter-matched SAE comparison.}
Each baseline uses a wider sparse dictionary so that its number of trainable parameters matches that of DualSAE.
We report $\Delta\text{metric}
= \text{metric}_{\text{matched SAE}}
- \text{metric}_{\text{DualSAE}}$; negative values favor DualSAE.
DualSAE outperforms the parameter-matched baselines in every setting, indicating  its gains are not due to additional parameters.
    }
    \label{tab:parameter-matched-sae}

    \setlength{\tabcolsep}{3.5pt}
    \resizebox{\linewidth}{!}{%
    \begin{tabular}{lllccccc@{\hspace{8pt}}ccccc}
        \toprule
        \multirow{2}{*}{\textbf{Encoder}}
        & \multirow{2}{*}{\textbf{Activation}}
        & \multirow{2}{*}{\textbf{Par.}}
        & \multicolumn{5}{c}{\textbf{ImageNet-1K}}
        & \multicolumn{5}{c}{\textbf{OpenImagesV7}} \\
        \cmidrule(lr){4-8}
        \cmidrule(lr){9-13}
        & &
        & \(\Delta R^2\)
        & \(\Delta\mathrm{MS}\)
        & \(\Delta\mathrm{Pur.}\)
        & \(\Delta\mathrm{KNN}\)
        & \(\Delta\mathrm{LP}\)
        & \(\Delta R^2\)
        & \(\Delta\mathrm{MS}\)
        & \(\Delta\mathrm{Pur.}\)
        & \(\Delta\mathrm{KNN}\)
        & \(\Delta\mathrm{LP}\) \\
        \midrule

        \multirow{3}{*}{ViT}
        & TopK & 18M
        & -0.002 & -0.004 & -0.004 & -0.003 & -0.007
        & -0.006 & -0.001 & -0.001 & -0.009 & -0.001 \\

        & BatchTopK & 18M
        & -0.004 & -0.003 & -0.006 & -0.009 & -0.001
        & -0.003 & -0.001 & -0.002 & -0.002 & -0.008 \\

        & JumpReLU & 18M
        & -0.113 & -0.246 & -0.235 & -0.003 & -0.002
        & -0.251 & -0.059 & -0.086 & -0.019 & -0.034 \\
        \midrule

        \multirow{3}{*}{CLIP}
        & TopK & 13M
        & -0.003 & -0.003 & -0.002 & -0.001 & -0.005
        & -0.010 & -0.009 & -0.001 & -0.011 & -0.008 \\

        & BatchTopK & 13M
        & -0.009 & -0.001 & -0.001 & -0.001 & -0.001
        & -0.001 & -0.002 & -0.007 & -0.008 & -0.001 \\

        & JumpReLU & 13M
        & -0.137 & -0.113 & -0.179 & -0.013 & -0.017
        & -0.144 & -0.039 & -0.116 & -0.004 & -0.019 \\
        \midrule

        \multirow{3}{*}{SigLIP2}
        & TopK & 20M
        & -0.006 & -0.002 & -0.002 & -0.001 & -0.004
        & -0.001 & -0.001 & -0.005 & -0.009 & -0.002 \\

        & BatchTopK & 20M
        & -0.001 & -0.006 & -0.012 & -0.001 & -0.008
        & -0.002 & -0.003 & -0.004 & -0.006 & -0.001 \\

        & JumpReLU & 20M
        & -0.119 & -0.050 & -0.245 & -0.005 & -0.009
        & -0.040 & -0.006 & -0.105 & -0.017 & -0.010 \\
        \bottomrule
    \end{tabular}%
    }
    %\vspace{-0.5em}
\end{table}

\begin{figure}[!h]
    \centering
    \includegraphics[width=\linewidth]{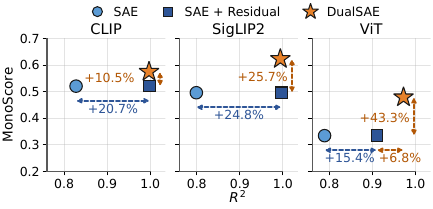}
    \caption{\textbf{DualSAE vs. SAE with Residual Branch} (BatchTopK on ImageNet-1K). SAE + Residual denotes a sequential baseline that freezes a pretrained SAE and trains only the Residual Branch, whereas DualSAE jointly optimizes both branches in its second training phase. Adding a Residual Branch improves $R^2$ by $15.4–24.8\%$ over the standard SAE. In contrast, DualSAE further improves MonoScore by $10.5–43.3\%$ over the sequential baseline while maintaining comparable or higher $R^2$, showing the benefits of joint optimization beyond additional residual capacity.}
    \vspace{-1em}
    \label{fig:joint_vs_seq}
    \end{figure}

    % \subsection{Reconstruction analysis}
    % \label{sec:exp_joint}
    \textbf{Residual analysis.} Adding the Residual Branch improves reconstruction fidelity, but does it also enhance interpretability? To test this, we compare DualSAE with a naive way of adding the Residual Branch: freeze a standard SAE and train the Residual Branch only. The main difference between the two designs is that the two branches are jointly trained in Stage 2 of DualSAE (Algorithm~\ref{alg:dualsae_training}), whereas the naive approach trains the two branches sequentially. This comparison also isolates the benefit of joint optimization beyond the additional reconstruction capacity provided by the Residual Branch.
    As shown in Figure~\ref{fig:joint_vs_seq} (BatchTopK on ImageNet-1K), the baseline raises \(R^2\) by 15--25\% but cannot improve interpretability: the frozen SAE keeps its sparse codes, so MonoScore is unchanged.
    In contrast, DualSAE trains both branches jointly and achieves comparable or higher \(R^2\) (within 0.3\% for CLIP and SigLIP2, 6.8\% higher for ViT) together with 10--43\% higher MonoScore.
    Similar results hold across other settings (Table~\ref{tab:frozen_vs_dualsae}).
    The results show that naively adding a Residual Branch improves reconstruction fidelity but provides no benefit to interpretability. In contrast, jointly optimizing both branches in DualSAE improves both reconstruction fidelity and interpretability.

\subsection{Branch decomposition and decorrelation loss}
\label{sec:exp_branch}\label{sec:exp_dcor}
    % We also use t-SNE to study the internal decomposition in DualSAE for three ImageNet-1K classes (Figure~\ref{fig:branch_decomposition}a). The sparse reconstruction \(x^M\) forms well-separated clusters (left), showing that the Sparse Branch captures the dominant semantics of each class. However, these clusters remain separated from the corresponding \(\hat{x}\) clusters (top-right), indicating that \(x^M\) alone does not fully recover the original embedding. The residual clusters $\hat{r}$ (bottom) overlap because they do not capture class-specific semantics. 

\begin{figure}[h]
    \centering
    % \vspace{-0.38em}
    \includegraphics[width=\linewidth]{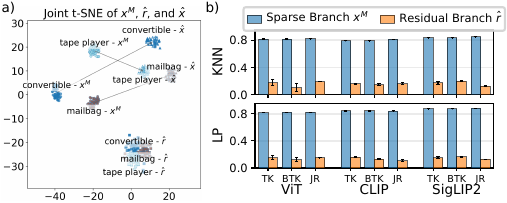}
    % \vspace{-1.3em}
    \caption{\textbf{Branch decomposition in DualSAE.} (a) Joint t-SNE of the Sparse Branch output \(x^M\), Residual Branch output \(\hat{r}\), and reconstruction \(\hat{x}\) for three ImageNet-1K classes: \(x^M\) forms class-separated clusters, while \(\hat{r}\) of all classes overlap. (b) KNN and LP top-1 accuracy on ImageNet-1K: \(x^M\) far exceeds \(\hat{r}\) in every setting, \ie, the Sparse Branch carries the class semantics.}
    \label{fig:branch_decomposition}
    %\vspace{-0.8em}
\end{figure}

    In this section, we investigate whether the Sparse Branch primarily captures semantic information, while the Residual Branch models non-semantic information.
    We use t-SNE to study the internal decomposition in DualSAE for three ImageNet-1K classes (Figure~\ref{fig:branch_decomposition}a). The sparse reconstruction \(x^M\) forms well-separated clusters (top-left), showing that the Sparse Branch captures the dominant semantic information of each class. However, the clusters remain distanced from the corresponding \(\hat{x}\) clusters (top-right), indicating that \(x^M\) alone does not fully recover the original embedding. 
    % The residual clusters $\hat{r}$ (bottom) lie closer together because they do not capture class-specific semantics. 
    % In contrast, the residual clusters \(\hat{r}\) overlap (bottom). Because \(x^M\) already captures the class-level component of each embedding, the Residual Branch mainly models variation within each class. With the class component removed, \(\hat{r}\) of different classes are centered near the same point and are therefore not separated by class.
    The residual outputs \(\hat{r}\) (bottom) overlap because they carry mainly non-semantic information: the Sparse Branch already encodes the semantics that distinguish the classes, so \(\hat{r}\) supplies the remaining non-semantic details that complete the reconstruction but are not class-specific.

    Additionally, we measure the class information in each branch by evaluating $x^M$ and $\hat{r}$ for classification on ImageNet-1K and OpenImagesV7 (KNN with $k=20$ cosine-similarity neighbors, LP with 10 training epochs, three seeds). In Figure~\ref{fig:branch_decomposition}b, \(x^M\) consistently retains high classification accuracy, whereas \(\hat{r}\) exhibits substantially lower accuracy (Supplementary~\ref{app:branch_oi} for OpenImagesV7). The results confirm that the Sparse Branch captures interpretable semantic information, while the Residual Branch provides non-semantic information for more faithful reconstruction.

    % To quantify this decomposition, we measure thehow much class information in each branch carries by evaluating $x^M$ and $\hat{r}$ for classification on ImageNet-1K and OpenImagesV7 (KNN with $k=20$ cosine-similarity neighbors, LP with 10 training epochs, three seeds). In all 36 comparisons, $x^M$ outperforms $\hat{r}$ (As shown in Figure~\ref{fig:branch_decomposition}b; for ImageNet-1K and in Supplementary~\ref{app:branch_oi} for OpenImagesV7). Thi, $x^M$ outperforms $\hat{r}$ in all 36 comparisons. The results supports that \textcolor{black}{\textit{the Sparse Branch captures interpretable semantic information, while the Residual Branch provides non-semantic information for more faithful reconstruction}}

\begin{figure}[h]
            %\vspace{-1em}
            \centering
            \includegraphics[width=\linewidth]{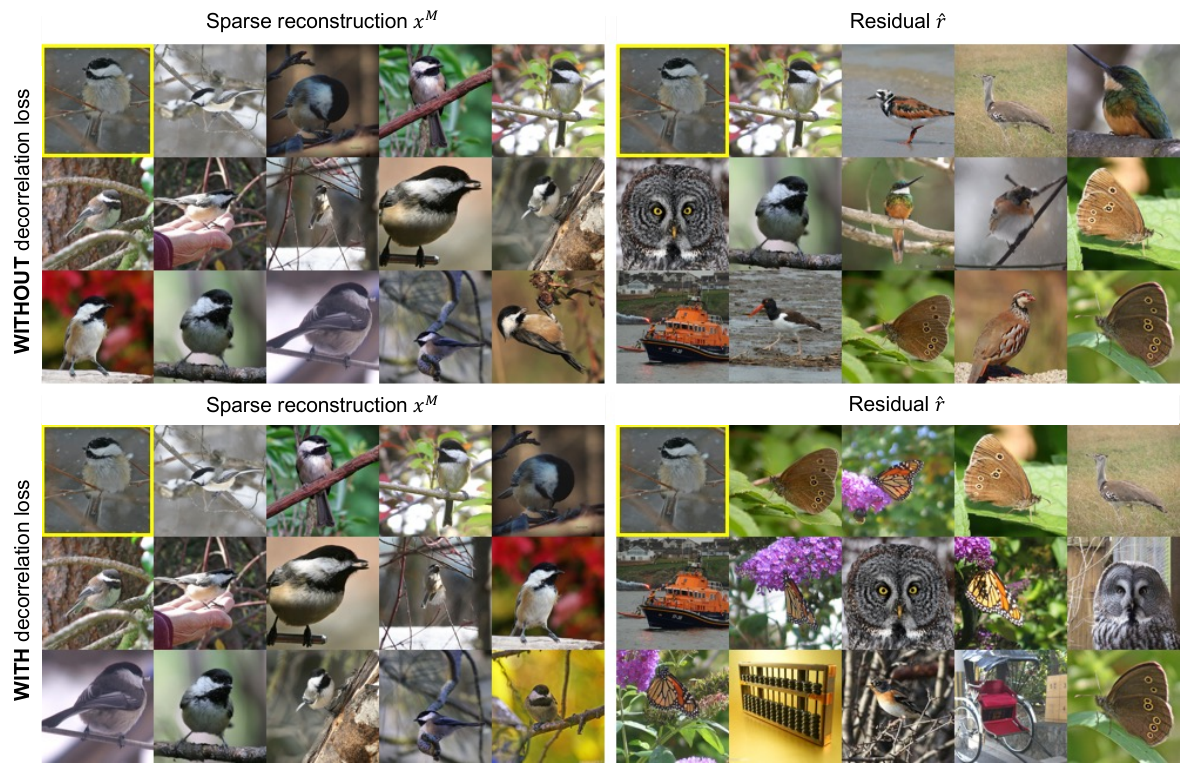}
            \caption{\textbf{Effect of decorrelation loss on branch separation.}
                For the same \textcolor{myyellow}{query image}, we compare nearest-neighbor retrievals from the sparse reconstruction \(x^M\) and residual \(\hat{r}\), with and without the decorrelation loss. The loss preserves coherent retrievals from \(x^M\) while making \(\hat{r}\) retrievals less similar to the query.}
            \label{fig:span_loss_retrieval}
            %\vspace{-1em}
        \end{figure}

    \textbf{Effect of the decorrelation loss.} We use nearest-neighbor retrieval to examine whether the decorrelation loss improves separation between the two branches. Figure~\ref{fig:span_loss_retrieval} compares retrievals from \(x^M\) and \(\hat r\) for Dual-BatchTopK trained on CLIP embeddings on ImageNet-1K. The intuition is that \(x^M\) should capture the dominant concept of the query, while \(\hat r\) should focus on information not already explained by the sparse reconstruction. Without the decorrelation loss, \(x^M\) retrieves coherent \textit{bird} images, while \(\hat r\) also retrieves several visually similar \textit{birds.} This overlap suggests that the Residual Branch partially reuses information already captured by the Sparse Branch rather than focusing on non-semantic information. With the decorrelation loss, retrievals from \(x^M\) remain semantically coherent and continue to capture the dominant \textit{bird} concept, whereas retrievals from \(\hat r\) become more diverse and less redundant with those from \(x^M\). This indicates that the decorrelation loss discourages the non-semantic residual from aligning with active sparse decoder directions. We further conduct a quantitative ablation by removing the decorrelation loss while keeping other parameters unchanged (Supplementary~\ref{sec:quan_dec_loss}), which leads to performance degradation across all settings. The results show that the decorrelation loss discourages branch overlap, encouraging the Sparse and Residual Branches to capture semantic and non-semantic information, respectively.

    \subsection{Ablation on expansion ratio}
    \label{sec:exp_expansion}
\begin{figure}[!h]
    \centering
    \includegraphics[width=\linewidth]{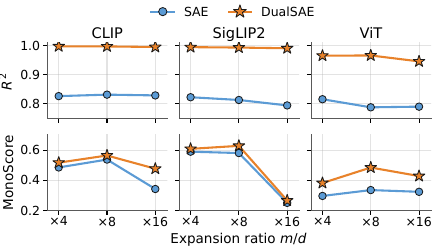}
    \caption{\textbf{Effect of the expansion ratio} (BatchTopK, ImageNet-1K). \(R^2\) (top) and MonoScore (bottom) of a standard SAE and DualSAE with expansion ratio of 4, 8, and 16: DualSAE is higher in both metrics at every ratio for all three encoders. Please refer to Figures~\ref{fig:expansion_in} and~\ref{fig:expansion_oi} for the results of other settings.}
    \label{fig:expansion_main}
    \end{figure}

    To test whether DualSAE's gains depend on the dictionary size, we train DualSAE and a standard SAE with expansion ratios $m/d\in\{4,8,16\}$ ($m$ is the dictionary size, and $d$ is the embedding dimension) for CLIP, SigLIP2, and ViT with TopK and BatchTopK on ImageNet-1K and OpenImagesV7.
    Figure~\ref{fig:expansion_main} show an representative result for BatchTopK on ImageNet-1K: DualSAE has higher \(R^2\) and MonoScore at every ratio for all three encoders. Supplementary~\ref{app:expansion} presents the results over all 36 pairs, which show DualSAE has higher \(R^2\) (by 0.12 to 0.33) and purity in every pair and higher MonoScore in 35 settings.
    The results indicate that DualSAE's advantages in reconstruction and interpretability generalize across dictionary sizes.

    \subsection{Language representation}
    \label{sec:exp_language}
    \begin{figure*}[t]
\centering
\scriptsize

\begin{minipage}[t]{0.55\textwidth}

\centering
\textbf{(a) ``Time-related" latent interpretation}

%\vspace{0.25em}

\begin{minipage}[t]{0.49\linewidth}%\vspace{0pt}% top-align the two boxes
\begin{tcolorbox}[
    colback=white,
    colframe=gray!25,
    boxrule=0.55pt,
    arc=1.5mm,
    left=1.3mm,
    right=1.3mm,
    top=1.2mm,
    bottom=1.2mm
]

\textbf{Dual-TopK}
\textcolor{gray!70}{latent 412}

%\vspace{0.35em}

\textbf{\textcolor{gray!80}{STRONGLY ACTIVATING}}
%\vspace{0.15em}

\begin{tabularx}{\linewidth}{@{}p{0.11\linewidth}X@{}}
\actscore{0.58} &
\texttt{two and a half }
\dualhl{years}
\texttt{ after it broke the news}
\\[0.35em]

\actscore{0.58} &
\texttt{which incorporated Atlus Online---two }
\dualhl{years}
\texttt{ ago}
\end{tabularx}

%\vspace{0.4em}

\textbf{\textcolor{gray!80}{MODERATELY ACTIVATING}}
%\vspace{0.15em}

\begin{tabularx}{\linewidth}{@{}p{0.11\linewidth}X@{}}
\actscore{0.14} &
\texttt{\$52.89 earlier, hovering close to a four-}
\dualhl{month}
\texttt{ low}
\\[0.35em]

\actscore{0.14} &
\texttt{the Titans in catches in his first }
\dualhl{year}
\end{tabularx}

\end{tcolorbox}
\end{minipage}
\hfill
\begin{minipage}[t]{0.49\linewidth}%\vspace{0pt}% top-align the two boxes
\begin{tcolorbox}[
    colback=white,
    colframe=gray!25,
    boxrule=0.55pt,
    arc=1.5mm,
    left=1.3mm,
    right=1.3mm,
    top=1.2mm,
    bottom=1.2mm
]

\textbf{TopK}
\textcolor{gray!70}{latent 2843}

%\vspace{0.35em}

\textbf{\textcolor{gray!80}{STRONGLY ACTIVATING}}
%\vspace{0.15em}

\begin{tabularx}{\linewidth}{@{}p{0.11\linewidth}X@{}}
\actscore{0.65} &
\texttt{Twenty-three }
\basehl{years}
\texttt{ after his death}
\\[0.35em]

\actscore{0.55} &
\missbl{five}
\basehl{years}
\texttt{ ago almost to the }
\basehl{day}
\end{tabularx}

%\vspace{0.4em}

\textbf{\textcolor{gray!80}{MODERATELY ACTIVATING}}
%\vspace{0.15em}

\begin{tabularx}{\linewidth}{@{}p{0.11\linewidth}X@{}}
\actscore{0.16} &
\missbl{Living}
\texttt{ a few }
\missbl{years}
\texttt{ of your life like}
\\[0.35em]

\actscore{0.15} &
\texttt{Having been in the established mental }
\missbl{health system}
\texttt{ for a }
\missbl{decade}
\texttt{ now}
\end{tabularx}

\end{tcolorbox}
\end{minipage}

\end{minipage}
\hfill
\begin{minipage}[t]{0.44\textwidth}

\centering
\textbf{(b) ``Positive-sentiment" steering}

%\vspace{0.25em}

\begin{tcolorbox}[
    colback=gray!3,
    colframe=gray!30,
    boxrule=0.5pt,
    arc=1.2mm,
    left=1.3mm,
    right=1.3mm,
    top=0.9mm,
    bottom=0.9mm
]
\textbf{Unsteered}

%\vspace{0.03em}

It took us two hours to find the place, and it took us two hours to get there.
It was a very small village. There were two houses and a barn.
I had to go in and put on my shoes. I had to go into the barn and put on
\end{tcolorbox}

%\vspace{0.08em}

\begin{tcolorbox}[
    colback=gray!3,
    colframe=gray!30,
    boxrule=0.5pt,
    arc=1.2mm,
    left=1.3mm,
    right=1.3mm,
    top=0.9mm,
    bottom=0.9mm
]
\textbf{TopK SAE}
\hfill

%\vspace{0.03em}

It took us two hours to find the spot,'' she said.
``And they were not even in their own area.
They didn't know where to get rid of the money.''
\end{tcolorbox}

%\vspace{0.08em}

\begin{tcolorbox}[
    colback=green!4,
    colframe=green!35!black,
    boxrule=0.55pt,
    arc=1.2mm,
    left=1.3mm,
    right=1.3mm,
    top=0.9mm,
    bottom=0.9mm
]
\textbf{Dual-TopK SAE}
\hfill

%\vspace{0.03em}

It took us two hours to find the
perfect combination of quality and price
to allow us to buy the property.
We were \textbf{very impressed} with the
quality of the property and the service we received.
\end{tcolorbox}

\end{minipage}

%\vspace{-0.3em}

\caption{
\textbf{Qualitative analysis on language representations.} \textbf{(a)} ``Time-related" latents learned by Dual-TopK and TopK. Dual-TopK responds more selectively to genuine \dualhlcap{``time-related" concepts}, whereas TopK exhibits \missblcap{missed temporal concepts} and activations on \missblcap{unrelated tokens}. \textbf{(b)} Positive-sentiment steering. Compared with the unsteered generation and TopK steering, Dual-TopK produces a coherent flow containing clear positive-sentiment expressions.
}
\label{fig:text_qualitative_combined}

%\vspace{-2em}
\vspace{-1em}
\end{figure*}
    We further evaluate DualSAE on language-model representations using Pythia~\cite{biderman2023pythia}
    activations on OpenWebText (OWT)~\cite{Gokaslan2019OpenWeb}
    and GitHub code (GHC)~\cite{codeparrot_github_code}.
    We evaluate SAEs' ability to preserve the
    original language-model behavior using $\Delta$CE and KL divergence,
    reconstruct activations using $R^2$, and learn interpretable features using
    MonoScore (MS).
    As shown in Table~\ref{tab:text_main_results}, DualSAE
    consistently improves reconstruction quality and interpretability while substantially
    reducing perturbations to the original model across both corpora, suggesting that DualSAE's advantages
    generalize beyond vision representations to both natural language and code.
    We also analyze the activating pattern in Figure~\ref{fig:text_qualitative_combined}a. Both TopK and Dual-TopK learn
    latents associated with temporal expressions, but the Dual-TopK latent
    responds more selectively to genuine time-related concepts.
    In contrast, the TopK SAE latent misses some temporal concepts and also
    activates on unrelated tokens. Please refer to Supplementary~\ref{app:llm} for the detailed setup, metric definitions, and more qualitative examples.

\begin{table}[t]
    \centering
    \caption{
        \textbf{Language-model evaluation} on OpenWebText (OWT) and
        GitHub code (GHC) at layer 12 of Pythia.
        Lower $\Delta$CE and KL indicate better preservation of the original
        language-model behavior; higher $R^2$ and MS indicate better
        reconstruction and monosemanticity. DualSAE improves all four metrics over its standard counterpart for every SAE variant on both corpora.
    }
    \label{tab:text_main_results}

    \setlength{\tabcolsep}{3.5pt}
    \resizebox{\linewidth}{!}{%
        \begin{tabular}{clcccc}
            \toprule
            \textbf{Corpus} &
            \textbf{Method} &
            \boldmath$\Delta$\textbf{CE} $\downarrow$ &
            \textbf{KL} $\downarrow$ &
            \boldmath$R^2$ $\uparrow$ &
            \textbf{MS} $\uparrow$ \\
            \midrule

            \multirow{6}{*}{\centering OWT}
            &
            JumpReLU &
            0.7412\textsubscript{0.0201} &
            0.5498\textsubscript{0.0011} &
            0.8994\textsubscript{0.1209} &
            0.4995\textsubscript{0.1225} \\

            &
            Dual-JumpReLU &
            \textbf{0.1423}\textsubscript{\textbf{0.0097}} &
            \textbf{0.0532}\textsubscript{\textbf{0.0101}} &
            \textbf{0.9732}\textsubscript{\textbf{0.0464}} &
            \textbf{0.6882}\textsubscript{\textbf{0.0743}} \\
            \cmidrule(lr){2-6}

            &
            TopK &
            0.5538\textsubscript{0.0193} &
            0.5608\textsubscript{0.0201} &
            0.9677\textsubscript{0.0559} &
            0.4317\textsubscript{0.0541} \\

            &
            Dual-TopK &
            \textbf{0.0105}\textsubscript{\textbf{0.0022}} &
            \textbf{0.0131}\textsubscript{\textbf{0.0016}} &
            \textbf{0.9896}\textsubscript{\textbf{0.0180}} &
            \textbf{0.6247}\textsubscript{\textbf{0.0236}} \\
            \cmidrule(lr){2-6}

            &
            BatchTopK &
            0.6587\textsubscript{0.0442} &
            0.6661\textsubscript{0.0441} &
            0.9432\textsubscript{0.0984} &
            0.4982\textsubscript{0.1455} \\

            &
            Dual-BatchTopK &
            \textbf{0.2118}\textsubscript{\textbf{0.0229}} &
            \textbf{0.3129}\textsubscript{\textbf{0.0206}} &
            \textbf{0.9887}\textsubscript{\textbf{0.0196}} &
            \textbf{0.9695}\textsubscript{\textbf{0.0078}} \\

            \midrule

            \multirow{6}{*}{\centering GHC}
            &
            JumpReLU &
            0.7843\textsubscript{0.1923} &
            0.6709\textsubscript{0.1009} &
            0.9223\textsubscript{0.0976} &
            0.5328\textsubscript{0.0843} \\

            &
            Dual-JumpReLU &
            \textbf{0.2544}\textsubscript{\textbf{0.0029}} &
            \textbf{0.3212}\textsubscript{\textbf{0.0084}} &
            \textbf{0.9865}\textsubscript{\textbf{0.0234}} &
            \textbf{0.8420}\textsubscript{\textbf{0.0498}} \\
            \cmidrule(lr){2-6}

            &
            TopK &
            0.4108\textsubscript{0.1877} &
            0.4310\textsubscript{0.1865} &
            0.9676\textsubscript{0.0561} &
            0.4194\textsubscript{0.1420} \\

            &
            Dual-TopK &
            \textbf{0.1954}\textsubscript{\textbf{0.0903}} &
            \textbf{0.2033}\textsubscript{\textbf{0.0861}} &
            \textbf{0.9937}\textsubscript{\textbf{0.0109}} &
            \textbf{0.5182}\textsubscript{\textbf{0.0534}} \\
            \cmidrule(lr){2-6}

            &
            BatchTopK &
            2.4065\textsubscript{0.1980} &
            2.4066\textsubscript{0.1890} &
            0.9386\textsubscript{0.1063} &
            0.5637\textsubscript{0.1236} \\

            &
            Dual-BatchTopK &
            \textbf{1.7612}\textsubscript{\textbf{0.1293}} &
            \textbf{0.7730}\textsubscript{\textbf{0.0302}} &
            \textbf{0.9790}\textsubscript{\textbf{0.0364}} &
            \textbf{0.9870}\textsubscript{\textbf{0.0039}} \\

            \bottomrule
        \end{tabular}%
    }
    % \vspace{-0.5em}
\end{table}

    \subsection{Downstream applications}
    \label{sec:exp_topic}
\begin{figure}[!h]
        \centering
        %\vspace{-1em}
        \vspace{-0.8em}
        \includegraphics[width=\linewidth]{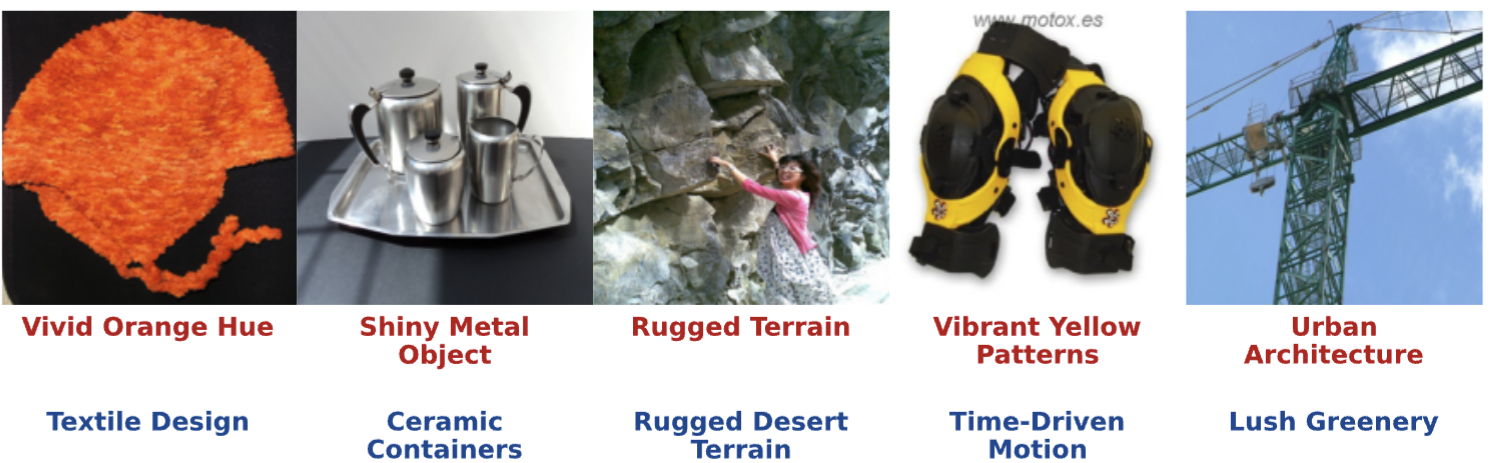}
        %\vspace{-0.3em}
        \caption{\textbf{Image--topic assignment comparison} (TopK on SigLIP2). Dominant topics from \textcolor{red}{{DualSAE}} and \textcolor{blue}{{SAE}}: DualSAE's are visually grounded, while SAE's are unrelated.}
        \label{fig:img2topic_sigclip}
        %\vspace{-1em}
        \vspace{-0.6em}
    \end{figure}

    \textbf{Topic modeling.} Each sparse feature can be viewed as a \emph{topic atom}: an image's active features indicate which visual concepts it contains. We apply the same topic-modeling pipeline in SAE-TM~\citep{girrbach2026saetm}, which clusters features into \(K\) topics summarized by their top caption words, to standard SAE features and DualSAE's sparse features. In Figure~\ref{fig:img2topic_sigclip}, DualSAE yields concrete labels for the dominant object, attribute, or scene (\eg, \textit{Shiny Metal Object} for a metal serving set), whereas the standard SAE assigns topics from weak or spurious associations (\eg, \textit{Ceramic Containers}). Quantitatively, on ImageNet-1K and OpenImagesV7, DualSAE increases topic diversity \(D\) (non-redundancy of top words) in 89 of 90 configurations and intruder accuracy \(C_I\) (whether an LLM judge spots an injected unrelated concept) in 74 of them (Table~\ref{tab:topic_modeling}; Supplementary~\ref{sec:topic_model}), \ie, it yields more concrete and interpretable topics.

    \begin{figure}[!b]
        \centering
        \vspace{-0.8em}
        {\scriptsize\textbf{JumpReLU}: ``dogs in a car with snow and trees in the background''\par}
        \includegraphics[width=\linewidth,trim=39.3 123.3 2.8 204.6,clip]{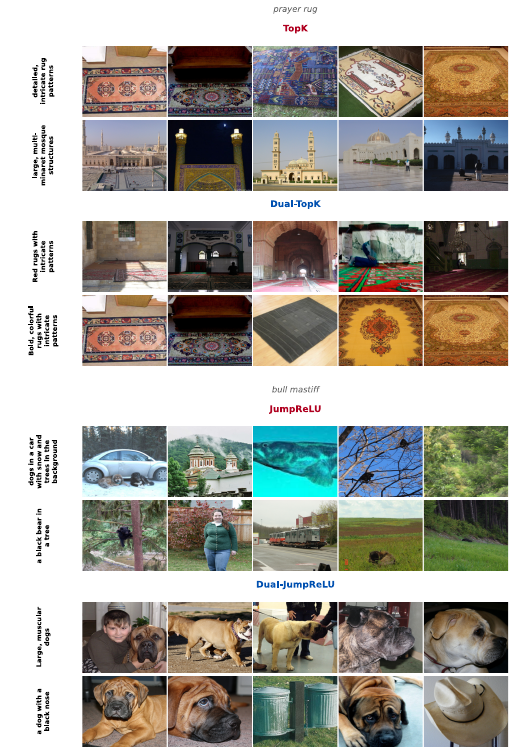}
        {\scriptsize\textbf{Dual-JumpReLU}: ``large, muscular dogs''\par}
        \includegraphics[width=\linewidth,trim=39.3 39.0 2.8 288.8,clip]{Figures/hypo_SAE.pdf}
        \caption{\textbf{Hypothesis generation for \textit{bull mastiff}.} Top latent of each model with its VLM hypothesis and five highest-activating images: JumpReLU's mixes unrelated scenes, whereas Dual-JumpReLU's is coherent and class-relevant.}
        \label{fig:hypo_main}
    \end{figure}

    \textbf{Hypothesis generation.} Following HypotheSAEs~\citep{movva2025sparse}, we select the 16 latents most discriminative of each class, have a VLM describe each as a hypothesis, and test on held-out images how many are significantly associated with the class and how well they predict it (AUROC). For \textit{bull mastiff}, the top JumpReLU latent mixes unrelated scenes, whereas Dual-JumpReLU's captures \textit{large, muscular dogs} (Figure~\ref{fig:hypo_main}). Overall, DualSAE yields more significant hypotheses in 17 of 18 configurations and higher AUROC in all six JumpReLU ones and at most 0.005 lower elsewhere (Supplementary~\ref{sec:hypo_sae}).

    \textbf{LLM steering.} Figure~\ref{fig:text_qualitative_combined}b illustrates feature steering using a \textit{positive-sentiment} latent to modify Pythia's activations at layer~12. While the unsteered and TopK-steered generations do not exhibit a clear positive shift, steering with Dual-TopK produces a coherent continuation containing explicit positive expressions (more in Supplementary~\ref{app:llm_steering}). Together, these results show that DualSAE's advantages extend to downstream applications.

    \section{Conclusion}
        We introduced DualSAE, which augments standard SAEs with a non-semantic Residual Branch to separate the reconstruction and interpretability objectives.
        Instead of forcing all information through a single sparse dictionary, DualSAE uses the Sparse Branch to learn semantic information and the Residual Branch to model the non-semantic residual, and our decorrelation loss keeps the Residual Branch from redundantly aligning with active sparse decoder directions.
        % Across multiple vision encoders, activation functions, and datasets, DualSAE improves reconstruction quality, monosemanticity, class purity, and downstream performance. Qualitative analyses further show that the Sparse Branch captures coherent semantic concepts, while the residual branch contributes fine-grained information.
        DualSAE improves both reconstruction fidelity and interpretability across diverse settings, and extends to language models and downstream applications.
        DualSAE has limitations. Our experiments use frozen representations of pre-trained vision encoders and a language model, so end-to-end trained and multimodal reasoning models remain untested, and DualSAE adds training choices, namely the decorrelation coefficient, the Residual Branch design, and the warm-up length. Future work should study more modalities and adaptive schedules that balance sparse and residual learning.
        Overall, our results suggest that separating sparse semantic structure from residual information is a promising direction for building more interpretable and faithful decompositions of modern representation spaces.

    \ifarxiv % acknowledgment only in the arXiv/camera-ready version; omitted from the anonymous review version
    \section*{Acknowledgment}
        This research was supported by NSERC-DG RGPIN-2022-05378, Amazon Research Award, and Gina Cody RIF, Computational resources were provided in part by Calcul Qu\'{e}bec (\url{www.calculquebec.ca}) and the Digital Research Alliance of Canada (\url{www.alliancecan.ca}) (Mahdi S. Hosseini); National Science Foundation (NSF) grants IIS-2123920, IIS-2212046 (Dimitris Samaras).
    \fi

{
    \small
    \bibliographystyle{ieeenat_fullname}
    \bibliography{main}

@String(IJCV = {Int. J. Comput. Vis.})

@String(ICLR = {Int. Conf. Learn. Represent.})

@String(IJCV  = {IJCV})

@String(ICLR  = {ICLR})

@inproceedings{
    monoscore,
    title={Sparse Autoencoders Learn Monosemantic Features in Vision-Language Models},
    author={Mateusz Pach and Shyamgopal Karthik and Quentin Bouniot and Serge Belongie and Zeynep Akata},
    booktitle={The Thirty-ninth Annual Conference on Neural Information Processing Systems},
    year={2025},
    url={https://openreview.net/forum?id=DaNnkQJSQf}
}

@article{
nasiri-sarvi2026sparc,
title={{SPARC}: Concept-Aligned Sparse Autoencoders for Cross-Model and Cross-Modal Interpretability},
author={Ali Nasiri-Sarvi and Hassan Rivaz and Mahdi S. Hosseini},
journal={Transactions on Machine Learning Research},
issn={2835-8856},
year={2026},
url={https://openreview.net/forum?id=IJfvoc2GbZ},
note={}
}

@article{tsne,
  title={Visualizing data using t-SNE.},
  author={Van der Maaten, Laurens and Hinton, Geoffrey},
  journal={Journal of machine learning research},
  volume={9},
  number={11},
  year={2008}
}

@article{templeton2024scaling,
       title={Scaling Monosemanticity: Extracting Interpretable Features from Claude 3 Sonnet},
       author={Templeton, Adly and Conerly, Tom and Marcus, Jonathan and Lindsey, Jack and Bricken, Trenton and Chen, Brian and Pearce, Adam and Citro, Craig and Ameisen, Emmanuel and Jones, Andy and Cunningham, Hoagy and Turner, Nicholas L and McDougall, Callum and MacDiarmid, Monte and Freeman, C. Daniel and Sumers, Theodore R. and Rees, Edward and Batson, Joshua and Jermyn, Adam and Carter, Shan and Olah, Chris and Henighan, Tom},
       year={2024},
       journal={Transformer Circuits Thread},
       url={https://transformer-circuits.pub/2024/scaling-monosemanticity/index.html}
    }

@inproceedings{cbsae,
    title={Interpretable and Steerable Concept Bottleneck Sparse Autoencoders},
    author={Kulkarni, Akshay and Weng, Tsui-Wei and Narayanaswamy, Vivek and Liu, Shusen and Sakla, Wesam and Thopalli, Kowshik},
    booktitle={IEEE/CVF Conference on Computer Vision and Pattern Recognition},
    year={2026},
}

@misc{codeparrot_github_code,
  author       = {{CodeParrot}},
  title        = {GitHub Code Dataset},
  year         = {2022},
  howpublished = {\url{https://huggingface.co/datasets/codeparrot/github-code}},
  note         = {Hugging Face dataset}
}

@misc{Gokaslan2019OpenWeb,
    title={OpenWebText Corpus},
    author={Gokaslan, Aaron and Cohen, Vanya and Pavlick, Ellie and Tellex, Stefanie},
    howpublished={\url{http://Skylion007.github.io/OpenWebTextCorpus}},
    year={2019}
}

@inproceedings{
gatedsae,
title={Improving Sparse Decomposition of Language Model Activations with Gated Sparse Autoencoders},
author={Senthooran Rajamanoharan and Arthur Conmy and Lewis Smith and Tom Lieberum and Vikrant Varma and Janos Kramar and Rohin Shah and Neel Nanda},
booktitle={The Thirty-eighth Annual Conference on Neural Information Processing Systems},
year={2024},
url={https://openreview.net/forum?id=zLBlin2zvW}
}

@misc{jumprelu,
      title={Jumping Ahead: Improving Reconstruction Fidelity with JumpReLU Sparse Autoencoders}, 
      author={Senthooran Rajamanoharan and Tom Lieberum and Nicolas Sonnerat and Arthur Conmy and Vikrant Varma and János Kramár and Neel Nanda},
      year={2024},
      eprint={2407.14435},
      archivePrefix={arXiv},
      primaryClass={cs.LG},
      url={https://arxiv.org/abs/2407.14435}, 
}

@inproceedings{
    batchtopksae,
    title={BatchTopK Sparse Autoencoders},
    author={Bart Bussmann and Patrick Leask and Neel Nanda},
    booktitle={NeurIPS 2024 Workshop on Scientific Methods for Understanding Deep Learning},
    year={2024},
    url={https://openreview.net/forum?id=d4dpOCqybL}
}

@misc{bussmann2025matryoshka,
      title={Learning Multi-Level Features with Matryoshka Sparse Autoencoders}, 
      author={Bart Bussmann and Noa Nabeshima and Adam Karvonen and Neel Nanda},
      year={2025},
      eprint={2503.17547},
      archivePrefix={arXiv},
      primaryClass={cs.LG},
      url={https://arxiv.org/abs/2503.17547}, 
}

@inproceedings{
korznikov2026ortsae,
title={Ort{SAE}: Orthogonal Sparse Autoencoders Uncover Atomic Features},
author={Anton Korznikov and Andrey V. Galichin and Alexey Dontsov and Oleg Rogov and Ivan Oseledets and Elena Tutubalina},
booktitle={ICLR 2026 Workshop on Representational Alignment (Re{\textasciicircum}4-Align)},
year={2026},
url={https://openreview.net/forum?id=f0xiIBVbNk}
}

@inproceedings{
    scalingsae,
    title={Scaling and evaluating sparse autoencoders},
    author={Leo Gao and Tom Dupre la Tour and Henk Tillman and Gabriel Goh and Rajan Troll and Alec Radford and Ilya Sutskever and Jan Leike and Jeffrey Wu},
    booktitle={The Thirteenth International Conference on Learning Representations},
    year={2025},
    url={https://openreview.net/forum?id=tcsZt9ZNKD}
}

@inproceedings{topksae,
    title={k-Sparse Autoencoders},
    author={Alireza Makhzani and Brendan Frey},
    booktitle={The Second International Conference on Learning Representations},
    year={2014},
    url={https://arxiv.org/abs/1312.5663}
}

@inproceedings{
    cunningham2023,
    title={Sparse Autoencoders Find Highly Interpretable Features in Language Models},
    author={Robert Huben and Hoagy Cunningham and Logan Riggs Smith and Aidan Ewart and Lee Sharkey},
    booktitle={The Twelfth International Conference on Learning Representations},
    year={2024},
    url={https://openreview.net/forum?id=F76bwRSLeK}
}

@InProceedings{clip,
  title = 	 {Learning Transferable Visual Models From Natural Language Supervision},
  author =       {Radford, Alec and Kim, Jong Wook and Hallacy, Chris and Ramesh, Aditya and Goh, Gabriel and Agarwal, Sandhini and Sastry, Girish and Askell, Amanda and Mishkin, Pamela and Clark, Jack and Krueger, Gretchen and Sutskever, Ilya},
  booktitle = 	 {Proceedings of the 38th International Conference on Machine Learning},
  pages = 	 {8748--8763},
  year = 	 {2021},
  editor = 	 {Meila, Marina and Zhang, Tong},
  volume = 	 {139},
  series = 	 {Proceedings of Machine Learning Research},
  month = 	 {18--24 Jul},
  publisher =    {PMLR},
  url = 	 {https://proceedings.mlr.press/v139/radford21a.html},
}

@article{imagenet15russakovsky,
    Author = {Olga Russakovsky and Jia Deng and Hao Su and Jonathan Krause and Sanjeev Satheesh and Sean Ma and Zhiheng Huang and Andrej Karpathy and Aditya Khosla and Michael Bernstein and Alexander C. Berg and Li Fei-Fei},
    Title = {{ImageNet Large Scale Visual Recognition Challenge}},
    Year = {2015},
    journal   = {International Journal of Computer Vision (IJCV)},
    doi = {10.1007/s11263-015-0816-y},
    volume={115},
    number={3},
    pages={211-252}
}

@misc{elhage2022toy,
      title={Toy Models of Superposition}, 
      author={Nelson Elhage and Tristan Hume and Catherine Olsson and Nicholas Schiefer and Tom Henighan and Shauna Kravec and Zac Hatfield-Dodds and Robert Lasenby and Dawn Drain and Carol Chen and Roger Grosse and Sam McCandlish and Jared Kaplan and Dario Amodei and Martin Wattenberg and Christopher Olah},
      year={2022},
      eprint={2209.10652},
      archivePrefix={arXiv},
      primaryClass={cs.LG},
      url={https://arxiv.org/abs/2209.10652}, 
}

@inproceedings{
    cui2026on,
    title={On the Limits of Sparse Autoencoders: A Theoretical Framework and Reweighted Remedy},
    author={Jingyi Cui and Qi Zhang and Yifei Wang and Yisen Wang},
    booktitle={The Fourteenth International Conference on Learning Representations},
    year={2026},
    url={https://openreview.net/forum?id=DSOTgzeH3w}
}

@article{lee2006efficient,
  title={Efficient sparse coding algorithms},
  author={Lee, Honglak and Battle, Alexis and Raina, Rajat and Ng, Andrew},
  journal={Advances in neural information processing systems},
  volume={19},
  year={2006}
}

@article{olshausen1996emergence,
  title={Emergence of simple-cell receptive field properties by learning a sparse code for natural images},
  author={Olshausen, Bruno A and Field, David J},
  journal={Nature},
  volume={381},
  number={6583},
  pages={607--609},
  year={1996},
  publisher={Nature Publishing Group UK London}
}

@article{bricken2023monosemanticity,
       title={Towards Monosemanticity: Decomposing Language Models With Dictionary Learning},
       author={Bricken, Trenton and Templeton, Adly and Batson, Joshua and Chen, Brian and Jermyn, Adam and Conerly, Tom and Turner, Nick and Anil, Cem and Denison, Carson and Askell, Amanda and Lasenby, Robert and Wu, Yifan and Kravec, Shauna and Schiefer, Nicholas and Maxwell, Tim and Joseph, Nicholas and Hatfield-Dodds, Zac and Tamkin, Alex and Nguyen, Karina and McLean, Brayden and Burke, Josiah E and Hume, Tristan and Carter, Shan and Henighan, Tom and Olah, Christopher},
       year={2023},
       journal={Transformer Circuits Thread},
       note={https://transformer-circuits.pub/2023/monosemantic-features/index.html}
    }

@misc{SigLIP2,
      title={SigLIP 2: Multilingual Vision-Language Encoders with Improved Semantic Understanding, Localization, and Dense Features}, 
      author={Michael Tschannen and Alexey Gritsenko and Xiao Wang and Muhammad Ferjad Naeem and Ibrahim Alabdulmohsin and Nikhil Parthasarathy and Talfan Evans and Lucas Beyer and Ye Xia and Basil Mustafa and Olivier Hénaff and Jeremiah Harmsen and Andreas Steiner and Xiaohua Zhai},
      year={2025},
      eprint={2502.14786},
      archivePrefix={arXiv},
      primaryClass={cs.CV},
      url={https://arxiv.org/abs/2502.14786}, 
}

@inproceedings{
    vit,
    title={An Image is Worth 16x16 Words: Transformers for Image Recognition at Scale},
    author={Alexey Dosovitskiy and Lucas Beyer and Alexander Kolesnikov and Dirk Weissenborn and Xiaohua Zhai and Thomas Unterthiner and Mostafa Dehghani and Matthias Minderer and Georg Heigold and Sylvain Gelly and Jakob Uszkoreit and Neil Houlsby},
    booktitle={International Conference on Learning Representations},
    year={2021},
    url={https://openreview.net/forum?id=YicbFdNTTy}
}

@article{reluSAE,
  title={Sparse autoencoder},
  author={Ng, Andrew},
  journal={CS294A Lecture notes},
  volume={72},
  number={2011},
  pages={1--19},
  year={2011}
}

@article{OpenImages,
  author = {Alina Kuznetsova and Hassan Rom and Neil Alldrin and Jasper Uijlings and Ivan Krasin and Jordi Pont-Tuset and Shahab Kamali and Stefan Popov and Matteo Malloci and Alexander Kolesnikov and Tom Duerig and Vittorio Ferrari},
  title = {The Open Images Dataset V4: Unified image classification, object detection, and visual relationship detection at scale},
  year = {2020},
  journal = {IJCV}
}

@inproceedings{mairal2009online,
  title={Online dictionary learning for sparse coding},
  author={Mairal, Julien and Bach, Francis and Ponce, Jean and Sapiro, Guillermo},
  booktitle={Proceedings of the 26th annual international conference on machine learning},
  pages={689--696},
  year={2009}
}

@inproceedings{girrbach2026saetm,
    title     = {Sparse Autoencoders are Topic Models},
    author    = {Girrbach, Leander and Akata, Zeynep},
    booktitle = {Proceedings of the 43rd International Conference on Machine Learning (ICML)},
    year      = {2026}
}

@inproceedings{
    movva2025sparse,
    title={Sparse Autoencoders for Hypothesis Generation},
    author={Rajiv Movva and Kenny Peng and Nikhil Garg and Jon Kleinberg and Emma Pierson},
    booktitle={Forty-second International Conference on Machine Learning},
    year={2025},
    url={https://openreview.net/forum?id=4R0pugRyN5}
}

@inproceedings{
    deng2026decompose,
    title={Decompose Sparsely Where You Should, Absorb Densely Where You Should Not},
    author={Ruixuan Deng and Zehao Jin and Zekun Wang and Zihan Dong},
    booktitle={Mechanistic Interpretability Workshop at ICML 2026},
    year={2026},
    url={https://openreview.net/forum?id=GhH9CY7Ohk}
}

@article{
yang2026alignsae,
title={Align{SAE}: Concept-Aligned Sparse Autoencoders},
author={Minglai Yang and Xinyu Guo and Zhengliang Shi and Jinhe Bi and Steven Bethard and Mihai Surdeanu and Liangming Pan},
journal={Transactions on Machine Learning Research},
issn={2835-8856},
year={2026},
url={https://openreview.net/forum?id=I9UjKxW4nq},
note={}
}

@inproceedings{
    wu2025axbench,
    title={AxBench: Steering {LLM}s? Even Simple Baselines Outperform Sparse Autoencoders},
    author={Zhengxuan Wu and Aryaman Arora and Atticus Geiger and Zheng Wang and Jing Huang and Dan Jurafsky and Christopher D Manning and Christopher Potts},
    booktitle={Forty-second International Conference on Machine Learning},
    year={2025},
    url={https://openreview.net/forum?id=K2CckZjNy0}
}

@misc{abdin2024phi4,
      title={Phi-4 Technical Report}, 
      author={Marah Abdin and Jyoti Aneja and Harkirat Behl and Sébastien Bubeck and Ronen Eldan and Suriya Gunasekar and Michael Harrison and Russell J. Hewett and Mojan Javaheripi and Piero Kauffmann and James R. Lee and Yin Tat Lee and Yuanzhi Li and Weishung Liu and Caio C. T. Mendes and Anh Nguyen and Eric Price and Gustavo de Rosa and Olli Saarikivi and Adil Salim and Shital Shah and Xin Wang and Rachel Ward and Yue Wu and Dingli Yu and Cyril Zhang and Yi Zhang},
      year={2024},
      eprint={2412.08905},
      archivePrefix={arXiv},
      primaryClass={cs.CL},
      url={https://arxiv.org/abs/2412.08905}, 
}

@inproceedings{xiao2024cpack,
author = {Xiao, Shitao and Liu, Zheng and Zhang, Peitian and Muennighoff, Niklas and Lian, Defu and Nie, Jian-Yun},
title = {C-Pack: Packed Resources For General Chinese Embeddings},
year = {2024},
isbn = {9798400704314},
publisher = {Association for Computing Machinery},
address = {New York, NY, USA},
url = {https://doi.org/10.1145/3626772.3657878},
doi = {10.1145/3626772.3657878},
booktitle = {Proceedings of the 47th International ACM SIGIR Conference on Research and Development in Information Retrieval},
pages = {641–649},
numpages = {9},
location = {Washington DC, USA},
series = {SIGIR '24}
}

@inproceedings{hou2024bridging,
    title = "Bridging Language and Items for Retrieval and Recommendation: Benchmarking {LLM}s as Semantic Encoders",
    author = "Hou, Yupeng  and
      Li, Jiacheng  and
      Fu, Xiangjun  and
      He, Zhankui  and
      Yan, An  and
      Chen, Xiusi  and
      McAuley, Julian",
    editor = "Liakata, Maria  and
      Moreira, Viviane P.  and
      Zhang, Jiajun  and
      Jurgens, David",
    booktitle = "Proceedings of the 64th Annual Meeting of the {A}ssociation for {C}omputational {L}inguistics (Volume 1: Long Papers)",
    month = jul,
    year = "2026",
    address = "San Diego, California, United States",
    publisher = "Association for Computational Linguistics",
    url = "https://aclanthology.org/2026.acl-long.147/",
    doi = "10.18653/v1/2026.acl-long.147",
    pages = "3251--3265",
    ISBN = "979-8-89176-390-6"
}

@inproceedings{zhang2015character,
 author = {Zhang, Xiang and Zhao, Junbo and LeCun, Yann},
 booktitle = {Advances in Neural Information Processing Systems},
 editor = {C. Cortes and N. Lawrence and D. Lee and M. Sugiyama and R. Garnett},
 pages = {},
 publisher = {Curran Associates, Inc.},
 title = {Character-level Convolutional Networks for Text Classification},
 url = {https://proceedings.neurips.cc/paper_files/paper/2015/file/250cf8b51c773f3f8dc8b4be867a9a02-Paper.pdf},
 volume = {28},
 year = {2015}
}

@inproceedings{demszky2020goemotions,
    title = "{G}o{E}motions: A Dataset of Fine-Grained Emotions",
    author = "Demszky, Dorottya  and
      Movshovitz-Attias, Dana  and
      Ko, Jeongwoo  and
      Cowen, Alan  and
      Nemade, Gaurav  and
      Ravi, Sujith",
    editor = "Jurafsky, Dan  and
      Chai, Joyce  and
      Schluter, Natalie  and
      Tetreault, Joel",
    booktitle = "Proceedings of the 58th Annual Meeting of the {A}ssociation for {C}omputational {L}inguistics",
    month = jul,
    year = "2020",
    address = "Online",
    publisher = "Association for Computational Linguistics",
    url = "https://aclanthology.org/2020.acl-main.372/",
    doi = "10.18653/v1/2020.acl-main.372",
    pages = "4040--4054"
}

@article{benjamini1995controlling,
    author = {Benjamini, Yoav and Hochberg, Yosef},
    title = {Controlling the False Discovery Rate: A Practical and Powerful Approach to Multiple Testing},
    journal = {Journal of the Royal Statistical Society: Series B (Methodological)},
    volume = {57},
    number = {1},
    pages = {289-300},
    year = {1995},
    month = {01},
    issn = {0035-9246},
    doi = {10.1111/j.2517-6161.1995.tb02031.x},
    url = {https://doi.org/10.1111/j.2517-6161.1995.tb02031.x},
    eprint = {https://academic.oup.com/jrsssb/article-pdf/57/1/289/49173396/jrsssb_57_1_289.pdf},
}

@inproceedings{biderman2023pythia,
  title={Pythia: A suite for analyzing large language models across training and scaling},
  author={Biderman, Stella and Schoelkopf, Hailey and Anthony, Quentin Gregory and Bradley, Herbie and O’Brien, Kyle and Hallahan, Eric and Khan, Mohammad Aflah and Purohit, Shivanshu and Prashanth, USVSN Sai and Raff, Edward and others},
  booktitle={International conference on machine learning},
  pages={2397--2430},
  year={2023},
  organization={PMLR}
}
}

% (the main paper refers to sections/figures/tables in the supplementary, so it is
%  compiled together here to resolve those references; split the PDF before submitting)
\clearpage
\setcounter{page}{1}
\maketitlesupplementary
\appendix
\supptableofcontents

\FloatBarrier

\section*{LLM Usage Statement}
% unnumbered table-of-contents entry: the appendix letters (A, B, ...) stay unchanged
\phantomsection\addcontentsline{toc}{section}{LLM Usage Statement}
    We used Claude Code (Anthropic) to assist with code development and implementation. We also used ChatGPT (OpenAI) for proofreading and correcting typographical and grammatical errors in the manuscript. All research decisions, experimental results, and scientific conclusions were made and verified by the authors.

\section{Experimental details}
\label{app:exp}

% Our code is available at
% \begin{center}
% \ifarxiv % official repository (arXiv version); anonymous mirror (review version), see the switch in main.tex
% \href{https://github.com/AtlasAnalyticsLab/DualSAE}{\nolinkurl{github.com/AtlasAnalyticsLab/DualSAE}}
% \else
% \href{https://anonymous.4open.science/r/DualSAE-F4B2}{\nolinkurl{anonymous.4open.science/r/DualSAE-F4B2}}
% \fi
% \end{center}

\FloatBarrier
\subsection{Hyper-parameters}
\label{app:hyper-param}

    Unless otherwise specified, all SAE variants are trained on frozen image embeddings for 50 epochs with batch size 2048, learning rate \(10^{-4}\), Adam \(\epsilon=6.25\times 10^{-10}\), and random seeds 42, 43, and 44. We use 8192 latent dimensions for all models. For TopK-based models, we use \(k=64\). Training is performed on a CUDA device with 4 dataloader workers. We use ImageNet-1K and OpenImagesV7 as datasets, and train separate models on embeddings extracted from frozen vision encoders, including CLIP, ViT, and SigLIP2.

For the standard TopK SAE, the encoder uses TopK activation with \(k=64\), tied encoder-decoder weights, and an auxiliary TopK loss with coefficient \(1/32\). The auxiliary TopK parameter is set to the same value as the main TopK parameter when not otherwise specified. The TopK SAE uses input normalization in our training configuration.

For BatchTopK SAE, we use the same number of latents and \(k=64\). The model is trained with the same optimizer settings as TopK SAE and uses gradient clipping with maximum norm 1.0. The auxiliary TopK loss coefficient is also set to \(1/32\). For JumpReLU SAE, we use the same 8192 latent dimensions and optimizer settings. JumpReLU SAE uses the same sparsity coefficient and a bandwidth of \(10^{-3}\) for the JumpReLU threshold smoothing.

DualSAE uses the same sparse latent dimensionality and the same sparse activation setting as the corresponding baseline SAE. In the main experiments, the Sparse Branch uses the selected sparsity mode, such as TopK, BatchTopK, or JumpReLU, while the Residual Branch models the residual component. The Residual Branch is parameterized as a one-layer projector. We use a decorrelation coefficient of 0.01.

\begin{table}[t]
\centering
\small
\setlength{\tabcolsep}{4pt}
\renewcommand{\arraystretch}{1.05}
\caption{\textbf{Common training hyperparameters.}
Unless otherwise specified, these settings are shared across SAE and DualSAE variants.
TK = TopK.}
\label{tab:training_hyperparams_common}
\begin{tabular}{lc}
\toprule
Hyperparameter & Value \\
\midrule
Number of latents & 8192 \\
Training epochs & 50 \\
Batch size & 2048 \\
Learning rate & \(10^{-4}\) \\
Adam \(\epsilon\) & \(6.25\times 10^{-10}\) \\
Random seeds & 42, 43, 44 \\
Number of workers & 4 \\
Device & 1 NVIDIA H100 \\
TK \(k\) & 64 \\
Maximum gradient norm & 1.0 \\
Auxiliary TopK coefficient & \(1/32\) \\
Dead neuron threshold & \(10^7\) steps \\
Dead neuron check interval & 100 steps \\
\bottomrule
\end{tabular}
\end{table}

\begin{table}[t]
\centering
\small
\setlength{\tabcolsep}{4pt}
\renewcommand{\arraystretch}{1.08}
\caption{\textbf{Model-specific hyperparameters.}
DualSAE uses the same Sparse Branch configuration as the corresponding SAE baseline, with an additional Residual Branch and decorrelation loss.
TK = TopK, BTK = BatchTopK, JR = JumpReLU, MAT = MatryoshkaSAE, GATE = GatedSAE, CB = CB-SAE.}
\label{tab:training_hyperparams_models}
\begin{tabularx}{\linewidth}{lX}
\toprule
Model & Hyperparameters \\
\midrule
TK &
TopK activation, \(k=64\), tied weights, normalization enabled, AuxK coefficient \(1/32\). \\

BTK &
BatchTopK activation, \(k=64\), AuxK coefficient \(1/32\), maximum gradient norm 1.0. \\

JR &
Sparsity coefficient \(10^{-4}\), bandwidth \(10^{-3}\), maximum gradient norm 1.0. \\

MAT &
BatchTopK activation, group fractions of 1/32, 1/16, 1/8, 1/4, and 17/32. \\

GATE &
Learning rate of 5e--5, and no grad-norm clip. \\

CB &
Concept set of VLG--CBM ImageNet-filtered, retain of 1024 latents, $k_{cb}$ of 64, and CB epochs/batch/lr: 25/2048/1e--4. \\

DualSAE &
Same sparse mode as the corresponding SAE baseline. For TopK variants, \(k=64\). The Residual Branch is a single linear layer \(\mathrm{Linear}(d,d)\) followed by GELU as the nonlinear operator (no hidden layer). We use a decorrelation coefficient of 0.01 and warm-up steps of 2500.\\
\bottomrule
\end{tabularx}
\end{table}

\FloatBarrier
\subsection{Datasets}
\label{app:datasets}

We evaluate DualSAE on two large-scale image datasets: ImageNet-1K~\citep{imagenet15russakovsky} and OpenImagesV7~\citep{OpenImages}. These datasets provide complementary evaluation settings. ImageNet-1K is a single-label classification benchmark with 1,000 object categories, making it suitable for measuring class-level semantic coherence, retrieval purity, and downstream classification accuracy. OpenImagesV7 is a large-scale multi-label dataset, where each image may contain multiple object categories. This provides a more challenging setting for evaluating whether SAE and DualSAE representations preserve semantic information when images contain multiple concepts, objects, and visual contexts.

For both datasets, we first extract frozen image embeddings using pretrained vision encoders, including ViT, CLIP, and SigLIP2. The SAE and DualSAE models are then trained directly on these fixed embeddings rather than on raw pixels. This setup isolates the representation-decomposition problem from encoder training and allows us to compare how different SAE variants reconstruct and decompose the same pretrained embedding space.

On ImageNet-1K, we use the standard class labels to compute class purity, KNN accuracy, and linear probing accuracy. Since each image has a single ground-truth class, KNN and LP are reported as top-1 accuracy. On OpenImagesV7, we adapt the evaluation to the multi-label setting. Class purity is computed based on the maximum label support among retrieved or activated samples, and KNN/LP performance is reported using micro-F1 with a threshold of \(0.5\). Together, these two datasets allow us to evaluate DualSAE under both clean single-label semantics and more realistic multi-label visual scenes.

\FloatBarrier
\subsection{Vision encoders}
\label{app:vision_encoders}

We study SAE and DualSAE decompositions on frozen image embeddings extracted from pretrained vision encoders. In our main experiments, we use ViT~\citep{vit}, CLIP~\citep{clip}, and SigLIP2~\citep{SigLIP2}. These encoders provide different representation spaces and allow us to test whether the proposed decomposition is robust across model families. ViT provides a standard supervised vision representation, while CLIP and SigLIP2 provide vision-language-aligned embeddings that often contain richer semantic and contextual information.

For each image \(I\), a frozen encoder \(f\) produces an embedding \(x=f(I)\in\mathbb{R}^d\). We train all SAE variants directly on these fixed embeddings rather than updating the vision encoder. This isolates the representation decomposition problem from encoder training and ensures that differences among SAE variants are due to the learned sparse and residual decompositions rather than changes in the backbone model.

\FloatBarrier
\subsection{Quantitative metrics}
\label{app:metrics}

Let $\mathcal{D}=\{(x_i,y_i)\}_{i=1}^N$ denote the evaluation set, where $x_i \in \mathbb{R}^d$ is the input embedding and $y_i$ is its label. For single-label datasets, $y_i \in \{1,\dots,C\}$; for multi-label datasets, $y_i \in \{0,1\}^C$ is a multi-hot vector. Given an autoencoder, we obtain latent activations $z_i \in \mathbb{R}^m$ and reconstruction $\hat{x}_i \in \mathbb{R}^d$. For latent $j$, we define its activating set as
\[
A_j = \{ i \in \{1,\dots,N\} : z_{ij} > 0 \},
\]
and denote the set of active latents by
\[
\mathcal{J}_{\mathrm{act}} = \{ j \in \{1,\dots,m\} : |A_j| > 0 \}.
\]

\textbf{Reconstruction $R^2$.}
We measure reconstruction fidelity using the coefficient of determination,
\[
R^2
=
1 - \frac{\sum_{i=1}^N \|x_i-\hat{x}_i\|_2^2}
{\sum_{i=1}^N \|x_i-\bar{x}\|_2^2},
\qquad
\bar{x}=\frac{1}{N}\sum_{i=1}^N x_i.
\]
Higher $R^2$ indicates better reconstruction quality.

\textbf{MonoScore (MS).}
MonoScore~\citep{monoscore} measures whether each latent consistently activates on semantically
similar images. Let
$\mathcal{D}=\{x_n\}_{n=1}^{N}$ be the evaluation set, where $x_n=f(I_n)$ is the embedding of image $I_n$ produced by the frozen encoder $f$ whose features the SAE is trained on (CLIP, SigLIP2, or ViT), and let
$a_{jn}$ denote the activation of latent $j$ on $x_n$.
We compute the semantic similarity between two images in the embedding space of the same encoder $f$:
\[
    h_n
    =
    \frac{x_n}{\lVert x_n\rVert_2},
    \qquad
    s_{nm}
    =
    h_n^\top h_m,
\]
so that $s_{nm}$ is the cosine similarity between the embeddings $x_n$ and $x_m$.
This differs from the original definition~\citep{monoscore}, which computes $s_{nm}$ with a separate image encoder; we discuss this choice at the end of this paragraph.

To make activation magnitudes comparable across latents, we independently
min--max normalize the activations of each latent:
\[
    \widetilde{a}_{jn}
    =
    \frac{
        a_{jn}-\min_{n'}a_{jn'}
    }{
        \max_{n'}a_{jn'}-\min_{n'}a_{jn'}
    }
    \quad
    \text{if }
    \max_{n'}a_{jn'}>\min_{n'}a_{jn'},
\]
and $\widetilde{a}_{jn}=0$ otherwise.

The MonoScore of latent $j$ is the activation-weighted average semantic
similarity over all unordered image pairs:
\[
    MS_j
    =
    \frac{
        \displaystyle
        \sum_{1\leq n<m\leq N}
        \widetilde{a}_{jn}\widetilde{a}_{jm}
        h_n^\top h_m
    }{
        \displaystyle
        \sum_{1\leq n<m\leq N}
        \widetilde{a}_{jn}\widetilde{a}_{jm}
    }.
    \label{eq:latent_monoscore}
\]
Here,
$\widetilde{a}_{jn}\widetilde{a}_{jm}$ is the co-activation weight of
the pair $(x_n,x_m)$, assigning greater importance to pairs that strongly
activate latent $j$.

We evaluate only latents for which the denominator is positive:
\[
    \mathcal{J}_{\mathrm{MS}}
    =
    \left\{
        j \;\middle|\;
        \sum_{1\leq n<m\leq N}
        \widetilde{a}_{jn}\widetilde{a}_{jm}>0
    \right\}.
\]
The model-level MonoScore is then the mean across these valid latents:
\[
    MS
    =
    \frac{1}{\lvert\mathcal{J}_{\mathrm{MS}}\rvert}
    \sum_{j\in\mathcal{J}_{\mathrm{MS}}} MS_j.
    \label{eq:model_monoscore}
\]
A larger $MS_j$ indicates that images strongly activating latent $j$ are
more semantically similar; consequently, a larger model-level $MS$ indicates
more monosemantic active latents on average.

\textbf{Choice of the similarity encoder in MonoScore.}
The original MonoScore~\citep{monoscore} computes the image similarities $s_{nm}$ with an encoder that differs from the one being explained, \eg, DINOv2 ViT-B for SAEs trained on CLIP ViT-L.
We instead compute $s_{nm}$ with the encoder $f$ whose embeddings the SAE decomposes, for two reasons.
First, an SAE decomposes the embedding space of $f$, so we measure whether each latent groups inputs that are coherent in that space.
Second, the score does not depend on a second model: \citet{monoscore} report that absolute MonoScore values change substantially with the choice of the external encoder.
Consequently, our absolute MonoScore values are not comparable to those reported in~\citep{monoscore}.
All comparisons in this paper, however, are between models evaluated with the same encoder and the same protocol: each standard SAE and its DualSAE counterpart are trained on the same embeddings and scored identically.

Computing similarities in the embedding space of $f$ can favor latents that group inputs that are close in that space, independently of their semantics.
We therefore also report class purity (below), which uses only the ground-truth labels and no encoder.
DualSAE achieves higher class purity than the corresponding standard SAE in all 36 settings of Table~\ref{tab:sae_comparison_imagenet_openimages_sidebyside} (six SAE variants, three encoders, and two datasets), and its MonoScore is higher in 34 of them.
DualSAE also achieves higher class purity in all 12 settings of the joint vs.\ sequential training comparison (Table~\ref{tab:frozen_vs_dualsae}) and in all 36 pairs of the expansion-ratio ablation (Supplementary~\ref{app:expansion}).
The interpretability improvements of DualSAE are therefore supported by an encoder-free metric as well as by MonoScore.

\textbf{Class purity.}
Class purity measures whether a latent is associated with a dominant class among the samples that activate it.

For a single-label dataset, the purity of latent $j$ is
\[
\mathrm{Purity}_j
=
\max_{c \in \{1,\dots,C\}}
\frac{1}{|A_j|}
\sum_{i \in A_j} \mathbb{I}[y_i=c].
\]

For a multi-label dataset with multi-hot labels $y_i \in \{0,1\}^C$, we define the class support of latent $j$ for class $c$ as
\[
\mathrm{Supp}_j(c)
=
\frac{1}{|A_j|}
\sum_{i \in A_j} y_{i,c},
\]
and define purity as the maximum class support,
\[
\mathrm{Purity}_j
=
\max_{c \in \{1,\dots,C\}} \mathrm{Supp}_j(c).
\]
This matches the implementation in our code: purity is the largest fraction of activating samples that share any one binary label. In both cases, the reported class purity is the mean over active latents,
\[
\mathrm{Purity}
=
\frac{1}{|\mathcal{J}_{\mathrm{act}}|}
\sum_{j \in \mathcal{J}_{\mathrm{act}}} \mathrm{Purity}_j.
\]
Note that this latent-level class purity differs from the retrieval-based class purity of the motivation experiment (Figure~\ref{fig:motivation_exp} of the main paper). There, class purity is the fraction of a query image's 50 nearest neighbors, retrieved by cosine similarity of $\hat{x}$ (or $r$), that belong to the query's class; it evaluates a representation rather than individual latents.

\textbf{KNN accuracy from reconstructed embeddings.}
To evaluate whether semantic label information is preserved after reconstruction, we build a training bank of reconstructed embeddings
\[
\mathcal{B}
=
\{(\hat{x}^{\mathrm{tr}}_i, y^{\mathrm{tr}}_i)\}_{i=1}^{N_{\mathrm{tr}}},
\]
where each $\hat{x}^{\mathrm{tr}}_i$ is L2-normalized. For an evaluation example, we reconstruct and normalize its embedding,
\[
q = \frac{\hat{x}}{\|\hat{x}\|_2},
\]
retrieve its $k$ nearest neighbors in the bank by cosine similarity, and denote their labels by $\{y_{n_1},\dots,y_{n_k}\}$.

For single-label datasets, we use majority voting:
\[
\hat{y}
=
\arg\max_{c \in \{1,\dots,C\}}
\sum_{t=1}^k \mathbb{I}[y_{n_t}=c],
\]
and report top-1 accuracy,
\[
\mathrm{KNN}
=
\frac{1}{N}
\sum_{i=1}^N \mathbb{I}[\hat{y}_i = y_i].
\]

For multi-label datasets, we average the neighbors' multi-hot labels,
\[
s = \frac{1}{k}\sum_{t=1}^k y_{n_t} \in [0,1]^C,
\qquad
\hat{y}_c = \mathbb{I}[s_c \ge 0.5].
\]
We then report micro-F1 over all sample-class pairs,
\[
\mathrm{KNN}
=
\mathrm{MicroF1}(\hat{Y}, Y),
\]
where
\[
\mathrm{MicroF1}
=
\frac{2\,TP}{2\,TP + FP + FN}.
\]
Here $TP$, $FP$, and $FN$ are accumulated over all classes and all evaluation examples.

\textbf{Linear probing accuracy from reconstructed embeddings.}
We also train a linear classifier on reconstructed training embeddings. Let
\[
\hat{X}^{\mathrm{tr}} = \{\hat{x}^{\mathrm{tr}}_i\}_{i=1}^{N_{\mathrm{tr}}},
\qquad
\hat{X}^{\mathrm{ev}} = \{\hat{x}^{\mathrm{ev}}_i\}_{i=1}^{N_{\mathrm{ev}}}.
\]
For single-label datasets, we train a linear classifier
\[
f(\hat{x}) = W\hat{x} + b \in \mathbb{R}^C
\]
using cross-entropy loss,
\[
\mathcal{L}_{\mathrm{CE}}
=
-\frac{1}{N_{\mathrm{tr}}}
\sum_{i=1}^{N_{\mathrm{tr}}}
\log
\frac{\exp(f(\hat{x}^{\mathrm{tr}}_i)_{y_i})}
{\sum_{c=1}^C \exp(f(\hat{x}^{\mathrm{tr}}_i)_c)}.
\]
Predictions are obtained by
\[
\hat{y} = \arg\max_c f(\hat{x})_c,
\]
and LP is reported as top-1 accuracy,
\[
\mathrm{LP}
=
\frac{1}{N_{\mathrm{ev}}}
\sum_{i=1}^{N_{\mathrm{ev}}}
\mathbb{I}[\hat{y}_i = y_i].
\]

For multi-label datasets, we use a linear layer with $C$ outputs and optimize binary cross-entropy with logits,
\[
\begin{aligned}
\mathcal{L}_{\mathrm{BCE}}
=
-\frac{1}{N_{\mathrm{tr}}C}
\sum_{i=1}^{N_{\mathrm{tr}}}
\sum_{c=1}^C
\Big(
&y_{i,c}\log \sigma(f(\hat{x}^{\mathrm{tr}}_i)_c) \\
&+
(1-y_{i,c})\log(1-\sigma(f(\hat{x}^{\mathrm{tr}}_i)_c))
\Big),
\end{aligned}
\]
where $\sigma(\cdot)$ is the sigmoid function. At evaluation time, predictions are thresholded at $0.5$,
\[
\hat{y}_{i,c}
=
\mathbb{I}[\sigma(f(\hat{x}^{\mathrm{ev}}_i)_c) \ge 0.5],
\]
and LP is reported as micro-F1,
\[
\mathrm{LP}
=
\mathrm{MicroF1}(\hat{Y}, Y).
\]

\textbf{Clarification about KNN and LP on $\hat{x}$ instead of the sparse code $z$.} Our goal is to evaluate reconstruction quality by measuring how well the reconstructed embedding $\hat{x}$ retains the semantic information of the original embedding. The sparse code $z$ instead indicates which monosemantic dictionary atoms are active in an input and is expressed in a different feature space. Therefore, directly using $z$ to evaluate semantic preservation would be misleading. MonoScore and Purity are two more reasonable metrics for evaluating the sparse code $z$.

\FloatBarrier
\section{Additional results in the vision domain}

    \FloatBarrier
    \subsection{Detailed results on ImageNet-1K and OpenImagesV7}
        \label{app:detail_results}

\definecolor{dualrow}{RGB}{235,242,255} % light blue

\begin{table*}[p]
    \centering
    \scriptsize
    \setlength{\tabcolsep}{2pt}
    \renewcommand{\arraystretch}{1.55}
    \caption{
    Comparison of SAE variants across vision encoders on ImageNet-1K (IN) and OpenImagesV7 (OI).
    DualSAE generally improves reconstruction ($R^2$), interpretability (MS, Pur.), and KNN over its standard counterpart across encoders, activations, and both datasets, with comparable or better LP. SL2 = SigLIP2. SAE abbreviations: TK = TopK, BTK = BatchTopK, JR = JumpReLU,
    MAT = MatryoshkaSAE, GATE = GatedSAE, and CB = CB-SAE.
    Prefix D- denotes the corresponding DualSAE variant; CB-DUAL denotes CB-SAE
    augmented with the DualSAE Residual Branch.
    MS denotes MonoScore; Pur. denotes class purity; KNN and LP denote
    KNN- and linear-probe-based downstream performance, respectively.
    For IN, both are reported as top-1 accuracy; for OI,
    both are reported as micro-F1.
    }
    \label{tab:sae_comparison_imagenet_openimages_sidebyside}

    \resizebox*{!}{0.89\textheight}{%
    \begin{tabular}{
        >{\centering\arraybackslash}m{0.035\textwidth}
        l
        ccccc
        @{\hspace{8pt}}
        ccccc
    }
    \toprule
    & &
    \multicolumn{5}{c}{\textbf{IN}} &
    \multicolumn{5}{c}{\textbf{OI}} \\
    \cmidrule(lr){3-7}
    \cmidrule(lr){8-12}

    \textbf{Enc.} &
    \textbf{SAE} &
    \textbf{\boldmath$R^2$} &
    \textbf{MS} &
    \textbf{Pur.} &
    \textbf{KNN} &
    \textbf{LP} &
    \textbf{\boldmath$R^2$} &
    \textbf{MS} &
    \textbf{Pur.} &
    \textbf{KNN} &
    \textbf{LP} \\

    \midrule

    \multirow{12}{=}{%
        \centering
        \rotatebox[origin=c]{90}{\textbf{ViT}}
    }
    & TK
    & \stdval{0.8081}{0.0008}
    & \stdval{0.2375}{0.0011}
    & \stdval{0.1271}{0.0007}
    & \stdval{0.8203}{0.0011}
    & \stdval{0.8230}{0.0006}
    & \stdval{0.7037}{0.0011}
    & \stdval{0.1283}{0.0007}
    & \stdval{0.2259}{0.0003}
    & \stdval{0.4675}{0.0011}
    & \stdval{0.4022}{0.0008} \\

    & \dc{D-TK}
    & \dc{\stdbf{0.9661}{0.0003}}
    & \dc{\stdbf{0.4874}{0.0008}}
    & \dc{\stdbf{0.5666}{0.0002}}
    & \dc{\stdbf{0.8226}{0.0008}}
    & \dc{\stdbf{0.8239}{0.0004}}
    & \dc{\stdbf{0.9611}{0.0006}}
    & \dc{\stdbf{0.2446}{0.0008}}
    & \dc{\stdbf{0.5028}{0.0001}}
    & \dc{\stdbf{0.4701}{0.0010}}
    & \dc{\stdbf{0.4049}{0.0005}} \\

    \cmidrule(lr){2-12}

    & BTK
    & \stdval{0.7879}{0.0007}
    & \stdval{0.2450}{0.0009}
    & \stdval{0.1513}{0.0004}
    & \stdval{0.8207}{0.0005}
    & \stdval{0.8230}{0.0005}
    & \stdval{0.6782}{0.0012}
    & \stdval{0.1369}{0.0008}
    & \stdval{0.2569}{0.0004}
    & \stdval{0.4672}{0.0013}
    & \stdbf{0.4050}{0.0009} \\

    & \dc{D-BTK}
    & \dc{\stdbf{0.9725}{0.0004}}
    & \dc{\stdbf{0.4775}{0.0005}}
    & \dc{\stdbf{0.5481}{0.0002}}
    & \dc{\stdbf{0.8228}{0.0003}}
    & \dc{\stdbf{0.8242}{0.0005}}
    & \dc{\stdbf{0.9593}{0.0006}}
    & \dc{\stdbf{0.2708}{0.0003}}
    & \dc{\stdbf{0.5284}{0.0002}}
    & \dc{\stdbf{0.4700}{0.0006}}
    & \dc{\stdval{0.4035}{0.0003}} \\

    \cmidrule(lr){2-12}

    & JR
    & \stdval{0.8020}{0.0007}
    & \stdval{0.2840}{0.0008}
    & \stdval{0.2391}{0.0006}
    & \stdval{0.8201}{0.0008}
    & \stdval{0.8221}{0.0007}
    & \stdval{0.7238}{0.0008}
    & \stdbf{0.1075}{0.0009}
    & \stdval{0.2273}{0.0004}
    & \stdval{0.4675}{0.0006}
    & \stdbf{0.4041}{0.0007} \\

    & \dc{D-JR}
    & \dc{\stdbf{0.9843}{0.0003}}
    & \dc{\stdbf{0.3576}{0.0007}}
    & \dc{\stdbf{0.4240}{0.0002}}
    & \dc{\stdbf{0.8239}{0.0006}}
    & \dc{\stdbf{0.8239}{0.0006}}
    & \dc{\stdbf{0.9701}{0.0006}}
    & \dc{\stdval{0.0940}{0.0003}}
    & \dc{\stdbf{0.2863}{0.0003}}
    & \dc{\stdbf{0.4703}{0.0003}}
    & \dc{\stdval{0.4030}{0.0003}} \\

    \cmidrule(lr){2-12}

    & MAT
    & \stdval{0.8273}{0.0002}
    & \stdval{0.2733}{0.0017}
    & \stdval{0.2233}{0.0018}
    & \stdval{0.8213}{0.0002}
    & \stdval{0.8233}{0.0001}
    & \stdval{0.7316}{0.0003}
    & \stdval{0.1288}{0.0004}
    & \stdval{0.3023}{0.0010}
    & \stdval{0.4670}{0.0003}
    & \stdval{0.4041}{0.0003} \\

    & \dc{D-MAT}
    & \dc{\stdbf{0.9612}{0.0008}}
    & \dc{\stdbf{0.5201}{0.0011}}
    & \dc{\stdbf{0.6305}{0.0016}}
    & \dc{\stdbf{0.8242}{0.0001}}
    & \dc{\stdbf{0.8245}{0.0003}}
    & \dc{\stdbf{0.9025}{0.0008}}
    & \dc{\stdbf{0.3596}{0.0016}}
    & \dc{\stdbf{0.6373}{0.0042}}
    & \dc{\stdbf{0.4705}{0.0002}}
    & \dc{\stdbf{0.4052}{0.0003}} \\

    \cmidrule(lr){2-12}

    & GATE
    & \stdval{0.8225}{0.0007}
    & \stdval{0.2371}{0.0005}
    & \stdval{0.1462}{0.0005}
    & \stdval{0.8208}{0.0005}
    & \stdval{0.8220}{0.0008}
    & \stdval{0.6908}{0.0014}
    & \stdval{0.1421}{0.0016}
    & \stdval{0.3400}{0.0028}
    & \stdval{0.4665}{0.0002}
    & \stdbf{0.4060}{0.0001} \\

    & \dc{D-GATE}
    & \dc{\stdbf{0.9268}{0.0075}}
    & \dc{\stdbf{0.3657}{0.0032}}
    & \dc{\stdbf{0.3044}{0.0036}}
    & \dc{\stdbf{0.8220}{0.0005}}
    & \dc{\stdbf{0.8236}{0.0002}}
    & \dc{\stdbf{0.7723}{0.0030}}
    & \dc{\stdbf{0.2214}{0.0007}}
    & \dc{\stdbf{0.4594}{0.0003}}
    & \dc{\stdbf{0.4682}{0.0001}}
    & \dc{\stdval{0.4059}{0.0004}} \\

    \cmidrule(lr){2-12}

    & CB
    & \stdval{0.6550}{0.0027}
    & \stdval{0.2809}{0.0009}
    & \stdval{0.3640}{0.0066}
    & \stdval{0.7988}{0.0014}
    & \stdval{0.8078}{0.0019}
    & \stdval{0.5572}{0.0013}
    & \stdval{0.2564}{0.0009}
    & \stdval{0.5018}{0.0007}
    & \stdval{0.4655}{0.0006}
    & \stdbf{0.4091}{0.0011} \\

    & \dc{CB-DUAL}
    & \dc{\stdbf{0.9823}{0.0024}}
    & \dc{\stdbf{0.3511}{0.0023}}
    & \dc{\stdbf{0.5963}{0.0145}}
    & \dc{\stdbf{0.8133}{0.0004}}
    & \dc{\stdbf{0.8093}{0.0003}}
    & \dc{\stdbf{0.9418}{0.0031}}
    & \dc{\stdbf{0.2919}{0.0091}}
    & \dc{\stdbf{0.5820}{0.0612}}
    & \dc{\stdbf{0.4800}{0.0003}}
    & \dc{\stdval{0.4067}{0.0002}} \\

    \midrule

    \multirow{12}{=}{%
        \centering
        \rotatebox[origin=c]{90}{\textbf{CLIP}}
    }
    & TK
    & \stdval{0.8189}{0.0010}
    & \stdval{0.4680}{0.0007}
    & \stdval{0.1572}{0.0003}
    & \stdval{0.8036}{0.0007}
    & \stdval{0.8326}{0.0005}
    & \stdval{0.7575}{0.0007}
    & \stdval{0.3798}{0.0005}
    & \stdval{0.2472}{0.0003}
    & \stdval{0.4682}{0.0007}
    & \stdval{0.3990}{0.0007} \\

    & \dc{D-TK}
    & \dc{\stdbf{0.9964}{0.0002}}
    & \dc{\stdbf{0.5603}{0.0007}}
    & \dc{\stdbf{0.4256}{0.0002}}
    & \dc{\stdbf{0.8155}{0.0002}}
    & \dc{\stdbf{0.8450}{0.0004}}
    & \dc{\stdbf{0.9858}{0.0006}}
    & \dc{\stdbf{0.3879}{0.0006}}
    & \dc{\stdbf{0.5800}{0.0002}}
    & \dc{\stdbf{0.4699}{0.0002}}
    & \dc{\stdbf{0.4077}{0.0005}} \\

    \cmidrule(lr){2-12}

    & BTK
    & \stdval{0.8472}{0.0009}
    & \stdval{0.4357}{0.0008}
    & \stdval{0.1410}{0.0006}
    & \stdval{0.8069}{0.0008}
    & \stdval{0.8347}{0.0009}
    & \stdval{0.7863}{0.0008}
    & \stdval{0.3607}{0.0006}
    & \stdval{0.2584}{0.0008}
    & \stdval{0.4673}{0.0008}
    & \stdval{0.4053}{0.0007} \\

    & \dc{D-BTK}
    & \dc{\stdbf{0.9972}{0.0004}}
    & \dc{\stdbf{0.5750}{0.0003}}
    & \dc{\stdbf{0.4287}{0.0004}}
    & \dc{\stdbf{0.8156}{0.0004}}
    & \dc{\stdbf{0.8451}{0.0003}}
    & \dc{\stdbf{0.9957}{0.0003}}
    & \dc{\stdbf{0.4510}{0.0003}}
    & \dc{\stdbf{0.6543}{0.0006}}
    & \dc{\stdbf{0.4700}{0.0005}}
    & \dc{\stdbf{0.4070}{0.0006}} \\

    \cmidrule(lr){2-12}

    & JR
    & \stdval{0.7792}{0.0005}
    & \stdval{0.4477}{0.0006}
    & \stdval{0.1920}{0.0003}
    & \stdval{0.8002}{0.0009}
    & \stdval{0.8272}{0.0009}
    & \stdval{0.7361}{0.0009}
    & \stdval{0.3557}{0.0009}
    & \stdval{0.2553}{0.0004}
    & \stdval{0.4658}{0.0010}
    & \stdval{0.4035}{0.0009} \\

    & \dc{D-JR}
    & \dc{\stdbf{0.9745}{0.0006}}
    & \dc{\stdbf{0.4947}{0.0005}}
    & \dc{\stdbf{0.2642}{0.0003}}
    & \dc{\stdbf{0.8124}{0.0006}}
    & \dc{\stdbf{0.8424}{0.0007}}
    & \dc{\stdbf{0.9901}{0.0002}}
    & \dc{\stdbf{0.3599}{0.0004}}
    & \dc{\stdbf{0.3443}{0.0004}}
    & \dc{\stdbf{0.4699}{0.0007}}
    & \dc{\stdbf{0.4068}{0.0008}} \\

    \cmidrule(lr){2-12}

    & MAT
    & \stdval{0.8351}{0.0003}
    & \stdval{0.4889}{0.0003}
    & \stdval{0.2763}{0.0017}
    & \stdval{0.8064}{0.0006}
    & \stdval{0.8359}{0.0003}
    & \stdval{0.7734}{0.0002}
    & \stdval{0.3825}{0.0008}
    & \stdval{0.3276}{0.0026}
    & \stdval{0.4679}{0.0002}
    & \stdval{0.4047}{0.0007} \\

    & \dc{D-MAT}
    & \dc{\stdbf{0.9998}{0.0002}}
    & \dc{\stdbf{0.5289}{0.0045}}
    & \dc{\stdbf{0.4721}{0.0077}}
    & \dc{\stdbf{0.8218}{0.0008}}
    & \dc{\stdbf{0.8488}{0.0002}}
    & \dc{\stdbf{0.9572}{0.0008}}
    & \dc{\stdbf{0.3939}{0.0087}}
    & \dc{\stdbf{0.6932}{0.0044}}
    & \dc{\stdbf{0.4712}{0.0001}}
    & \dc{\stdbf{0.4095}{0.0002}} \\

    \cmidrule(lr){2-12}

    & GATE
    & \stdval{0.8256}{0.0006}
    & \stdval{0.4013}{0.0005}
    & \stdval{0.0920}{0.0008}
    & \stdval{0.8015}{0.0009}
    & \stdval{0.8324}{0.0004}
    & \stdval{0.7264}{0.0005}
    & \stdval{0.3431}{0.0017}
    & \stdval{0.2891}{0.0021}
    & \stdval{0.4648}{0.0002}
    & \stdval{0.4015}{0.0008} \\

    & \dc{D-GATE}
    & \dc{\stdbf{0.9940}{0.0005}}
    & \dc{\stdbf{0.4608}{0.0026}}
    & \dc{\stdbf{0.2995}{0.0033}}
    & \dc{\stdbf{0.8159}{0.0002}}
    & \dc{\stdbf{0.8450}{0.0001}}
    & \dc{\stdbf{0.9246}{0.0043}}
    & \dc{\stdbf{0.3741}{0.0054}}
    & \dc{\stdbf{0.4466}{0.0049}}
    & \dc{\stdbf{0.4695}{0.0002}}
    & \dc{\stdbf{0.4088}{0.0003}} \\

    \cmidrule(lr){2-12}

    & CB
    & \stdval{0.5959}{0.0012}
    & \stdval{0.3679}{0.0137}
    & \stdval{0.4027}{0.0052}
    & \stdval{0.7709}{0.0035}
    & \stdval{0.7978}{0.0036}
    & \stdval{0.5653}{0.0066}
    & \stdval{0.4653}{0.0022}
    & \stdval{0.4857}{0.0041}
    & \stdval{0.4692}{0.0009}
    & \stdval{0.3972}{0.0013} \\

    & \dc{CB-DUAL}
    & \dc{\stdbf{0.9865}{0.0003}}
    & \dc{\stdbf{0.3989}{0.0088}}
    & \dc{\stdbf{0.6872}{0.0044}}
    & \dc{\stdbf{0.8641}{0.0001}}
    & \dc{\stdbf{0.8283}{0.0003}}
    & \dc{\stdbf{0.9715}{0.0009}}
    & \dc{\stdbf{0.4952}{0.0123}}
    & \dc{\stdbf{0.5981}{0.0842}}
    & \dc{\stdbf{0.4771}{0.0004}}
    & \dc{\stdbf{0.4125}{0.0002}} \\

    \midrule

    \multirow{12}{=}{%
        \centering
        \rotatebox[origin=c]{90}{\textbf{SL2}}
    }
    & TK
    & \stdval{0.6457}{0.0009}
    & \stdval{0.5903}{0.0010}
    & \stdval{0.3412}{0.0010}
    & \stdval{0.8311}{0.0010}
    & \stdval{0.8663}{0.0004}
    & \stdval{0.5846}{0.0011}
    & \stdbf{0.5473}{0.0011}
    & \stdval{0.3758}{0.0009}
    & \stdval{0.4679}{0.0009}
    & \stdval{0.3644}{0.0008} \\

    & \dc{D-TK}
    & \dc{\stdbf{0.9944}{0.0005}}
    & \dc{\stdbf{0.6354}{0.0008}}
    & \dc{\stdbf{0.4036}{0.0008}}
    & \dc{\stdbf{0.8517}{0.0008}}
    & \dc{\stdbf{0.8830}{0.0005}}
    & \dc{\stdbf{0.9690}{0.0008}}
    & \dc{\stdval{0.5416}{0.0005}}
    & \dc{\stdbf{0.5944}{0.0003}}
    & \dc{\stdbf{0.4723}{0.0004}}
    & \dc{\stdbf{0.4104}{0.0008}} \\

    \cmidrule(lr){2-12}

    & BTK
    & \stdval{0.8462}{0.0008}
    & \stdval{0.4689}{0.0009}
    & \stdval{0.2270}{0.0009}
    & \stdval{0.8477}{0.0009}
    & \stdval{0.8763}{0.0008}
    & \stdval{0.7604}{0.0012}
    & \stdval{0.4473}{0.0013}
    & \stdval{0.3356}{0.0005}
    & \stdval{0.4705}{0.0008}
    & \stdval{0.4056}{0.0006} \\

    & \dc{D-BTK}
    & \dc{\stdbf{0.9942}{0.0003}}
    & \dc{\stdbf{0.6231}{0.0009}}
    & \dc{\stdbf{0.3958}{0.0009}}
    & \dc{\stdbf{0.8516}{0.0005}}
    & \dc{\stdbf{0.8830}{0.0004}}
    & \dc{\stdbf{0.9814}{0.0006}}
    & \dc{\stdbf{0.5635}{0.0010}}
    & \dc{\stdbf{0.6191}{0.0002}}
    & \dc{\stdbf{0.4720}{0.0005}}
    & \dc{\stdbf{0.4102}{0.0004}} \\

    \cmidrule(lr){2-12}

    & JR
    & \stdval{0.7853}{0.0006}
    & \stdval{0.5120}{0.0006}
    & \stdval{0.2631}{0.0006}
    & \stdval{0.8451}{0.0007}
    & \stdval{0.8732}{0.0007}
    & \stdval{0.7172}{0.0009}
    & \stdval{0.4673}{0.0009}
    & \stdval{0.2899}{0.0003}
    & \stdval{0.4702}{0.0007}
    & \stdval{0.4051}{0.0008} \\

    & \dc{D-JR}
    & \dc{\stdbf{0.9806}{0.0002}}
    & \dc{\stdbf{0.5770}{0.0003}}
    & \dc{\stdbf{0.3046}{0.0003}}
    & \dc{\stdbf{0.8504}{0.0002}}
    & \dc{\stdbf{0.8816}{0.0005}}
    & \dc{\stdbf{0.9583}{0.0005}}
    & \dc{\stdbf{0.4873}{0.0004}}
    & \dc{\stdbf{0.3421}{0.0002}}
    & \dc{\stdbf{0.4724}{0.0005}}
    & \dc{\stdbf{0.4095}{0.0003}} \\

    \cmidrule(lr){2-12}

    & MAT
    & \stdval{0.8362}{0.0000}
    & \stdval{0.5894}{0.0005}
    & \stdval{0.3261}{0.0012}
    & \stdval{0.8466}{0.0008}
    & \stdval{0.8768}{0.0005}
    & \stdval{0.7645}{0.0004}
    & \stdval{0.5261}{0.0003}
    & \stdval{0.3805}{0.0008}
    & \stdval{0.4712}{0.0002}
    & \stdval{0.4064}{0.0003} \\

    & \dc{D-MAT}
    & \dc{\stdbf{0.9083}{0.0003}}
    & \dc{\stdbf{0.5911}{0.0023}}
    & \dc{\stdbf{0.4819}{0.0043}}
    & \dc{\stdbf{0.8490}{0.0003}}
    & \dc{\stdbf{0.8812}{0.0003}}
    & \dc{\stdbf{0.8731}{0.0010}}
    & \dc{\stdbf{0.5581}{0.0019}}
    & \dc{\stdbf{0.4812}{0.0038}}
    & \dc{\stdbf{0.4999}{0.0002}}
    & \dc{\stdbf{0.4112}{0.0001}} \\

    \cmidrule(lr){2-12}

    & GATE
    & \stdval{0.8372}{0.0004}
    & \stdval{0.5236}{0.0016}
    & \stdval{0.1291}{0.0024}
    & \stdval{0.8452}{0.0007}
    & \stdval{0.8758}{0.0006}
    & \stdval{0.7107}{0.0015}
    & \stdval{0.4668}{0.0023}
    & \stdval{0.2902}{0.0022}
    & \stdval{0.4694}{0.0002}
    & \stdval{0.4027}{0.0009} \\

    & \dc{D-GATE}
    & \dc{\stdbf{0.8812}{0.0052}}
    & \dc{\stdbf{0.5512}{0.0048}}
    & \dc{\stdbf{0.1923}{0.0022}}
    & \dc{\stdbf{0.8512}{0.0011}}
    & \dc{\stdbf{0.8822}{0.0009}}
    & \dc{\stdbf{0.9110}{0.0002}}
    & \dc{\stdbf{0.4812}{0.0004}}
    & \dc{\stdbf{0.3011}{0.0002}}
    & \dc{\stdbf{0.4783}{0.0002}}
    & \dc{\stdbf{0.4110}{0.0002}} \\

    \cmidrule(lr){2-12}

    & CB
    & \stdval{0.5399}{0.0011}
    & \stdval{0.4693}{0.0139}
    & \stdval{0.4312}{0.0016}
    & \stdval{0.8108}{0.0036}
    & \stdval{0.8437}{0.0034}
    & \stdval{0.4865}{0.0028}
    & \stdval{0.5710}{0.0085}
    & \stdval{0.5088}{0.0080}
    & \stdval{0.4715}{0.0004}
    & \stdval{0.3996}{0.0031} \\

    & \dc{CB-DUAL}
    & \dc{\stdbf{0.9348}{0.0004}}
    & \dc{\stdbf{0.5291}{0.0102}}
    & \dc{\stdbf{0.5341}{0.0132}}
    & \dc{\stdbf{0.8444}{0.0003}}
    & \dc{\stdbf{0.8791}{0.0002}}
    & \dc{\stdbf{0.9712}{0.0009}}
    & \dc{\stdbf{0.5891}{0.0112}}
    & \dc{\stdbf{0.6012}{0.0065}}
    & \dc{\stdbf{0.4755}{0.0003}}
    & \dc{\stdbf{0.4202}{0.0005}} \\

    \bottomrule
    \end{tabular}%
    }
\end{table*}

        Table~\ref{tab:sae_comparison_imagenet_openimages_sidebyside} compares standard SAE variants with their DualSAE counterparts across three vision encoders, six activation mechanisms, and two datasets (with results averaged over three random seeds). The most consistent trend is the improvement in reconstruction quality. Across all encoder and activation settings, DualSAE substantially increases reconstruction \(R^2\) on both ImageNet-1K and OpenImagesV7. This indicates that the Residual Branch effectively recovers information that is difficult to preserve through a single sparse bottleneck.

        DualSAE also improves latent interpretability in most settings. On ImageNet-1K, DualSAE consistently increases MonoScore and class purity across nearly all SAE variants, showing that the Sparse Branch learns more semantically coherent and class-consistent latent features. The improvement in purity is especially large, suggesting that active latents in DualSAE tend to correspond to more concentrated visual concepts. On OpenImagesV7, the gains in purity remain strong across encoders and activation functions, even though the dataset is multi-label and therefore more semantically complex. MonoScore improvements are more mixed on OpenImagesV7, which is expected because images may contain multiple objects and overlapping concepts, making monosemanticity harder to measure with a single dominant visual concept.

        Downstream performance also benefits from the DualSAE decomposition. On ImageNet-1K, DualSAE improves both KNN and LP accuracy for nearly every encoder and activation setting, indicating that the reconstructed embeddings preserve more label-relevant information. On OpenImagesV7, KNN performance also improves consistently, while LP performance is generally competitive and often slightly better. The smaller LP gains on OpenImagesV7 suggest that multi-label classification is a harder and more saturated setting, but DualSAE still improves reconstruction and latent purity without substantially harming downstream utility.

        Overall, the table shows that DualSAE improves the reconstruction-interpretability trade-off across different encoders, activation mechanisms, and datasets. The consistent \(R^2\) gains demonstrate the benefit of modeling residual information separately, while the improvements in MonoScore, purity, and KNN/LP performance suggest that this additional capacity does not simply make the representation less interpretable. Instead, separating the sparse reconstruction from the residual component allows DualSAE to preserve semantic structure while reconstructing the original embedding more faithfully.

        \begin{figure*}[t]
            \centering
            \includegraphics[width=0.8\linewidth]{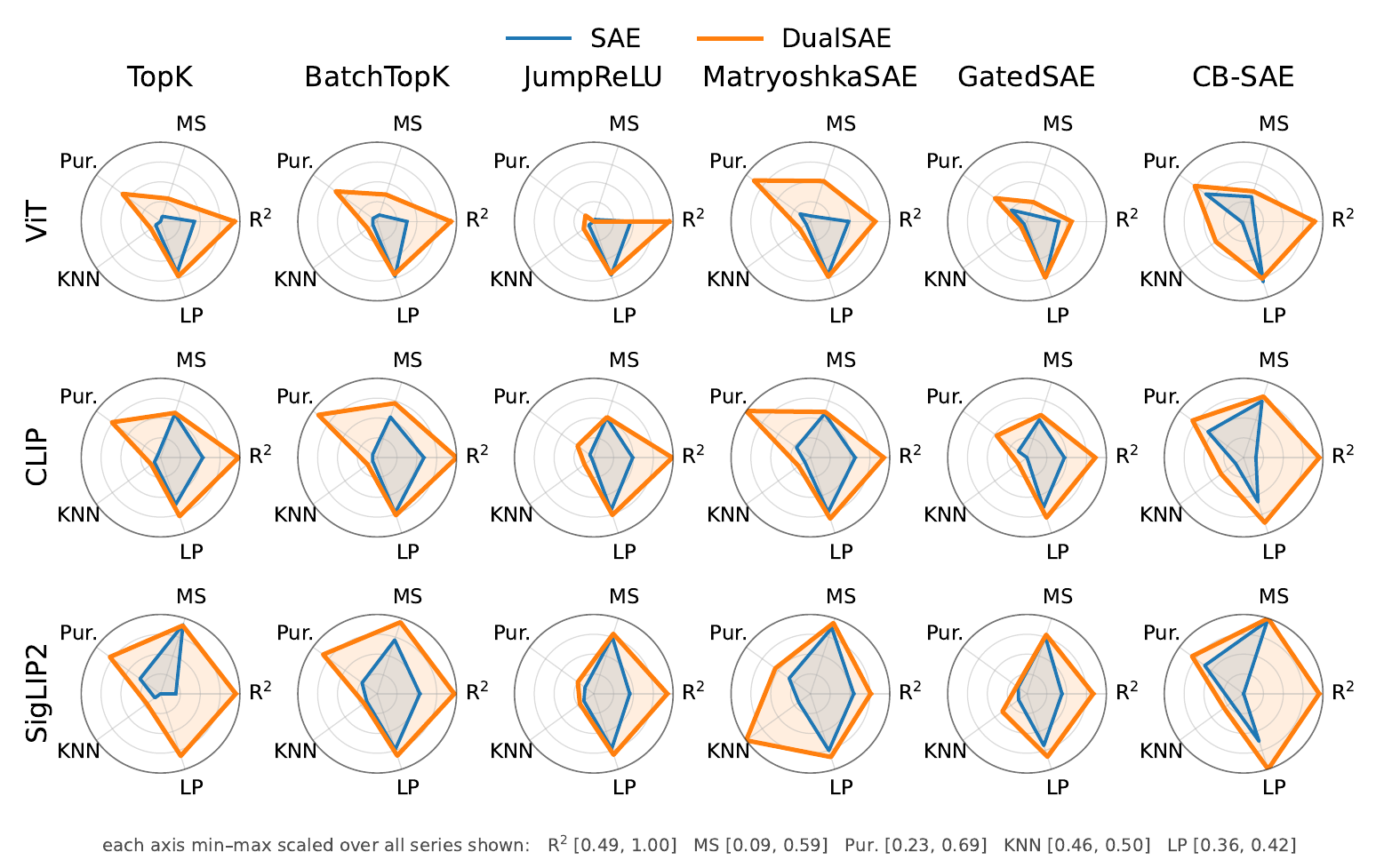}
            \caption{\textbf{Comparison of SAE and DualSAE on OpenImagesV7.}
    Radar plots compare each SAE variant with its DualSAE counterpart across three encoders and six SAE variants using reconstruction \(R^2\), MonoScore, class purity, KNN accuracy, and LP accuracy. Across 18 settings, DualSAE generally improves reconstruction fidelity, monosemanticity, and downstream performance.}
        \label{fig:radar_openimagesv7}
        \end{figure*}

    \subsection{Input--reconstruction alignment}
    \label{app:recon_alignment}
\begin{figure}[h]
        \centering
        \includegraphics[width=\linewidth]{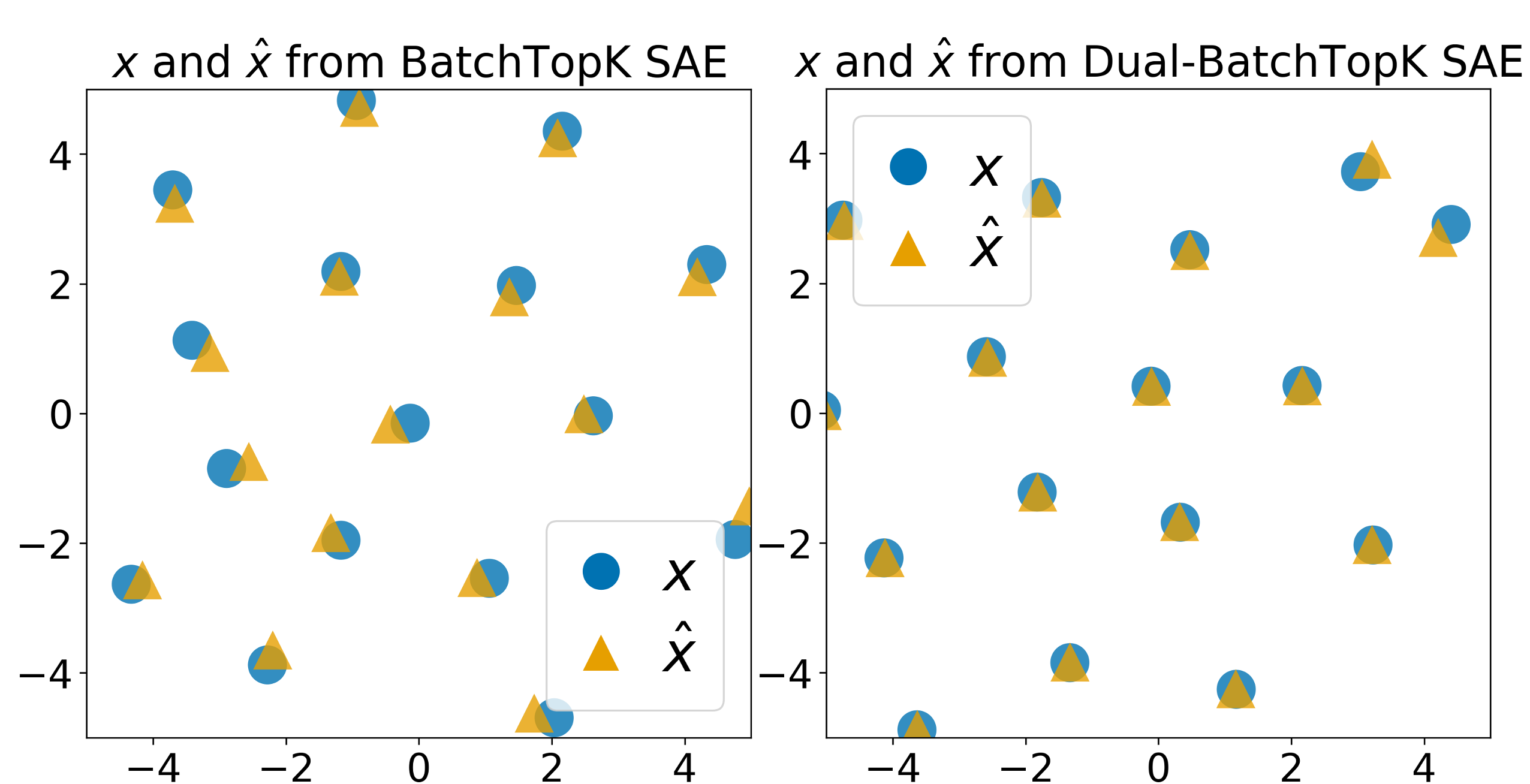}
        \caption{\textbf{Input--reconstruction alignment.} Dual-BatchTopK preserves the original embedding geometry better than BatchTopK, shown by the overlap between \textcolor{x_fig4}{input $x$} and \textcolor{xhat_fig4}{reconstructed $\hat{x}$}.}
        \label{fig:recon_alignment}
    \end{figure}

    We use t-SNE~\citep{tsne} to compare the geometry of the reconstructed input between BatchTopK and Dual-BatchTopK SAE. In Figure~\ref{fig:recon_alignment}, the standard SAE shows visible displacement between original embeddings \(x\) and reconstructed embeddings \(\hat{x}\), indicating weaker reconstruction fidelity. In contrast, Dual-BatchTopK produces much tighter overlap, which means higher reconstruction quality.

    \subsection{Class information in each branch on OpenImagesV7}
    \label{app:branch_oi}
\begin{figure}[h]
        \centering
        \includegraphics[width=\linewidth]{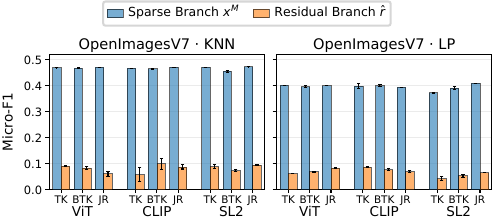}
        \caption{\textbf{Class information in each DualSAE branch on OpenImagesV7.} KNN and LP micro-F1 of the Sparse Branch output $x^M$ and the Residual Branch output $\hat{r}$ (mean and standard deviation over three seeds). $x^M$ far exceeds $\hat{r}$ in all 18 settings, so the class information resides in the Sparse Branch. SL2: SigLIP2; TK, BTK, JR: TopK, BatchTopK, JumpReLU.}
        \label{fig:xm_vs_rhat_oi}
    \end{figure}

    Figure~\ref{fig:xm_vs_rhat_oi} complements Figure~\ref{fig:branch_decomposition}b of the main paper with the KNN and LP results on OpenImagesV7, measured by micro-F1 with the same settings ($k=20$ cosine-similarity neighbors for KNN, 10 training epochs for LP, three seeds). The Sparse Branch output $x^M$ reaches 0.45--0.47 (KNN) and 0.37--0.41 (LP), whereas the Residual Branch output $\hat{r}$ reaches only 0.06--0.10 and 0.04--0.09, so $x^M$ outperforms $\hat{r}$ in all 18 comparisons on OpenImagesV7.

\FloatBarrier
\twocolumn[{%
% full-width block so the section and subsection headings and the two tables share a page
\section{Ablation studies}
\label{app:ablation}

\subsection{Coefficient of the decorrelation loss}
\label{app:gamma_ablation}

\begin{center}
\captionof{table}{Reconstruction performance ($R^2$) under different
values of $\gamma$. The best result for each configuration
is highlighted in bold.
Our default $\gamma=10^{-2}$ gives the highest $R^2$ in 4 of 12 configurations and is within 0.004 of the best in 5 others (largest gap 0.021, IN with ViT and TK); the lowest $R^2$ in every configuration is at $\gamma=1$ or $\gamma=10$. IN = ImageNet-1K, OI = OpenImagesV7, SL2 = SigLIP2, TK = TopK, BTK = BatchTopK.}
\label{tab:gamma_r2}
\resizebox{0.69\textwidth}{!}{% width scaled with the column count to keep the previous font size
\begin{tabular}{lllccccc}
\toprule
\textbf{Data} & \textbf{Enc.} & \textbf{SAE}
& $\boldsymbol{\gamma=10^{-3}}$
& $\boldsymbol{\gamma=10^{-2}}$
& $\boldsymbol{\gamma=10^{-1}}$
& $\boldsymbol{\gamma=1}$
& $\boldsymbol{\gamma=10}$ \\
\midrule

IN & CLIP & TK
& \stdbf{0.9979}{0.0001}
& \stdval{0.9964}{0.0002}
& \stdval{0.9978}{0.0000}
& \stdval{0.9965}{0.0003}
& \stdval{0.9733}{0.0105} \\

IN & CLIP & BTK
& \stdval{0.9956}{0.0002}
& \stdval{0.9972}{0.0004}
& \stdbf{0.9975}{0.0003}
& \stdval{0.9961}{0.0001}
& \stdval{0.9937}{0.0006} \\

IN & SL2 & TK
& \stdbf{0.9945}{0.0005}
& \stdval{0.9944}{0.0005}
& \stdval{0.9900}{0.0030}
& \stdval{0.9918}{0.0028}
& \stdval{0.9590}{0.0281} \\

IN & SL2 & BTK
& \stdval{0.9923}{0.0004}
& \stdbf{0.9942}{0.0003}
& \stdval{0.9894}{0.0004}
& \stdval{0.9709}{0.0003}
& \stdval{0.9283}{0.0119} \\

IN & ViT & TK
& \stdbf{0.9872}{0.0001}
& \stdval{0.9661}{0.0003}
& \stdval{0.9867}{0.0002}
& \stdval{0.9824}{0.0010}
& \stdval{0.9555}{0.0062} \\

IN & ViT & BTK
& \stdval{0.9868}{0.0013}
& \stdval{0.9725}{0.0004}
& \stdbf{0.9870}{0.0005}
& \stdval{0.9843}{0.0006}
& \stdval{0.9620}{0.0026} \\

\midrule

OI & CLIP & TK
& \stdval{0.9764}{0.0015}
& \stdbf{0.9858}{0.0006}
& \stdval{0.9831}{0.0003}
& \stdval{0.9816}{0.0001}
& \stdval{0.9587}{0.0017} \\

OI & CLIP & BTK
& \stdval{0.9916}{0.0004}
& \stdbf{0.9957}{0.0003}
& \stdval{0.9944}{0.0002}
& \stdval{0.9847}{0.0007}
& \stdval{0.9453}{0.0006} \\

OI & SL2 & TK
& \stdval{0.9680}{0.0005}
& \stdbf{0.9690}{0.0008}
& \stdval{0.9674}{0.0004}
& \stdval{0.9665}{0.0002}
& \stdval{0.9257}{0.0055} \\

OI & SL2 & BTK
& \stdval{0.9864}{0.0004}
& \stdval{0.9814}{0.0006}
& \stdbf{0.9889}{0.0009}
& \stdval{0.9818}{0.0081}
& \stdval{0.9338}{0.0238} \\

OI & ViT & TK
& \stdbf{0.9629}{0.0012}
& \stdval{0.9611}{0.0006}
& \stdval{0.9601}{0.0014}
& \stdval{0.9521}{0.0022}
& \stdval{0.9117}{0.0078} \\

OI & ViT & BTK
& \stdbf{0.9625}{0.0016}
& \stdval{0.9593}{0.0006}
& \stdval{0.9499}{0.0096}
& \stdval{0.9036}{0.0356}
& \stdval{0.9059}{0.0075} \\

\bottomrule
\end{tabular}%
}
\end{center}

\begin{center}
\captionof{table}{Monosemanticity (MonoScore) under different
values of $\gamma$. The best result for each configuration
is highlighted in bold.
Our default $\gamma=10^{-2}$ gives the highest MonoScore in 10 of 12 configurations. IN = ImageNet-1K, OI = OpenImagesV7, SL2 = SigLIP2, TK = TopK, BTK = BatchTopK.}
\label{tab:gamma_ms}
\resizebox{0.61\textwidth}{!}{% width scaled with the column count to keep the previous font size
\begin{tabular}{lllccccc}
\toprule
\textbf{Data} & \textbf{Enc.} & \textbf{SAE}
& $\boldsymbol{\gamma=10^{-3}}$
& $\boldsymbol{\gamma=10^{-2}}$
& $\boldsymbol{\gamma=10^{-1}}$
& $\boldsymbol{\gamma=1}$
& $\boldsymbol{\gamma=10}$ \\
\midrule

IN & CLIP & TK
& \stdval{0.396}{0.004}
& \stdbf{0.560}{0.001}
& \stdval{0.187}{0.006}
& \stdval{0.212}{0.008}
& \stdval{0.086}{0.056} \\

IN & CLIP & BTK
& \stdval{0.391}{0.002}
& \stdbf{0.575}{0.000}
& \stdval{0.428}{0.006}
& \stdval{0.364}{0.024}
& \stdval{0.124}{0.014} \\

IN & SL2 & TK
& \stdval{0.537}{0.008}
& \stdbf{0.635}{0.001}
& \stdval{0.566}{0.012}
& \stdval{0.440}{0.156}
& \stdval{0.436}{0.043} \\

IN & SL2 & BTK
& \stdval{0.419}{0.005}
& \stdbf{0.623}{0.001}
& \stdval{0.578}{0.059}
& \stdval{0.495}{0.007}
& \stdval{0.460}{0.038} \\

IN & ViT & TK
& \stdval{0.261}{0.000}
& \stdbf{0.487}{0.001}
& \stdval{0.282}{0.001}
& \stdval{0.223}{0.001}
& \stdval{0.207}{0.013} \\

IN & ViT & BTK
& \stdval{0.275}{0.000}
& \stdbf{0.478}{0.001}
& \stdval{0.284}{0.000}
& \stdval{0.223}{0.002}
& \stdval{0.237}{0.004} \\

\midrule

OI & CLIP & TK
& \stdbf{0.399}{0.013}
& \stdval{0.388}{0.001}
& \stdval{0.263}{0.037}
& \stdval{0.166}{0.008}
& \stdval{0.125}{0.018} \\

OI & CLIP & BTK
& \stdbf{0.471}{0.001}
& \stdval{0.451}{0.000}
& \stdval{0.200}{0.008}
& \stdval{0.142}{0.006}
& \stdval{0.112}{0.003} \\

OI & SL2 & TK
& \stdval{0.517}{0.008}
& \stdbf{0.542}{0.001}
& \stdval{0.321}{0.021}
& \stdval{0.420}{0.068}
& \stdval{0.243}{0.026} \\

OI & SL2 & BTK
& \stdval{0.382}{0.020}
& \stdbf{0.564}{0.001}
& \stdval{0.429}{0.018}
& \stdval{0.410}{0.137}
& \stdval{0.177}{0.185} \\

OI & ViT & TK
& \stdval{0.146}{0.002}
& \stdbf{0.245}{0.001}
& \stdval{0.152}{0.002}
& \stdval{0.148}{0.003}
& \stdval{0.150}{0.013} \\

OI & ViT & BTK
& \stdval{0.268}{0.002}
& \stdbf{0.271}{0.000}
& \stdval{0.269}{0.012}
& \stdval{0.253}{0.005}
& \stdval{0.220}{0.000} \\

\bottomrule
\end{tabular}%
}
\end{center}
\vspace{1em}
}]

    \FloatBarrier
    \subsection{Decorrelation loss}
    \label{sec:quan_dec_loss}

    To isolate the contribution of the decorrelation loss, we keep all the settings fixed while removing \(\mathcal{L}_{\mathrm{dcor}}\). Table~\ref{tab:dcor-ablation} reports the performance difference between the ablated and full models. Removing the decorrelation loss reduces every metric across all encoders, activation functions, and datasets. The degradation is particularly pronounced for MonoScore and purity, indicating that allowing the two branches to reuse the same directions weakens the monosemantic structure of the Sparse Branch. The consistent reduction in \(R^2\), KNN accuracy, and LP accuracy further shows that the loss improves the branch decomposition without sacrificing reconstruction or downstream utility. These results suggest that \textit{{\(\mathcal{L}_{\mathrm{dcor}}\) is important for both interpretability and reconstruction in DualSAE.}}

    \begin{table*}[t]
        \centering
        \caption{
            \textbf{Ablation of the decorrelation loss.}
            We report the metric changes after removing
            \(\mathcal{L}_{\mathrm{dcor}}\). Every entry is negative, indicating that removing the decorrelation loss degrades performance. In other words, \(\mathcal{L}_{\mathrm{dcor}}\) allows the Sparse Branch to learn more interpretable representations (higher MS and Purity) and better reconstruction (higher $R^2$, KNN, and LP).
IN = ImageNet-1K, OI = OpenImagesV7, SL2 = SigLIP2, TK = TopK, BTK = BatchTopK, JR = JumpReLU.}
        \label{tab:dcor-ablation}

        \setlength{\tabcolsep}{4pt}
        \resizebox{0.7\linewidth}{!}{%
            \begin{tabular}{llccccc@{\hspace{8pt}}ccccc}
                \toprule
            \multirow{2}{*}{\textbf{Enc.}}
            & \multirow{2}{*}{\textbf{SAE}}
            & \multicolumn{5}{c}{\textbf{IN}}
            & \multicolumn{5}{c}{\textbf{OI}} \\
            \cmidrule(lr){3-7}
            \cmidrule(lr){8-12}
            &
            & \(\Delta R^2\)
            & \(\Delta\mathrm{MS}\)
            & \(\Delta\mathrm{Pur.}\)
            & \(\Delta\mathrm{KNN}\)
            & \(\Delta\mathrm{LP}\)
            & \(\Delta R^2\)
            & \(\Delta\mathrm{MS}\)
            & \(\Delta\mathrm{Pur.}\)
            & \(\Delta\mathrm{KNN}\)
            & \(\Delta\mathrm{LP}\) \\
            \midrule

            \multirow{3}{*}{ViT}
            & TK
            & -0.014 & -0.098 & -0.230 & -0.029 & -0.175
            & -0.028 & -0.033 & -0.042 & -0.019 & -0.005 \\

            & BTK
            & -0.021 & -0.072 & -0.179 & -0.045 & -0.001
            & -0.027 & -0.029 & -0.041 & -0.077 & -0.004 \\

            & JR
            & -0.013 & -0.232 & -0.362 & -0.004 & -0.015
            & -0.016 & -0.045 & -0.040 & -0.001 & -0.006 \\
            \midrule

            \multirow{3}{*}{CLIP}
            & TK
            & -0.022 & -0.051 & -0.015 & -0.022 & -0.001
            & -0.037 & -0.023 & -0.038 & -0.004 & -0.001 \\

            & BTK
            & -0.027 & -0.075 & -0.007 & -0.056 & -0.001
            & -0.046 & -0.052 & -0.021 & -0.003 & -0.001 \\

            & JR
            & -0.025 & -0.037 & -0.111 & -0.003 & -0.003
            & -0.040 & -0.051 & -0.031 & -0.001 & -0.005 \\
            \midrule

            \multirow{3}{*}{SL2}
            & TK
            & -0.022 & -0.013 & -0.054 & -0.012 & -0.008
            & -0.029 & -0.227 & -0.002 & -0.010 & -0.006 \\

            & BTK
            & -0.025 & -0.009 & -0.028 & -0.015 & -0.001
            & -0.060 & -0.273 & -0.026 & -0.002 & -0.009 \\

            & JR
            & -0.016 & -0.017 & -0.077 & -0.001 & -0.002
            & -0.061 & -0.031 & -0.039 & -0.001 & -0.005 \\
            \bottomrule
            \end{tabular}
        }
    \end{table*}

\FloatBarrier
\subsection{Projector of the Residual Branch}
\label{app:projector_ablation}

The Residual Branch maps the residual $r = x - x^M$ to the correction $\hat{r} = Q(W_P r)$ through a projector.
To test how sensitive DualSAE is to this choice, we vary only the projector and keep everything else
(Sparse Branch, losses, decorrelation coefficient, and two-stage schedule) fixed.
Let $A$ denote the encoder embedding dimension ($A=768$ for CLIP, $1024$ for ViT, and $1152$ for SigLIP2). We compare the default setting with:
\begin{itemize}
    \item \textbf{Default}: a single linear projector $\mathrm{Linear}(A,A)$ followed by GELU, without a hidden layer (590,592 parameters for CLIP).
    \item \textbf{B256} / \textbf{B512}: a bottleneck projector $\mathrm{Linear}(A,h) \rightarrow \mathrm{GELU} \rightarrow \mathrm{Linear}(h,A)$ with hidden width $h=256$ / $h=512$ (394K / 788K parameters for CLIP). A width $h<A$ restricts the residual correction to an $h$-dimensional subspace.
    \item \textbf{D3} / \textbf{D4}: deeper MLP projectors $A \rightarrow 512 \rightarrow 512 \rightarrow A$ and $A \rightarrow 512 \rightarrow 512 \rightarrow 512 \rightarrow A$ with GELU between layers (1.05M / 1.31M parameters for CLIP).
\end{itemize}
Each configuration is trained on ImageNet-1K and OpenImagesV7 with CLIP, SigLIP2, and ViT encoders and TopK and BatchTopK sparsity, using three seeds (42, 43, 44).
Results are shown in Table~\ref{tab:projector_ablation}.
The default single-layer projector gives the highest $R^2$ and the highest purity in every setting; the bottleneck and deeper projectors reduce $R^2$ by up to 0.18 (D4 on OpenImagesV7 with SigLIP2 and TopK).
MS is highest with the default projector in 8 of the 12 settings and within 0.031 of the best in the others, while on ImageNet-1K with ViT the alternative projectors reduce MS by 0.08--0.14.
KNN and LP vary by less than 0.005 across projectors.
Thus, a simple single-layer projector is sufficient, and adding depth or a bottleneck does not improve DualSAE.

\begin{table*}[p]
\centering
\caption{Projector ablation of the Residual Branch. Default: $\mathrm{Linear}(A,A) \rightarrow \mathrm{GELU}$;
B$h$: $\mathrm{Linear}(A,h) \rightarrow \mathrm{GELU} \rightarrow \mathrm{Linear}(h,A)$;
D3/D4: three-/four-layer MLP projectors with hidden width 512.
The default single-layer projector gives the highest $R^2$ and purity in every setting, while KNN and LP vary by less than 0.005 across projectors. IN = ImageNet-1K, OI = OpenImagesV7, SL2 = SigLIP2, TK = TopK, BTK = BatchTopK.
MS denotes MonoScore; Pur. denotes class purity; KNN and LP denote KNN- and linear-probe-based
downstream performance. Bold marks the best mean per setting.
Results are reported as mean $\pm$ standard deviation across three seeds.}
\label{tab:projector_ablation}
\scriptsize
\setlength{\tabcolsep}{5pt}
\renewcommand{\arraystretch}{0.99}
\begin{tabular}{@{}llllccccc@{}}
\toprule
Data & Enc. & SAE & Projector & $R^2$ & MS & Pur. & KNN & LP \\
\midrule
\multirow{30}{*}{IN} & \multirow{10}{*}{CLIP} & \multirow{5}{*}{TK} & \dc{Default} & \dc{\stdbf{0.9964}{0.0002}} & \dc{\stdval{0.5603}{0.0007}} & \dc{\stdbf{0.4256}{0.0002}} & \dc{\stdval{0.8155}{0.0002}} & \dc{\stdbf{0.8450}{0.0004}} \\
 &  &  & B256 & \stdval{0.9440}{0.0010} & \stdval{0.5612}{0.0038} & \stdval{0.2808}{0.0019} & \stdval{0.8153}{0.0004} & \stdval{0.8437}{0.0002} \\
 &  &  & B512 & \stdval{0.9886}{0.0001} & \stdbf{0.5851}{0.0028} & \stdval{0.3453}{0.0006} & \stdbf{0.8157}{0.0001} & \stdval{0.8449}{0.0002} \\
 &  &  & D3 & \stdval{0.9514}{0.0023} & \stdval{0.5622}{0.0048} & \stdval{0.2853}{0.0051} & \stdval{0.8148}{0.0002} & \stdval{0.8436}{0.0007} \\
 &  &  & D4 & \stdval{0.9290}{0.0013} & \stdval{0.5585}{0.0039} & \stdval{0.2718}{0.0071} & \stdval{0.8141}{0.0009} & \stdval{0.8427}{0.0004} \\
\cmidrule(lr){3-9}
 &  & \multirow{5}{*}{BTK} & \dc{Default} & \dc{\stdbf{0.9972}{0.0004}} & \dc{\stdval{0.5750}{0.0003}} & \dc{\stdbf{0.4287}{0.0004}} & \dc{\stdbf{0.8156}{0.0004}} & \dc{\stdbf{0.8451}{0.0003}} \\
 &  &  & B256 & \stdval{0.9397}{0.0027} & \stdval{0.5668}{0.0066} & \stdval{0.3062}{0.0083} & \stdval{0.8140}{0.0002} & \stdval{0.8434}{0.0003} \\
 &  &  & B512 & \stdval{0.9808}{0.0010} & \stdbf{0.5843}{0.0050} & \stdval{0.3460}{0.0057} & \stdval{0.8154}{0.0002} & \stdval{0.8450}{0.0001} \\
 &  &  & D3 & \stdval{0.9276}{0.0087} & \stdval{0.5680}{0.0027} & \stdval{0.3030}{0.0060} & \stdval{0.8133}{0.0006} & \stdval{0.8423}{0.0007} \\
 &  &  & D4 & \stdval{0.9033}{0.0122} & \stdval{0.5664}{0.0077} & \stdval{0.3000}{0.0127} & \stdval{0.8116}{0.0004} & \stdval{0.8409}{0.0009} \\
\cmidrule(lr){2-9}
 & \multirow{10}{*}{SL2} & \multirow{5}{*}{TK} & \dc{Default} & \dc{\stdbf{0.9944}{0.0005}} & \dc{\stdbf{0.6354}{0.0008}} & \dc{\stdbf{0.4036}{0.0008}} & \dc{\stdbf{0.8517}{0.0008}} & \dc{\stdbf{0.8830}{0.0005}} \\
 &  &  & B256 & \stdval{0.9041}{0.0195} & \stdval{0.6261}{0.0023} & \stdval{0.2741}{0.0132} & \stdval{0.8505}{0.0006} & \stdval{0.8809}{0.0006} \\
 &  &  & B512 & \stdval{0.9629}{0.0022} & \stdval{0.6227}{0.0032} & \stdval{0.2756}{0.0034} & \stdbf{0.8517}{0.0005} & \stdval{0.8829}{0.0002} \\
 &  &  & D3 & \stdval{0.9155}{0.0029} & \stdval{0.6193}{0.0019} & \stdval{0.2662}{0.0035} & \stdval{0.8511}{0.0001} & \stdval{0.8810}{0.0008} \\
 &  &  & D4 & \stdval{0.8761}{0.0131} & \stdval{0.6260}{0.0081} & \stdval{0.2746}{0.0145} & \stdval{0.8501}{0.0015} & \stdval{0.8791}{0.0006} \\
\cmidrule(lr){3-9}
 &  & \multirow{5}{*}{BTK} & \dc{Default} & \dc{\stdbf{0.9942}{0.0003}} & \dc{\stdbf{0.6231}{0.0009}} & \dc{\stdbf{0.3958}{0.0009}} & \dc{\stdbf{0.8516}{0.0005}} & \dc{\stdbf{0.8830}{0.0004}} \\
 &  &  & B256 & \stdval{0.9235}{0.0034} & \stdval{0.6011}{0.0006} & \stdval{0.2899}{0.0077} & \stdval{0.8507}{0.0002} & \stdval{0.8817}{0.0004} \\
 &  &  & B512 & \stdval{0.9660}{0.0009} & \stdval{0.6004}{0.0024} & \stdval{0.2934}{0.0037} & \stdval{0.8513}{0.0004} & \stdval{0.8829}{0.0002} \\
 &  &  & D3 & \stdval{0.9247}{0.0010} & \stdval{0.5969}{0.0071} & \stdval{0.2792}{0.0035} & \stdval{0.8507}{0.0003} & \stdval{0.8812}{0.0003} \\
 &  &  & D4 & \stdval{0.9057}{0.0028} & \stdval{0.5904}{0.0030} & \stdval{0.2718}{0.0046} & \stdval{0.8497}{0.0003} & \stdval{0.8801}{0.0005} \\
\cmidrule(lr){2-9}
 & \multirow{10}{*}{ViT} & \multirow{5}{*}{TK} & \dc{Default} & \dc{\stdbf{0.9661}{0.0003}} & \dc{\stdbf{0.4874}{0.0008}} & \dc{\stdbf{0.5666}{0.0002}} & \dc{\stdval{0.8226}{0.0008}} & \dc{\stdval{0.8239}{0.0004}} \\
 &  &  & B256 & \stdval{0.8880}{0.0012} & \stdval{0.3521}{0.0014} & \stdval{0.2717}{0.0031} & \stdval{0.8223}{0.0007} & \stdbf{0.8240}{0.0002} \\
 &  &  & B512 & \stdval{0.9285}{0.0014} & \stdval{0.3770}{0.0007} & \stdval{0.3129}{0.0002} & \stdbf{0.8228}{0.0004} & \stdval{0.8239}{0.0002} \\
 &  &  & D3 & \stdval{0.8618}{0.0038} & \stdval{0.3852}{0.0007} & \stdval{0.3329}{0.0017} & \stdval{0.8224}{0.0002} & \stdbf{0.8240}{0.0006} \\
 &  &  & D4 & \stdval{0.8744}{0.0004} & \stdval{0.3699}{0.0005} & \stdval{0.3056}{0.0003} & \stdval{0.8221}{0.0002} & \stdval{0.8236}{0.0007} \\
\cmidrule(lr){3-9}
 &  & \multirow{5}{*}{BTK} & \dc{Default} & \dc{\stdbf{0.9725}{0.0004}} & \dc{\stdbf{0.4775}{0.0005}} & \dc{\stdbf{0.5481}{0.0002}} & \dc{\stdbf{0.8228}{0.0003}} & \dc{\stdbf{0.8242}{0.0005}} \\
 &  &  & B256 & \stdval{0.8939}{0.0007} & \stdval{0.3668}{0.0011} & \stdval{0.3048}{0.0019} & \stdval{0.8223}{0.0001} & \stdval{0.8239}{0.0004} \\
 &  &  & B512 & \stdval{0.9332}{0.0003} & \stdval{0.3934}{0.0003} & \stdval{0.3467}{0.0004} & \stdval{0.8227}{0.0005} & \stdbf{0.8242}{0.0002} \\
 &  &  & D3 & \stdval{0.8601}{0.0037} & \stdval{0.4012}{0.0011} & \stdval{0.3642}{0.0008} & \stdval{0.8223}{0.0002} & \stdval{0.8237}{0.0004} \\
 &  &  & D4 & \stdval{0.8764}{0.0001} & \stdval{0.3845}{0.0001} & \stdval{0.3353}{0.0010} & \stdval{0.8221}{0.0006} & \stdval{0.8235}{0.0000} \\
\midrule
\multirow{30}{*}{OI} & \multirow{10}{*}{CLIP} & \multirow{5}{*}{TK} & \dc{Default} & \dc{\stdbf{0.9858}{0.0006}} & \dc{\stdval{0.3879}{0.0006}} & \dc{\stdbf{0.5800}{0.0002}} & \dc{\stdval{0.4699}{0.0002}} & \dc{\stdbf{0.4077}{0.0005}} \\
 &  &  & B256 & \stdval{0.8947}{0.0012} & \stdbf{0.4185}{0.0033} & \stdval{0.5220}{0.0026} & \stdval{0.4693}{0.0001} & \stdbf{0.4077}{0.0002} \\
 &  &  & B512 & \stdval{0.9682}{0.0004} & \stdval{0.4151}{0.0044} & \stdval{0.5524}{0.0092} & \stdbf{0.4700}{0.0001} & \stdval{0.4074}{0.0002} \\
 &  &  & D3 & \stdval{0.8951}{0.0012} & \stdval{0.3891}{0.0038} & \stdval{0.5263}{0.0039} & \stdval{0.4696}{0.0003} & \stdval{0.4074}{0.0004} \\
 &  &  & D4 & \stdval{0.8564}{0.0018} & \stdval{0.3883}{0.0067} & \stdval{0.5208}{0.0033} & \stdval{0.4692}{0.0002} & \stdval{0.4072}{0.0008} \\
\cmidrule(lr){3-9}
 &  & \multirow{5}{*}{BTK} & \dc{Default} & \dc{\stdbf{0.9957}{0.0003}} & \dc{\stdval{0.4510}{0.0003}} & \dc{\stdbf{0.6543}{0.0006}} & \dc{\stdbf{0.4700}{0.0005}} & \dc{\stdval{0.4070}{0.0006}} \\
 &  &  & B256 & \stdval{0.9020}{0.0013} & \stdval{0.4490}{0.0015} & \stdval{0.5231}{0.0015} & \stdval{0.4695}{0.0001} & \stdval{0.4091}{0.0002} \\
 &  &  & B512 & \stdval{0.9776}{0.0041} & \stdbf{0.4656}{0.0014} & \stdval{0.5596}{0.0047} & \stdbf{0.4700}{0.0001} & \stdval{0.4077}{0.0004} \\
 &  &  & D3 & \stdval{0.9184}{0.0111} & \stdval{0.4504}{0.0041} & \stdval{0.5177}{0.0141} & \stdval{0.4697}{0.0002} & \stdval{0.4092}{0.0006} \\
 &  &  & D4 & \stdval{0.8710}{0.0027} & \stdval{0.4521}{0.0050} & \stdval{0.5313}{0.0117} & \stdval{0.4690}{0.0005} & \stdbf{0.4102}{0.0005} \\
\cmidrule(lr){2-9}
 & \multirow{10}{*}{SL2} & \multirow{5}{*}{TK} & \dc{Default} & \dc{\stdbf{0.9690}{0.0008}} & \dc{\stdbf{0.5416}{0.0005}} & \dc{\stdbf{0.5944}{0.0003}} & \dc{\stdval{0.4723}{0.0004}} & \dc{\stdval{0.4104}{0.0008}} \\
 &  &  & B256 & \stdval{0.8257}{0.0022} & \stdval{0.4568}{0.0031} & \stdval{0.5338}{0.0019} & \stdbf{0.4726}{0.0003} & \stdval{0.4107}{0.0007} \\
 &  &  & B512 & \stdval{0.9144}{0.0013} & \stdval{0.4442}{0.0066} & \stdval{0.5282}{0.0005} & \stdval{0.4725}{0.0002} & \stdbf{0.4110}{0.0002} \\
 &  &  & D3 & \stdval{0.8236}{0.0016} & \stdval{0.4015}{0.0038} & \stdval{0.5242}{0.0005} & \stdbf{0.4726}{0.0001} & \stdval{0.4102}{0.0005} \\
 &  &  & D4 & \stdval{0.7882}{0.0011} & \stdval{0.4291}{0.0035} & \stdval{0.5342}{0.0045} & \stdval{0.4721}{0.0003} & \stdval{0.4103}{0.0006} \\
\cmidrule(lr){3-9}
 &  & \multirow{5}{*}{BTK} & \dc{Default} & \dc{\stdbf{0.9814}{0.0006}} & \dc{\stdbf{0.5635}{0.0010}} & \dc{\stdbf{0.6191}{0.0002}} & \dc{\stdval{0.4720}{0.0005}} & \dc{\stdval{0.4102}{0.0004}} \\
 &  &  & B256 & \stdval{0.8844}{0.0010} & \stdval{0.5222}{0.0027} & \stdval{0.4946}{0.0020} & \stdbf{0.4723}{0.0000} & \stdval{0.4096}{0.0006} \\
 &  &  & B512 & \stdval{0.9579}{0.0004} & \stdval{0.5518}{0.0024} & \stdval{0.5133}{0.0042} & \stdval{0.4720}{0.0001} & \stdval{0.4102}{0.0002} \\
 &  &  & D3 & \stdval{0.8906}{0.0003} & \stdval{0.5150}{0.0026} & \stdval{0.5132}{0.0023} & \stdval{0.4722}{0.0001} & \stdval{0.4106}{0.0005} \\
 &  &  & D4 & \stdval{0.8560}{0.0011} & \stdval{0.5146}{0.0016} & \stdval{0.5078}{0.0017} & \stdval{0.4721}{0.0000} & \stdbf{0.4109}{0.0006} \\
\cmidrule(lr){2-9}
 & \multirow{10}{*}{ViT} & \multirow{5}{*}{TK} & \dc{Default} & \dc{\stdbf{0.9611}{0.0006}} & \dc{\stdbf{0.2446}{0.0008}} & \dc{\stdbf{0.5028}{0.0001}} & \dc{\stdbf{0.4701}{0.0010}} & \dc{\stdval{0.4049}{0.0005}} \\
 &  &  & B256 & \stdval{0.8293}{0.0009} & \stdval{0.2142}{0.0010} & \stdval{0.4538}{0.0026} & \stdval{0.4693}{0.0001} & \stdbf{0.4060}{0.0003} \\
 &  &  & B512 & \stdval{0.8912}{0.0003} & \stdval{0.2229}{0.0017} & \stdval{0.4681}{0.0030} & \stdval{0.4698}{0.0002} & \stdval{0.4042}{0.0003} \\
 &  &  & D3 & \stdval{0.7864}{0.0179} & \stdval{0.2221}{0.0010} & \stdval{0.4687}{0.0014} & \stdval{0.4695}{0.0001} & \stdval{0.4055}{0.0012} \\
 &  &  & D4 & \stdval{0.8152}{0.0029} & \stdval{0.2182}{0.0012} & \stdval{0.4646}{0.0020} & \stdval{0.4693}{0.0001} & \stdbf{0.4060}{0.0004} \\
\cmidrule(lr){3-9}
 &  & \multirow{5}{*}{BTK} & \dc{Default} & \dc{\stdbf{0.9593}{0.0006}} & \dc{\stdbf{0.2708}{0.0003}} & \dc{\stdbf{0.5284}{0.0002}} & \dc{\stdbf{0.4700}{0.0006}} & \dc{\stdval{0.4035}{0.0003}} \\
 &  &  & B256 & \stdval{0.7977}{0.0188} & \stdval{0.2463}{0.0027} & \stdval{0.5045}{0.0116} & \stdval{0.4691}{0.0002} & \stdbf{0.4069}{0.0009} \\
 &  &  & B512 & \stdval{0.8776}{0.0053} & \stdval{0.2539}{0.0007} & \stdval{0.5048}{0.0052} & \stdval{0.4697}{0.0001} & \stdval{0.4038}{0.0007} \\
 &  &  & D3 & \stdval{0.7914}{0.0061} & \stdval{0.2500}{0.0013} & \stdval{0.4945}{0.0027} & \stdval{0.4694}{0.0001} & \stdval{0.4057}{0.0004} \\
 &  &  & D4 & \stdval{0.8091}{0.0034} & \stdval{0.2459}{0.0013} & \stdval{0.4888}{0.0010} & \stdval{0.4691}{0.0003} & \stdval{0.4061}{0.0005} \\
\bottomrule
\end{tabular}
\end{table*}

\subsection{Joint training vs.\ sequential training}
\label{app:frozen_sae}

This section gives the full results of the joint vs.\ sequential training experiment in the main paper (Sec.~\ref{exp:why}).
Sequential training first trains a standard SAE that uses DualSAE's Sparse Branch without a Residual Branch ($\hat{x}=x^M$), then freezes this Sparse Branch and trains a freshly initialized Residual Branch of the same architecture alone on $\mathcal{L}_{\mathrm{rec}}+\gamma\mathcal{L}_{\mathrm{dcor}}$, with the hyperparameters of the corresponding DualSAE run, including $\gamma$ and the warm-up length.
It is evaluated after the same number of Residual Branch updates that the jointly trained model receives up to its selected checkpoint.
Joint training is DualSAE.
Table~\ref{tab:frozen_vs_dualsae} reports all 12 settings (ImageNet-1K and OpenImagesV7; CLIP, SigLIP2, and ViT; TopK and BatchTopK).

% \begin{figure*}[t]
% \centering
% \includegraphics[width=\textwidth]{Figures/joint_vs_seq_tradeoff_all.pdf}
% \caption{\textbf{Reconstruction--interpretability trade-off of joint vs.\ sequential training in all 12 settings} (rows: dataset and SAE; columns: encoder). \(R^2\) against MonoScore for a standard SAE (SAE), the same SAE frozen with a Residual Branch trained on top (SAE + Residual, \ie, sequential training), and DualSAE (both branches trained jointly). Dashed arrows give relative changes: blue, \(R^2\) from SAE to SAE + Residual; orange, MonoScore (vertical) and, where it differs by at least 2\%, \(R^2\) (horizontal) from SAE + Residual to DualSAE. The Residual Branch raises \(R^2\) by 15--45\% without changing MonoScore; joint training reaches a similar \(R^2\) (within 0.7\% for CLIP and SigLIP2, higher for ViT) and raises MonoScore in 11 of 12 settings (by 10--49\%). Mean over three seeds, all from the same evaluation.}
% \label{fig:joint_vs_seq_all}
% \end{figure*}

\begin{table*}[t]
\centering
\caption{Sequential Training vs.\ Joint Training. Sequential Training: a pre-trained SAE whose Sparse Branch is frozen while a Residual Branch is trained afterwards; Joint Training: both branches trained jointly. 
Sequential training recovers most of DualSAE's reconstruction fidelity and downstream performance, but joint training gives higher MonoScore and purity in all 12 settings. IN = ImageNet-1K, OI = OpenImagesV7, SL2 = SigLIP2, TK = TopK, BTK = BatchTopK.
MS denotes MonoScore; Pur. denotes class purity; KNN and LP denote KNN- and linear-probe-based
downstream performance (top-1 accuracy on ImageNet-1K, micro-F1 on OpenImagesV7).
Bold marks the higher mean in each setting.
Results are reported as mean $\pm$ standard deviation across three seeds.}
\label{tab:frozen_vs_dualsae}
\footnotesize
\setlength{\tabcolsep}{5pt}
\begin{tabular}{@{}llllccccc@{}}
\toprule
Data & Enc. & SAE & Method & $R^2$ & MS & Pur. & KNN & LP \\
\midrule
\multirow{12}{*}{IN}
& \multirow{4}{*}{CLIP} & \multirow{2}{*}{TK} & Sequential Training & \stdbf{0.9983}{0.0001} & \stdval{0.5563}{0.0025} & \stdval{0.2746}{0.0049} & \stdbf{0.8155}{0.0002} & \stdbf{0.8452}{0.0000} \\
& & & \dc{Joint Training} & \dc{\stdval{0.9964}{0.0002}} & \dc{\stdbf{0.5603}{0.0007}} & \dc{\stdbf{0.4256}{0.0002}} & \dc{\stdbf{0.8155}{0.0002}} & \dc{\stdval{0.8450}{0.0004}} \\
&  & \multirow{2}{*}{BTK} & Sequential Training & \stdbf{0.9987}{0.0000} & \stdval{0.5206}{0.0149} & \stdval{0.2822}{0.0010} & \stdval{0.8155}{0.0002} & \stdbf{0.8451}{0.0001} \\
& & & \dc{Joint Training} & \dc{\stdval{0.9972}{0.0004}} & \dc{\stdbf{0.5750}{0.0003}} & \dc{\stdbf{0.4287}{0.0004}} & \dc{\stdbf{0.8156}{0.0004}} & \dc{\stdbf{0.8451}{0.0003}} \\
\cmidrule(lr){2-9}
& \multirow{4}{*}{SL2} & \multirow{2}{*}{TK} & Sequential Training & \stdbf{0.9969}{0.0001} & \stdval{0.5596}{0.0702} & \stdval{0.2742}{0.0098} & \stdbf{0.8517}{0.0000} & \stdbf{0.8837}{0.0001} \\
& & & \dc{Joint Training} & \dc{\stdval{0.9944}{0.0005}} & \dc{\stdbf{0.6354}{0.0008}} & \dc{\stdbf{0.4036}{0.0008}} & \dc{\stdbf{0.8517}{0.0008}} & \dc{\stdval{0.8830}{0.0005}} \\
&  & \multirow{2}{*}{BTK} & Sequential Training & \stdbf{0.9970}{0.0001} & \stdval{0.4958}{0.0323} & \stdval{0.3022}{0.0033} & \stdbf{0.8517}{0.0001} & \stdbf{0.8838}{0.0001} \\
& & & \dc{Joint Training} & \dc{\stdval{0.9942}{0.0003}} & \dc{\stdbf{0.6231}{0.0009}} & \dc{\stdbf{0.3958}{0.0009}} & \dc{\stdval{0.8516}{0.0005}} & \dc{\stdval{0.8830}{0.0004}} \\
\cmidrule(lr){2-9}
& \multirow{4}{*}{ViT} & \multirow{2}{*}{TK} & Sequential Training & \stdval{0.9328}{0.0004} & \stdval{0.3281}{0.0004} & \stdval{0.2444}{0.0001} & \stdbf{0.8227}{0.0002} & \stdbf{0.8242}{0.0003} \\
& & & \dc{Joint Training} & \dc{\stdbf{0.9661}{0.0003}} & \dc{\stdbf{0.4874}{0.0008}} & \dc{\stdbf{0.5666}{0.0002}} & \dc{\stdval{0.8226}{0.0008}} & \dc{\stdval{0.8239}{0.0004}} \\
&  & \multirow{2}{*}{BTK} & Sequential Training & \stdval{0.9104}{0.0020} & \stdval{0.3342}{0.0005} & \stdval{0.2647}{0.0005} & \stdbf{0.8231}{0.0004} & \stdbf{0.8242}{0.0002} \\
& & & \dc{Joint Training} & \dc{\stdbf{0.9725}{0.0004}} & \dc{\stdbf{0.4775}{0.0005}} & \dc{\stdbf{0.5481}{0.0002}} & \dc{\stdval{0.8228}{0.0003}} & \dc{\stdbf{0.8242}{0.0005}} \\
\midrule
\multirow{12}{*}{OI}
& \multirow{4}{*}{CLIP} & \multirow{2}{*}{TK} & Sequential Training & \stdbf{0.9936}{0.0000} & \stdval{0.2952}{0.0052} & \stdval{0.5019}{0.0053} & \stdbf{0.4702}{0.0001} & \stdbf{0.4078}{0.0000} \\
& & & \dc{Joint Training} & \dc{\stdval{0.9858}{0.0006}} & \dc{\stdbf{0.3879}{0.0006}} & \dc{\stdbf{0.5800}{0.0002}} & \dc{\stdval{0.4699}{0.0002}} & \dc{\stdval{0.4077}{0.0005}} \\
&  & \multirow{2}{*}{BTK} & Sequential Training & \stdbf{0.9972}{0.0002} & \stdval{0.3012}{0.0248} & \stdval{0.5095}{0.0045} & \stdbf{0.4701}{0.0001} & \stdbf{0.4070}{0.0001} \\
& & & \dc{Joint Training} & \dc{\stdval{0.9957}{0.0003}} & \dc{\stdbf{0.4510}{0.0003}} & \dc{\stdbf{0.6543}{0.0006}} & \dc{\stdval{0.4700}{0.0005}} & \dc{\stdbf{0.4070}{0.0006}} \\
\cmidrule(lr){2-9}
& \multirow{4}{*}{SL2} & \multirow{2}{*}{TK} & Sequential Training & \stdbf{0.9753}{0.0013} & \stdval{0.3214}{0.0028} & \stdval{0.5035}{0.0038} & \stdval{0.4722}{0.0001} & \stdbf{0.4112}{0.0001} \\
& & & \dc{Joint Training} & \dc{\stdval{0.9690}{0.0008}} & \dc{\stdbf{0.5416}{0.0005}} & \dc{\stdbf{0.5944}{0.0003}} & \dc{\stdbf{0.4723}{0.0004}} & \dc{\stdval{0.4104}{0.0008}} \\
&  & \multirow{2}{*}{BTK} & Sequential Training & \stdbf{0.9867}{0.0026} & \stdval{0.3805}{0.0442} & \stdval{0.5377}{0.0097} & \stdbf{0.4722}{0.0001} & \stdbf{0.4109}{0.0001} \\
& & & \dc{Joint Training} & \dc{\stdval{0.9814}{0.0006}} & \dc{\stdbf{0.5635}{0.0010}} & \dc{\stdbf{0.6191}{0.0002}} & \dc{\stdval{0.4720}{0.0005}} & \dc{\stdval{0.4102}{0.0004}} \\
\cmidrule(lr){2-9}
& \multirow{4}{*}{ViT} & \multirow{2}{*}{TK} & Sequential Training & \stdval{0.9545}{0.0005} & \stdval{0.2075}{0.0008} & \stdval{0.4527}{0.0010} & \stdval{0.4699}{0.0001} & \stdval{0.4043}{0.0000} \\
& & & \dc{Joint Training} & \dc{\stdbf{0.9611}{0.0006}} & \dc{\stdbf{0.2446}{0.0008}} & \dc{\stdbf{0.5028}{0.0001}} & \dc{\stdbf{0.4701}{0.0010}} & \dc{\stdbf{0.4049}{0.0005}} \\
&  & \multirow{2}{*}{BTK} & Sequential Training & \stdval{0.9491}{0.0023} & \stdval{0.2060}{0.0006} & \stdval{0.4592}{0.0026} & \stdval{0.4699}{0.0002} & \stdbf{0.4146}{0.0003} \\
& & & \dc{Joint Training} & \dc{\stdbf{0.9593}{0.0006}} & \dc{\stdbf{0.2708}{0.0003}} & \dc{\stdbf{0.5284}{0.0002}} & \dc{\stdbf{0.4700}{0.0006}} & \dc{\stdval{0.4035}{0.0003}} \\
\bottomrule
\end{tabular}
\end{table*}

\FloatBarrier
\subsection{Expansion ratio}
\label{app:expansion}

\begin{figure*}[t]
\centering
\includegraphics[width=0.8\linewidth]{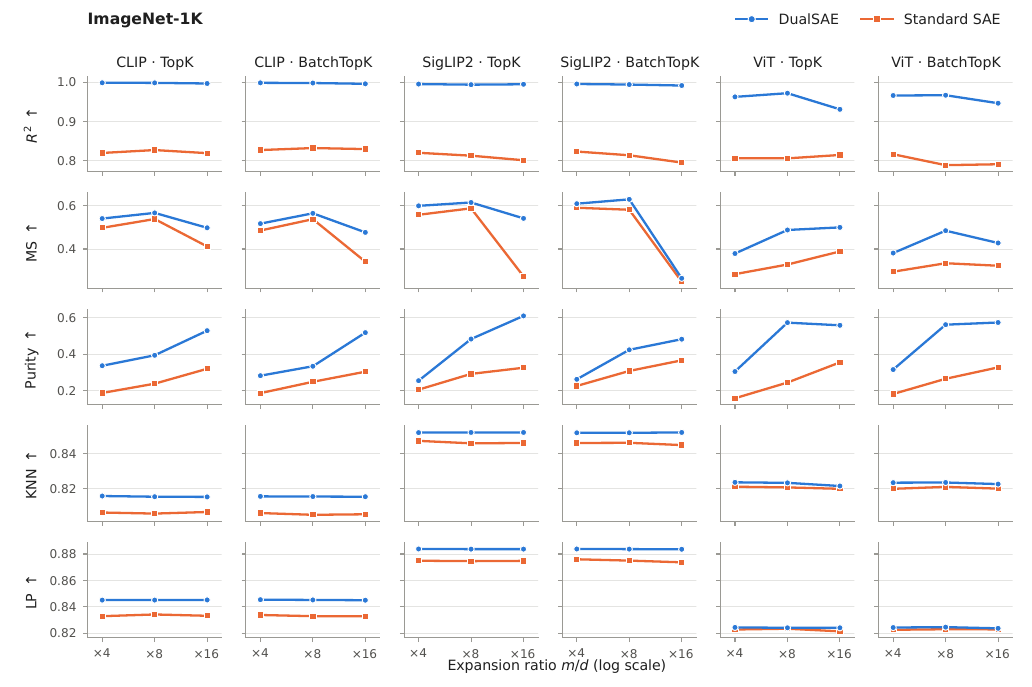}
\caption{Expansion-ratio ablation on ImageNet-1K.
DualSAE has higher $R^2$, MS, purity, KNN, and LP than the standard SAE at every expansion ratio in all six settings. Each panel plots one metric (rows) against the expansion ratio $m/d$ (number of latents over the encoder's embedding dimension, log scale) for one encoder and activation (columns).
Standard SAE: DualSAE's Sparse Branch trained without the Residual Branch (as in Supplementary~\ref{app:frozen_sae}), not the baselines of Table~\ref{tab:sae_comparison_imagenet_openimages_sidebyside}.
All panels in a row share the same y-axis.
MS denotes MonoScore and Purity class purity; KNN and LP are top-1 accuracy.}
\label{fig:expansion_in}
\end{figure*}

\begin{figure*}[t]
\centering
\includegraphics[width=0.8\linewidth]{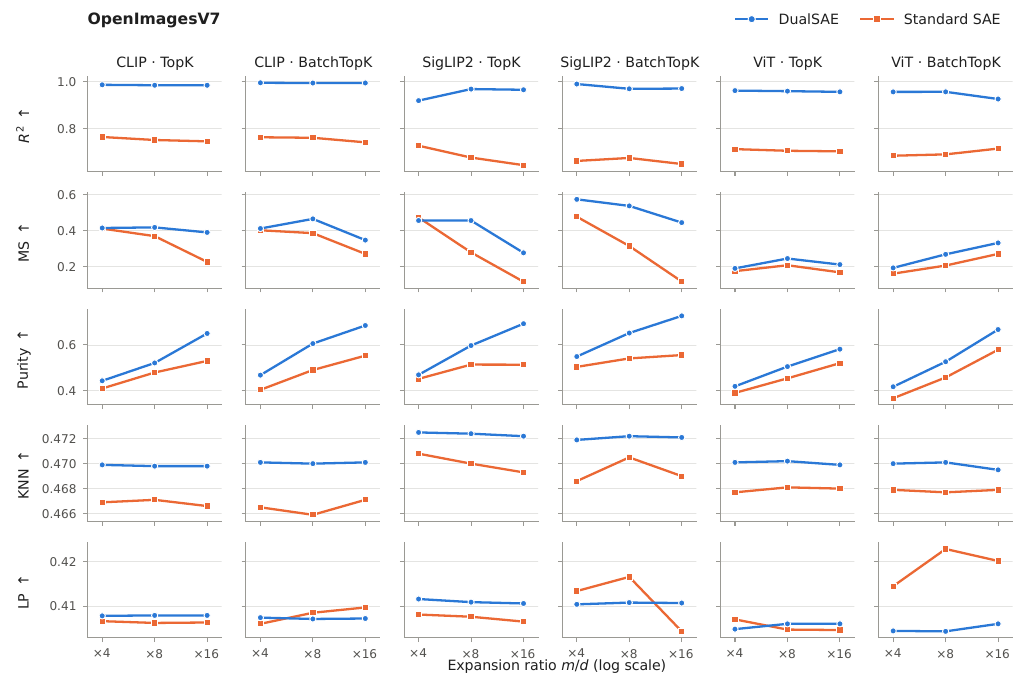}
\caption{Expansion-ratio ablation on OpenImagesV7.
DualSAE has higher $R^2$, purity, and KNN than the standard SAE in all 18 pairs and higher MS in 17; its LP is higher in 10 and at most 0.019 lower in the others. Each panel plots one metric (rows) against the expansion ratio $m/d$ (number of latents over the encoder's embedding dimension, log scale) for one encoder and activation (columns).
Standard SAE: DualSAE's Sparse Branch trained without the Residual Branch (as in Supplementary~\ref{app:frozen_sae}), not the baselines of Table~\ref{tab:sae_comparison_imagenet_openimages_sidebyside}.
All panels in a row share the same y-axis.
MS denotes MonoScore and Purity class purity; KNN and LP are micro-F1.}
\label{fig:expansion_oi}
\end{figure*}

All other experiments use a dictionary of $m=8192$ latents. Relative to the embedding dimension $d$ of each encoder ($d=768$ for CLIP, $1024$ for ViT, and $1152$ for SigLIP2), this is an expansion ratio $m/d$ of about 10.7 for CLIP, 8 for ViT, and 7.1 for SigLIP2.
To test whether the gains of DualSAE depend on this choice, we train DualSAE and a standard SAE with $m/d\in\{4,8,16\}$ on ImageNet-1K and OpenImagesV7, using CLIP, SigLIP2, and ViT encoders with TopK and BatchTopK sparsity.
We keep $k=64$ and all other hyperparameters at their default values (Supplementary~\ref{app:hyper-param}), including the decorrelation coefficient $\gamma$, and we do not resample dead latents.
The standard SAE is DualSAE's Sparse Branch trained without the Residual Branch, as in Supplementary~\ref{app:frozen_sae}, so the two models differ only in the Residual Branch.
In two runs on ImageNet-1K at $\times16$, DualSAE with SigLIP2 and BatchTopK and the standard SAE with ViT and BatchTopK, the validation MonoScore used for checkpoint selection peaked early in training (before the Residual Branch is unfrozen, and in the first epoch, respectively).
We retrain these two runs and select their checkpoints only from epochs after the Residual Branch is unfrozen (DualSAE) or from epoch 10 onward (standard SAE).
This gives 36 DualSAE--standard SAE pairs.
Figures~\ref{fig:expansion_in} and~\ref{fig:expansion_oi} show the results on ImageNet-1K and OpenImagesV7, respectively.

\textbf{Reconstruction.}
DualSAE achieves higher $R^2$ than the standard SAE in all 36 pairs, by 0.12 to 0.33.
Its $R^2$ stays above 0.91 at every expansion ratio, and even at $\times4$ it exceeds the $R^2$ of the standard SAE at every expansion ratio in the same setting.

\textbf{Interpretability.}
DualSAE achieves higher MS in 35 of the 36 pairs (the exception is OpenImagesV7 with SigLIP2 and TopK at $\times4$, by 0.017) and higher class purity in all 36.
For DualSAE, MS is highest at $\times8$ in 8 of the 12 settings; the exceptions are the two SigLIP2 settings on OpenImagesV7 (highest at $\times4$) and two ViT settings (highest at $\times16$).
For both models, MS decreases from $\times8$ to $\times16$ in 10 of the 12 settings; the exceptions are ViT with TopK on ImageNet-1K and ViT with BatchTopK on OpenImagesV7.
This decrease coincides with a growing fraction of dead latents (latents that never activate on the evaluation set), which at $\times16$ reaches 60\% for DualSAE and 80\% for the standard SAE; DualSAE has fewer dead latents than the standard SAE in 9 of the 12 pairs at $\times16$.
Purity, in contrast, increases with the expansion ratio for both models: in all 12 settings, the $\times16$ dictionary has higher purity than the $\times4$ dictionary.

\textbf{Downstream performance.}
KNN and LP change by less than 0.003 across expansion ratios, except for the LP of the standard SAE (up to 0.012).
DualSAE achieves higher KNN in all 36 pairs and higher LP in 28; the 8 exceptions are all on OpenImagesV7, 7 of them with BatchTopK, and DualSAE trails by at most 0.019.

Overall, the advantages of DualSAE in reconstruction, MS, purity, and KNN hold at every expansion ratio from $\times4$ to $\times16$, with a single MS exception, so they do not depend on the default dictionary size.

\clearpage
\FloatBarrier % after \clearpage, restores the full column height so Section D starts in the left column (see preamble.tex)
\section{Evaluation on language-model representations}
\label{app:llm}

\textbf{Language model and activation extraction.}
We evaluate DualSAE on residual-stream activations from
\texttt{EleutherAI/pythia-410m-deduped}.
The model contains 24 transformer blocks with hidden dimension $d=1024$.
We extract residual-stream activations after block 12, corresponding to
a middle-layer representation.
Activations are collected using forward hooks and stored once for each
corpus and data split, ensuring that all SAE variants evaluated under the
same setting receive identical activations.

\textbf{Corpora.}
We consider two substantially different training corpora.
\textbf{OpenWebText} represents general web text, while
\textbf{GitHub code} is obtained from
\texttt{codeparrot/github-code-clean} and contains mixed-language source
code.
This allows us to evaluate whether DualSAE generalizes across distinct
residual-stream distributions.
Each SAE is trained on a single corpus, and activations are never mixed
across corpora.

For the GitHub corpus, we remove empty and duplicated documents.
We additionally verify that the corpus is not ordered by programming
language: the first shard contains 30 languages across 108,751 runs and
126,925 rows, with different languages interleaved throughout the shard.

\textbf{Data preparation.}
Documents are packed into contexts of 128 tokens without padding.
Each training stream contains exactly 64M tokens, matching the optimization
budget such that each SAE observes one pass over its training stream.
We construct two disjoint held-out sets of 200K tokens each from documents
occurring strictly after those used for training.
The first held-out split is used for model selection, while the second is
reserved for final evaluation.

\textbf{Activation normalization.}
For each corpus, we normalize activations using a single global scale
computed as the root-mean-square activation norm over a 2M-token subset
of the corresponding training split:
\[
    s =
    \sqrt{
        \mathbb{E}_{x}
        \left[\lVert x\rVert_2^2\right]
    },
    \qquad
    \tilde{x} = \frac{x}{s}.
\]
We use a global scale rather than per-token normalization so that variation
in activation magnitude is preserved, which is important when reconstructed
activations are spliced back into the language model.
The resulting scale at block 12 is $125.65$ for OpenWebText and $127.97$
for GitHub code.

\textbf{SAE configuration.}
All SAE variants use 8,192 latent features with target sparsity
$L_0=64$.
We train each model for 31,250 optimization steps with batch size 2,048
using Adam with learning rate $10^{-4}$.
Each configuration is evaluated over three random seeds.
DualSAE follows our two-stage training scheme: during the first stage,
the Residual Branch is frozen and only the Sparse Branch is optimized;
the Residual Branch is subsequently enabled for joint training.
All comparisons are made between models trained on the same corpus using
the same block-12 activations.

\textbf{Evaluation metrics.}
We evaluate each SAE using four metrics:
$\Delta$CE, KL divergence, $R^2$, and MonoScore (MS).

\textbf{Cross-entropy change ($\Delta$CE).}
To evaluate whether the reconstructed representation preserves the
language model's behavior, we splice the SAE reconstruction $\hat{x}$
back into the residual stream at block 12 and run the remainder of the
language model normally.
We compute
\[
    \Delta \mathrm{CE}
    =
    \mathrm{CE}_{\mathrm{splice}}
    -
    \mathrm{CE}_{\mathrm{clean}},
\]
where $\mathrm{CE}_{\mathrm{clean}}$ is the next-token cross-entropy of
the original model and $\mathrm{CE}_{\mathrm{splice}}$ is obtained after
replacing the residual-stream activation with its SAE reconstruction.
A smaller $\Delta\mathrm{CE}$ indicates better preservation of the
original model behavior.

\textbf{KL divergence.}
We additionally measure the KL divergence between the clean and spliced
next-token distributions:
\[
    D_{\mathrm{KL}}
    \left(
        p_{\mathrm{clean}}
        \,\Vert\,
        p_{\mathrm{splice}}
    \right).
\]
Unlike $\Delta$CE, which depends only on the probability assigned to the
ground-truth next token, KL divergence measures changes across the full
output distribution.
Lower KL indicates better preservation of the model's predictive
distribution.

\textbf{Reconstruction quality ($R^2$).}
We measure reconstruction fidelity using pooled $R^2$:
\[
    R^2 =
    1 -
    \frac{
        \sum_i \lVert x_i-\hat{x}_i\rVert_2^2
    }{
        \sum_i \lVert x_i-\bar{x}\rVert_2^2
    },
\]
where $\bar{x}$ denotes the mean activation over the evaluation set.
Higher $R^2$ indicates more faithful reconstruction.

\textbf{Monosemanticity (MS).}
We evaluate the semantic coherence of the learned sparse features using
MonoScore (MS).
MonoScore measures the semantic consistency of examples that strongly
activate each latent feature, with larger values indicating more
monosemantic features.

Overall, lower $\Delta$CE and KL indicate better behavioral preservation,
while higher $R^2$ and MS indicate better reconstruction fidelity and
monosemanticity, respectively.

\textbf{More qualitative examples.} Beside the example shown in main manuscript, we present more qualitative examples here for more comprehensive view. Please refer to Figures~\ref{fig:text_qualitative_name_latent}--\ref{fig:text_class_attribute_latent}.

\begin{figure*}[t]
\centering
\small

\noindent
\textbf{\large ``Person Name" latents}

\vspace{0.6em}

\begin{minipage}[t]{0.485\textwidth}
\begin{tcolorbox}[
    colback=white,
    colframe=gray!25,
    boxrule=0.6pt,
    arc=2mm,
    left=3mm,
    right=3mm,
    top=2mm,
    bottom=2mm
]

\textbf{\large Dual-BatchTopK}
\textcolor{gray!70}{latent 7597}

\vspace{0.45em}

\vspace{0.8em}

\textbf{\textcolor{gray!80}{STRONGLY ACTIVATING (5)}}
\vspace{0.3em}

\begin{tabularx}{\linewidth}{@{}p{0.09\linewidth}X@{}}
\actscore{0.60} &
\texttt{a show I heard on NPR last week. The reporter is }
\dualhl{Stephen Beard}
\texttt{, 's bureau chief and what his commentary was nothing}
\\[0.6em]

\actscore{0.57} &
\texttt{Sturgeon's Law still hold true? You can connect with }
\dualhl{Richard}
\texttt{ on Google+. <|endoftext|>The Advertising Standards Authority (ASA}
\\[0.6em]

\actscore{0.53} &
\texttt{.com/R2suBp06fS -- }
\dualhl{Jim McElwain}
\texttt{  (@CoachMcElwain)}
\\[0.6em]

\actscore{0.49} &
\texttt{trade globalization. To keep investment and jobs at home, }
\dualhl{William Greider}
\texttt{ has urged that Washington "rewrite trade law,"}
\\[0.6em]

\actscore{0.47} &
\texttt{-- Editor's note: When reached for comment, }
\dualhl{Michael Burton}
\texttt{, the commercial broker for the bindery building and its}
\end{tabularx}

\vspace{0.7em}

\textbf{\textcolor{gray!80}{MODERATELY ACTIVATING (3)}}

\vspace{0.3em}

\begin{tabularx}{\linewidth}{@{}p{0.09\linewidth}X@{}}
\actscore{0.28} &
\texttt{Khouri, Undersecretary for the Ministry of Finance, }
\dualhl{Abdullah Salem Al Dhaheri}
\texttt{, CEO of Adnoc}
\\[0.6em]

\actscore{0.24} &
\texttt{now." In other words-- he likes him. -- }
\dualhl{Christy Chirinos}
\texttt{August 7}
\\[0.6em]

\actscore{0.14} &
\texttt{rights, his spokesman, }
\dualhl{Stephane Dujarric}
\texttt{, said on Thursday in New York.}
\end{tabularx}

\end{tcolorbox}
\end{minipage}
\hfill
\begin{minipage}[t]{0.485\textwidth}
\begin{tcolorbox}[
    colback=white,
    colframe=gray!25,
    boxrule=0.6pt,
    arc=2mm,
    left=3mm,
    right=3mm,
    top=2mm,
    bottom=2mm
]

\textbf{\large BatchTopK}
\textcolor{gray!70}{latent 3329}

\vspace{0.45em}

\vspace{0.8em}

\textbf{\textcolor{gray!80}{STRONGLY ACTIVATING (5)}}
\vspace{0.3em}

\begin{tabularx}{\linewidth}{@{}p{0.09\linewidth}X@{}}
\actscore{0.55} &
\texttt{yster selection and entrees at this jewel box seafood restaurant from }
\basehl{Larry Maguire}
\texttt{ 's team. Counter Cafe}
\\[0.6em]

\actscore{0.45} &
\texttt{electronic cigarettes. As a blog, I have competition, like }
\basehl{Steve Vape,}
\missbl{Ecig Advanced,}
\missbl{Vapor Joes}
\\[0.6em]

\actscore{0.42} &
\texttt{!" For more, read: Here's who }
\basehl{Steve}
\missbl{Wozniak}
\texttt{ is voting for in 2016 The}
\\[0.6em]

\actscore{0.41} &
\texttt{L) co-founder }
\basehl{Steve}
\missbl{Wozniak}
\texttt{has a vision to create a new kind}
\\[0.6em]

\actscore{0.40} &
\texttt{of hard work from}
\missbl{IndyCar}
\texttt{helping us being the scenes,}
\basehl{Michael Andretti}
\texttt{too, and it wouldn't have happened}
\end{tabularx}

\vspace{0.7em}

\textbf{\textcolor{gray!80}{MODERATELY ACTIVATING (3)}}

\vspace{0.3em}

\begin{tabularx}{\linewidth}{@{}p{0.09\linewidth}X@{}}
\actscore{0.23} &
\basehl{Franz}
\missbl{Liszt}
\texttt{,the other towering virtuoso pianist}
\\[0.6em]

\actscore{0.13} &
\texttt{this kind of stuff," he said. Premier }
\basehl{Brad}
\missbl{Wall}
\texttt{,who took to Twitter over the weekend to ask for}
\\[0.6em]

\actscore{0.12} &
\texttt{Salzberg, Atlanta Gastroenterology Specialists; }
\basehl{Pierre}
\missbl{Desreumaux}
\texttt{, Hopital }
\basehl{Claude}
\missbl{ Huriez}
\end{tabularx}

\end{tcolorbox}
\end{minipage}

\caption{\textbf{Qualitative comparison of ``person name'' latents learned by Dual-BatchTopK and standard BatchTopK.}
Both latents respond strongly to \dualhlcap{person names}. Dual-BatchTopK exhibits more consistent activation on complete person-name spans, whereas the standard BatchTopK latent also activates on \missblcap{unrelated or only partially relevant tokens}, as highlighted in gray. }
\label{fig:text_qualitative_name_latent}
\end{figure*}

\begin{figure*}[t]
\centering
\small

\textbf{\large ``Professional Title'' latents}

\vspace{0.6em}

\begin{minipage}[t]{0.485\textwidth}
\begin{tcolorbox}[
    colback=white,
    colframe=gray!25,
    boxrule=0.6pt,
    arc=2mm,
    left=3mm,
    right=3mm,
    top=2mm,
    bottom=2mm
]

\textbf{\large Dual-JumpReLU}
\textcolor{gray!70}{latent 4637}

\vspace{0.75em}

\textbf{\textcolor{gray!70}{STRONGLY ACTIVATING (5)}}

\vspace{0.3em}

\begin{tabularx}{\linewidth}{@{}p{0.10\linewidth}X@{}}

\actscore{0.48} &
\texttt{At a Monday news conference, White House deputy}
\texttt{ national }
\dualhl{security adviser}
\texttt{ Ben Rhodes encouraged the would-be alliance members to ``stay}
\\[0.55em]

\actscore{0.45} &
\texttt{(AP) -- An investigation has found that former Portland Police }
\dualhl{Chief}
\texttt{ Larry O'Dea brought discredit to the city,}
\\[0.55em]

\actscore{0.43} &
\texttt{Faulconer said the }
\dualhl{mayor}
\texttt{ spoke with San Diego  }
\dualhl{Police Chief}
\texttt{ Shelley Zimmerman who said the claims were unfounded}
\\[0.55em]

\actscore{0.42} &
\texttt{These are the trends a rational person should expect following FCC}
\dualhl{ Chairman}
\texttt{ Pai's holiday gifts to the biggest ISPs.}
\\[0.55em]

\actscore{0.38} &
\texttt{judicial killings, forced labor, and torture,'' }
\dualhl{Acting}
\texttt{ }
\dualhl{Undersecretary}
\texttt{ for Terrorism and}

\end{tabularx}

\vspace{0.75em}

\textbf{\textcolor{gray!70}{MODERATELY ACTIVATING (3)}}

\vspace{0.3em}

\begin{tabularx}{\linewidth}{@{}p{0.10\linewidth}X@{}}

\actscore{0.23} &
\texttt{land Basin,'' said Samuel Moody, Rockhopper's }
\dualhl{CEO}
\texttt{. Related: Christie's White House Ambitions}
\\[0.55em]

\actscore{0.20} &
\texttt{Interview with Sundance DiGiovanni (}
\dualhl{CEO}
\texttt{ MLG) Interview with Liquid Haypr}
\\[0.55em]

\actscore{0.19} &
\texttt{also noted a potential conflict of interest where PC Gamer }
\dualhl{executive editor}
\texttt{ Tyler Wilde was also roommates with an employee of}

\end{tabularx}

\end{tcolorbox}
\end{minipage}
\hfill
\begin{minipage}[t]{0.485\textwidth}
\begin{tcolorbox}[
    colback=white,
    colframe=gray!25,
    boxrule=0.6pt,
    arc=2mm,
    left=3mm,
    right=3mm,
    top=2mm,
    bottom=2mm
]

\textbf{\large JumpReLU}
\textcolor{gray!70}{latent 3533}

\vspace{0.75em}

\textbf{\textcolor{gray!70}{STRONGLY ACTIVATING (5)}}

\vspace{0.3em}

\begin{tabularx}{\linewidth}{@{}p{0.10\linewidth}X@{}}

\actscore{0.46} &
\texttt{(AP) -- An investigation has found that former Portland }
\missbl{Police}
\basehl{Chief}
\texttt{ Larry O'Dea brought discredit to the city,}
\\[0.55em]

\actscore{0.45} &
\texttt{judicial killings, forced labor, and torture,'' }
\dualhl{Acting}
\missbl{Undersecretary}
\\[0.55em]

\actscore{0.44} &
\texttt{Faulconer said the }
\missbl{mayor}
\texttt{ spoke with San Diego }
\missbl{Police}
\texttt{ }
\basehl{Chief}
\texttt{ Shelley Zimmerman who said the claims were unfounded}
\\[0.55em]

\actscore{0.44} &
\texttt{At a Monday news conference, White House deputy}
\texttt{ national }
\basehl{security}
\texttt{ }
\missbl{adviser}
\texttt{ Ben Rhodes}
\texttt{ encouraged the would-be alliance members to ``stay}
\\[0.55em]

\actscore{0.39} &
\texttt{a public statement regarding disclosure PC Gamer }
\basehl{executive editor}
\texttt{ Tyler Wilde is in a personal relationship with a Ubiso}

\end{tabularx}

\vspace{0.75em}

\textbf{\textcolor{gray!70}{MODERATELY ACTIVATING (3)}}

\vspace{0.3em}

\begin{tabularx}{\linewidth}{@{}p{0.10\linewidth}X@{}}

\actscore{0.22} &
\texttt{water Horizon. Critics say former }
\missbl{chief}
\texttt{ }
\basehl{executive}
\texttt{ }
\missbl{Lord}
\texttt{ Browne created all these disasters with relentless cost-}
\\[0.55em]

\actscore{0.19} &
\texttt{AND IRRESPONSIBLE! Ilford North }
\basehl{MP}
\texttt{ }
\missbl{Mr}
\texttt{ Streeting said the public would conclude Mr Corbyn was ``}
\\[0.55em]

\actscore{0.12} &
\texttt{ops'' for his miscues during the presidential debates, and }
\basehl{Gov}
\texttt{ Jerry Brown dismissed the Texan's recruiting drive as ``}

\end{tabularx}

\end{tcolorbox}
\end{minipage}

\caption{\textbf{Qualitative comparison of ``professional title'' latents learned by Dual-JumpReLU and standard JumpReLU.}
Both latents respond to professional and organizational titles such as \textit{Chief}, \textit{CEO}, \textit{Chairman}, and \textit{editor}. Dual-JumpReLU activates more consistently on complete title-related spans, whereas the standard JumpReLU latent also responds to surrounding or semantically different tokens, highlighted in gray. The moderately activating examples further reveal this secondary behavior, suggesting that DualSAE yields a more semantically focused and less entangled latent representation.}
\label{fig:text_professional_title_latent}

\end{figure*}

\begin{figure*}[t]
\centering
\small

\textbf{\large ``End-of-Text'' latents}

\vspace{0.6em}
\begin{minipage}[t]{0.485\textwidth}
\begin{tcolorbox}[
    colback=white,
    colframe=gray!25,
    boxrule=0.6pt,
    arc=2mm,
    left=3mm,
    right=3mm,
    top=2mm,
    bottom=2mm
]

\textbf{\large Dual-GatedSAE}
\textcolor{gray!70}{latent 7795}

\vspace{0.75em}

\textbf{\textcolor{gray!70}{STRONGLY ACTIVATING (5)}}

\vspace{0.3em}

\begin{tabularx}{\linewidth}{@{}p{0.10\linewidth}X@{}}

\actscore{0.45} &
\texttt{actualResult->getList()); \} \}}
\newline
\dualhl{<|endoftext|>}
\dualhl{package}
\texttt{ com.winterwell.depot;}
\\[0.55em]

\actscore{0.44} &
\texttt{\} \} \} \#endif}
\newline
\dualhl{<|endoftext|>}
\texttt{package}
\texttt{ zorn}
\texttt{co}
\texttt{.reploid}
\texttt{craft}
\texttt{.}
\texttt{network}
\texttt{;}
\\[0.55em]

\actscore{0.44} &
\texttt{NAGatedVC : UIViewController @end}
\newline
\dualhl{<|endoftext|>}
\texttt{using System; using System.Collections.Generic}
\\[0.55em]

\actscore{0.42} &
\texttt{uint16\_t Delay; \#endif}
\newline
\dualhl{<|endoftext|>}
\texttt{package}
\texttt{info}
\texttt{.bati11.otameshi.db}
\\[0.55em]

\actscore{0.38} &
\texttt{\}, posts: posts \}); \}); }
\dualhl{<|endoftext|>}
\texttt{/*}
\newline
\texttt{* Copyright (c) 2008--2018 SL}

\end{tabularx}

\vspace{0.75em}

\textbf{\textcolor{gray!70}{MODERATELY ACTIVATING (3)}}

\vspace{0.3em}

\begin{tabularx}{\linewidth}{@{}p{0.10\linewidth}X@{}}

\actscore{0.12} &
\texttt{AGatedVC : UIViewController @end }
\dualhl{<|endoftext|>}
\texttt{using System; using System.Collections.Generic;}
\\[0.55em]

\actscore{0.12} &
\texttt{> </body> </html> }
\dualhl{<|endoftext|>}
\texttt{using System.Reflection; using System.Runtime.Compiler}
\\[0.55em]

\actscore{0.12} &
\texttt{</body> </html> }
\dualhl{<|endoftext|>}
\texttt{---}
\texttt{status}
\texttt{:}
\texttt{publish}

\end{tabularx}

\end{tcolorbox}
\end{minipage}
\hfill
\begin{minipage}[t]{0.485\textwidth}
\begin{tcolorbox}[
    colback=white,
    colframe=gray!25,
    boxrule=0.6pt,
    arc=2mm,
    left=3mm,
    right=3mm,
    top=2mm,
    bottom=2mm
]

\textbf{\large GatedSAE}
\textcolor{gray!70}{latent 3113}

\vspace{0.75em}

\textbf{\textcolor{gray!70}{STRONGLY ACTIVATING (5)}}

\vspace{0.3em}

\begin{tabularx}{\linewidth}{@{}p{0.10\linewidth}X@{}}

\actscore{0.45} &
\texttt{; double y; }
\missbl{\};}
\missbl{ \#endif}
\basehl{<|endoftext|>}
\missbl{---}
\texttt{ title}
\\[0.55em]

\actscore{0.44} &
\texttt{Ctrl.quote); module.exports = router;}
\basehl{<|endoftext|>}
\texttt{<?php /**}
\newline
\texttt{* Magento * *}
\\[0.55em]

\actscore{0.44} &
\texttt{port=Config.PORT, host=Config.HOST )}
\newline
\basehl{<|endoftext|>}
\newline
\texttt{** Form generated from reading UI file '}
\\[0.55em]

\actscore{0.40} &
\texttt{registerTask('default', ['uglify']); }
\missbl{\};}
\newline
\basehl{<|endoftext|>}
\texttt{use strict'; angular}
\missbl{:}
\texttt{module('adsApp}
\\[0.55em]

\actscore{0.40} &
\texttt{accessor :api\_key \ \ end }
\missbl{end}
\newline
\basehl{<|endoftext|>}
\texttt{\# Xamarin.Plugin.AudioState The}

\end{tabularx}

\vspace{0.75em}

\textbf{\textcolor{gray!70}{MODERATELY ACTIVATING (3)}}
\textcolor{gray!70}{ --- WHERE A SECOND SENSE SHOWS UP}

\vspace{0.3em}

\begin{tabularx}{\linewidth}{@{}p{0.10\linewidth}X@{}}

\actscore{0.23} &
\texttt{- 1)); \} }
\basehl{<|endoftext|>}
\texttt{require 'basalt/packages/repo'}
\newline
\texttt{require '}
\\[0.55em]

\actscore{0.22} &
\texttt{cg\_sfun.h'' \#endif }
\basehl{<|endoftext|>}
\missbl{---}
\texttt{ layout: page-disqus}
\newline
\texttt{title:}
\\[0.55em]

\actscore{0.17} &
\texttt{width: \ \ height:}
\missbl{:}
\missbl{---}
\basehl{<|endoftext|>}
\texttt{\# encoding: utf-8}
\newline
\texttt{require File.expand}

\end{tabularx}

\end{tcolorbox}
\end{minipage}

\caption{\textbf{Qualitative comparison of ``end-of-text'' latents learned by Dual-GatedSAE and standard GatedSAE.}
Both latents respond strongly to the \dualhlcap{\texttt{<|endoftext|>}} token at boundaries between code fragments or files. Dual-GatedSAE exhibits a cleaner and more consistent association with the boundary token, whereas the standard GatedSAE latent also activates on \missblcap{surrounding punctuation and formatting-related tokens}, highlighted in gray. The moderately activating examples further reveal this secondary behavior, suggesting that DualSAE yields a more semantically focused and less entangled boundary feature.}
\label{fig:text_endoftext_latent}

\end{figure*}

\begin{figure*}[t]
\centering
\small

\textbf{\large ``class Attribute'' latents}

\vspace{0.6em}

\begin{minipage}[t]{0.485\textwidth}
\begin{tcolorbox}[
    colback=white,
    colframe=gray!25,
    boxrule=0.6pt,
    arc=2mm,
    left=3mm,
    right=3mm,
    top=2mm,
    bottom=2mm
]

\textbf{\large Dual-MatryoshkaSAE}
\textcolor{gray!70}{latent 1106}

\vspace{0.75em}

\textbf{\textcolor{gray!70}{STRONGLY ACTIVATING (5)}}

\vspace{0.3em}

\begin{tabularx}{\linewidth}{@{}p{0.10\linewidth}X@{}}

\actscore{0.34} &
\texttt{/X5eoF"}
\texttt{target}
\texttt{="\_blank"><span }
\dualhl{class}
\texttt{="icon-right-open">}
\newline
\texttt{</span> View in map}
\\[0.55em]

\actscore{0.33} &
\texttt{</div> \ \ <!-- \ \ <div><a }
\dualhl{class}
\texttt{="picture-cta button button-lrg alert"}
\\[0.55em]

\actscore{0.33} &
\texttt{Codemotion</a> \ \ <div><span }
\dualhl{class}
\texttt{="picture-where">Madrid / Spain</span>}
\\[0.55em]

\actscore{0.30} &
\texttt{div }
\dualhl{class}
\texttt{="row"> \ \ <!-- \ \ <div }
\dualhl{class}
\texttt{="columns">}
\newline
\texttt{<h1 }
\texttt{style}
\texttt{="clear:}
\\[0.55em]

\actscore{0.30} &
\texttt{ui3-u first-col"> \ \ <div }
\dualhl{class}
\texttt{="content">}
\newline
\texttt{</div> \ \ </div}

\end{tabularx}

\vspace{0.75em}

\textbf{\textcolor{gray!70}{MODERATELY ACTIVATING (3)}}

\vspace{0.3em}

\begin{tabularx}{\linewidth}{@{}p{0.10\linewidth}X@{}}

\actscore{0.15} &
\texttt{preserve\textbackslash"> <path }
\texttt{fill}
\texttt{=\textbackslash"\#AAAAAA\textbackslash" }
\missbl{d}
\texttt{=\textbackslash"M15.1,4.3c-2}
\\[0.55em]

\actscore{0.15} &
\texttt{<span }
\basehl{class}
\texttt{='break'> </span>memory<span }
\basehl{class}
\texttt{='break'> </span>.<span }
\basehl{class}
\texttt{='break'>}
\\[0.55em]

\actscore{0.14} &
\texttt{<span }
\basehl{class}
\texttt{='break'> </span>widget<span }
\basehl{class}
\texttt{='break'> </span>.<span }
\basehl{class}
\texttt{='break'>}

\end{tabularx}

\end{tcolorbox}
\end{minipage}
\hfill
\begin{minipage}[t]{0.485\textwidth}
\begin{tcolorbox}[
    colback=white,
    colframe=gray!25,
    boxrule=0.6pt,
    arc=2mm,
    left=3mm,
    right=3mm,
    top=2mm,
    bottom=2mm
]

\textbf{\large MatryoshkaSAE}
\textcolor{gray!70}{latent 1893}

\vspace{0.75em}

\textbf{\textcolor{gray!70}{STRONGLY ACTIVATING (5)}}

\vspace{0.3em}

\begin{tabularx}{\linewidth}{@{}p{0.10\linewidth}X@{}}

\actscore{0.46} &
\texttt{<div }
\basehl{class}
\texttt{="picture-caption"> \ \ <a }
\basehl{class}
\texttt{="bottom-logo" }
\missbl{href}
\texttt{="/">Codemotion</}
\\[0.55em]

\actscore{0.42} &
\texttt{<link }
\missbl{href}
\texttt{="http://example.com/1" }
\missbl{rel}
\texttt{="alternate" }
\missbl{type}
\texttt{="application/atom+xml"/>}
\\[0.55em]

\actscore{0.42} &
\texttt{<link }
\missbl{href}
\texttt{="http://example.com/3" }
\missbl{rel}
\texttt{="alternate" }
\missbl{type}
\texttt{="application/rss+xml"/>}
\\[0.55em]

\actscore{0.40} &
\texttt{</div> \ \ <!-- \ \ <div><a }
\basehl{class}
\texttt{="picture-cta button button-lrg alert"}
\\[0.55em]

\actscore{0.39} &
\texttt{<link }
\missbl{href}
\texttt{="http://example.com/1" }
\missbl{rel}
\texttt{="alternate" }
\missbl{type}
\texttt{="application/atom+xml"}

\end{tabularx}

\vspace{0.75em}

\textbf{\textcolor{gray!70}{MODERATELY ACTIVATING (3)}}

\vspace{0.3em}

\begin{tabularx}{\linewidth}{@{}p{0.10\linewidth}X@{}}

\actscore{0.14} &
\texttt{</a></li> <li><a }
\missbl{href}
\texttt{="../symbols/}
\\[0.55em]

\actscore{0.13} &
\texttt{a></li> <li><a }
\missbl{href}
\texttt{="../symbols/}
\\[0.55em]

\actscore{0.12} &
\texttt{</a></li> <li><a }
\missbl{href}
\texttt{="../symbols/}

\end{tabularx}

\end{tcolorbox}
\end{minipage}

\caption{\textbf{Qualitative comparison of ``class attribute'' latents learned by Dual-MatryoshkaSAE and standard MatryoshkaSAE.}
Both latents respond to HTML attributes, but Dual-MatryoshkaSAE is more selectively aligned with the \dualhlcap{\texttt{class}} attribute across diverse contexts. In contrast, the standard MatryoshkaSAE latent also activates on other attributes such as \missblcap{\texttt{href}}, \missblcap{\texttt{rel}}, and \missblcap{\texttt{type}}, highlighted in gray. The moderately activating examples show the same pattern, with Dual-MatryoshkaSAE continuing to focus on \texttt{class} while the baseline shifts toward other attribute types, suggesting a cleaner and less entangled semantic feature.}
\label{fig:text_class_attribute_latent}

\end{figure*}

\FloatBarrier
\section{Downstream applications}
\label{app:downstream}

    \FloatBarrier
    \subsection{SAE as a topic model}
    \label{sec:topic_model}

    \begin{table*}[t]
\centering
\caption{\textbf{Topic modeling results} on ImageNet-1K and OpenImagesV7.
We compare SAE variants with different activation functions and encoders across topic counts $K \in \{50,100,200,300,500\}$.
Diversity $D$ evaluates topic non-redundancy, and intruder accuracy $C_I$ measures whether unrelated concepts can be identified from each topic.
DualSAE increases topic diversity in 89 of the 90 configurations and improves intruder accuracy in most of them. IN = ImageNet-1K, OI = OpenImagesV7, SL2 = SigLIP2, TK = TopK, BTK = BatchTopK, JR = JumpReLU.}
\label{tab:topic_modeling}
\scriptsize
\resizebox{0.8\textwidth}{!}{
\begin{tabular}{lccccc|ccccc}
\toprule
\textbf{SAE / Enc.}
& \multicolumn{5}{c|}{\textbf{IN}}
& \multicolumn{5}{c}{\textbf{OI}} \\
\cmidrule(lr){2-6} \cmidrule(lr){7-11}
& \textbf{K=50} & \textbf{K=100} & \textbf{K=200} & \textbf{K=300} & \textbf{K=500}
& \textbf{K=50} & \textbf{K=100} & \textbf{K=200} & \textbf{K=300} & \textbf{K=500} \\
\midrule

\multicolumn{11}{l}{\textbf{Diversity $D$}} \\
TK / CLIP        & \textbf{+0.11} & \textbf{+0.10} & \textbf{+0.10} & \textbf{+0.11} & \textbf{+0.10} & \textbf{+0.05} & \textbf{+0.03} & \textbf{+0.02} & \textbf{+0.04} & \textbf{+0.05} \\
TK / SL2      & -0.01 & \textbf{+0.00} & \textbf{+0.03} & \textbf{+0.06} & \textbf{+0.12} & \textbf{+0.04} & \textbf{+0.03} & \textbf{+0.03} & \textbf{+0.05} & \textbf{+0.08} \\
TK / ViT         & \textbf{+0.10} & \textbf{+0.08} & \textbf{+0.06} & \textbf{+0.06} & \textbf{+0.03} & \textbf{+0.05} & \textbf{+0.04} & \textbf{+0.03} & \textbf{+0.02} & \textbf{+0.03} \\
BTK / CLIP   & \textbf{+0.17} & \textbf{+0.16} & \textbf{+0.16} & \textbf{+0.16} & \textbf{+0.17} & \textbf{+0.12} & \textbf{+0.09} & \textbf{+0.10} & \textbf{+0.10} & \textbf{+0.13} \\
BTK / SL2 & \textbf{+0.07} & \textbf{+0.09} & \textbf{+0.11} & \textbf{+0.15} & \textbf{+0.20} & \textbf{+0.10} & \textbf{+0.08} & \textbf{+0.08} & \textbf{+0.12} & \textbf{+0.16} \\
BTK / ViT    & \textbf{+0.05} & \textbf{+0.06} & \textbf{+0.04} & \textbf{+0.04} & \textbf{+0.03} & \textbf{+0.04} & \textbf{+0.03} & \textbf{+0.04} & \textbf{+0.03} & \textbf{+0.04} \\
JR / CLIP    & \textbf{+0.13} & \textbf{+0.12} & \textbf{+0.09} & \textbf{+0.07} & \textbf{+0.09} & \textbf{+0.12} & \textbf{+0.12} & \textbf{+0.10} & \textbf{+0.10} & \textbf{+0.11} \\
JR / SL2  & \textbf{+0.07} & \textbf{+0.06} & \textbf{+0.04} & \textbf{+0.04} & \textbf{+0.05} & \textbf{+0.15} & \textbf{+0.14} & \textbf{+0.11} & \textbf{+0.10} & \textbf{+0.11} \\
JR / ViT     & \textbf{+0.11} & \textbf{+0.10} & \textbf{+0.08} & \textbf{+0.06} & \textbf{+0.05} & \textbf{+0.18} & \textbf{+0.17} & \textbf{+0.14} & \textbf{+0.13} & \textbf{+0.14} \\

\midrule
\multicolumn{11}{l}{\textbf{Intruder $C_I$ (\%)}} \\
TK / CLIP        & \textbf{+5.11} & \textbf{+5.22} & \textbf{+0.44} & \textbf{+2.48} & \textbf{+1.31} & \textbf{+5.11} & -2.11 & \textbf{+1.28} & -1.44 & -0.07 \\
TK / SL2      & \textbf{+0.89} & -0.89 & -3.50 & \textbf{+0.56} & \textbf{+2.27} & \textbf{+10.44} & \textbf{+5.89} & \textbf{+2.28} & \textbf{+1.26} & \textbf{+2.87} \\
TK / ViT         & \textbf{+9.11} & \textbf{+5.33} & \textbf{+3.50} & \textbf{+0.59} & \textbf{+2.04} & \textbf{+2.00} & \textbf{+7.44} & \textbf{+1.22} & \textbf{+1.78} & \textbf{+1.49} \\
BTK / CLIP   & \textbf{+7.78} & \textbf{+0.78} & \textbf{+0.17} & \textbf{+3.26} & \textbf{+3.29} & \textbf{+5.56} & \textbf{+4.11} & \textbf{+3.00} & \textbf{+1.41} & \textbf{+3.04} \\
BTK / SL2 & -4.44 & -2.00 & \textbf{+0.06} & \textbf{+1.59} & \textbf{+1.93} & \textbf{+3.56} & \textbf{+3.56} & \textbf{+0.28} & \textbf{+1.78} & \textbf{+1.44} \\
BTK / ViT    & \textbf{+3.78} & \textbf{+5.89} & \textbf{+1.11} & \textbf{+0.89} & \textbf{+2.53} & \textbf{+8.22} & \textbf{+4.22} & \textbf{+1.94} & \textbf{+0.52} & \textbf{+2.51} \\
JR / CLIP    & \textbf{+1.33} & \textbf{+3.33} & \textbf{+0.89} & \textbf{+0.04} & \textbf{+1.31} & \textbf{+0.89} & -0.78 & -0.67 & \textbf{+1.63} & \textbf{+0.11} \\
JR / SL2  & \textbf{+4.67} & \textbf{+6.56} & \textbf{+1.39} & \textbf{+1.63} & \textbf{+3.84} & \textbf{+2.00} & \textbf{+5.11} & \textbf{+3.22} & \textbf{+1.52} & \textbf{+2.64} \\
JR / ViT     & -0.67 & -0.11 & \textbf{+1.22} & \textbf{+2.96} & \textbf{+1.78} & -3.33 & -5.22 & -9.06 & -7.07 & -5.71 \\

\bottomrule
\end{tabular}
}
\end{table*}

    \textbf{Description.}
    We assess whether DualSAE's Sparse Branch yields better \emph{topic models} than
    conventional SAEs. Following the SAE-TM pipeline~\citep{girrbach2026saetm}, we interpret each dictionary
    feature with a distribution over caption vocabulary, cluster the features into $K$ topics, and
    summarize every topic by its highest-probability words. We report two complementary metrics.
    \emph{Diversity} $D$ measures topic non-redundancy (the semantic spread of the top words across
    topics)
    and \emph{intruder accuracy} $C_I$ measures whether an unrelated ``intruder'' concept injected into a
    topic can be correctly identified; higher values indicate a more self-consistent topic. For both metrics, higher is better. To isolate the effect of the two-branch decomposition, we compare
    each DualSAE only against its \emph{activation-matched} conventional baseline
    (TopK\,$\leftrightarrow$\,Dual-TopK, BatchTopK\,$\leftrightarrow$\,Dual-BatchTopK,
    JumpReLU\,$\leftrightarrow$\,Dual-JumpReLU) and report the difference
    $\Delta = \text{DualSAE} - \text{baseline}$; positive entries in Table~\ref{tab:topic_modeling} thus
    denote improvements attributable to DualSAE.

    \textbf{Settings.}
    We evaluate on two datasets, ImageNet-1K and OpenImagesV7, and three vision encoders: CLIP
    (ViT-L/14, DataComp), SigLIP2 (SoViT-400m/14 at $384$px), and a supervised ViT-L/16 (at $224$px).
    For each encoder we train SAEs with three sparsity mechanisms---TopK, BatchTopK, and JumpReLU---and
    their DualSAE counterparts, all with a dictionary of $8192$ atoms and matched sparsity budget
    ($k{=}64$). Topics are formed for $K \in \{50, 100, 200, 300, 500\}$. Intruder scores
    are produced by the same LLM judge for every method, and all reported values are averaged over three
    random seeds; we never average across $K$, encoder, or activation, reporting every configuration
    separately.

    \textbf{Diversity.}
    DualSAE improves topic diversity almost universally: $\Delta D$ is positive in nearly every one of the
    $90$ configurations, the sole exception being TopK/SigLIP2 at $K{=}50$ on ImageNet-1K ($-0.01$). The gains
    are stable across $K$ and both datasets, and are largest for BatchTopK/CLIP (consistently $\approx
    +0.16$). This indicates that decorrelating the Sparse Branch and Residual Branch removes
    redundant directions, yielding topics whose vocabularies overlap less.

    \textbf{Intruder accuracy.}
    Intruder accuracy is predominantly improved---DualSAE topics are more
    often self-consistent enough for the injected concept to be spotted. Gains are especially large for
    TopK/ViT on ImageNet-1K ($+9.11$ at $K{=}50$) and are positive for most configurations across both
    datasets. The same weak spot recurs: JumpReLU/ViT on OpenImagesV7 degrades ($-3$ to $-9$), and a few
    small-$K$ BatchTopK/SigLIP2 cells are mildly negative.

\begin{figure*}[t]
      \centering
      \includegraphics[width=\linewidth,height=0.9\textheight,keepaspectratio]{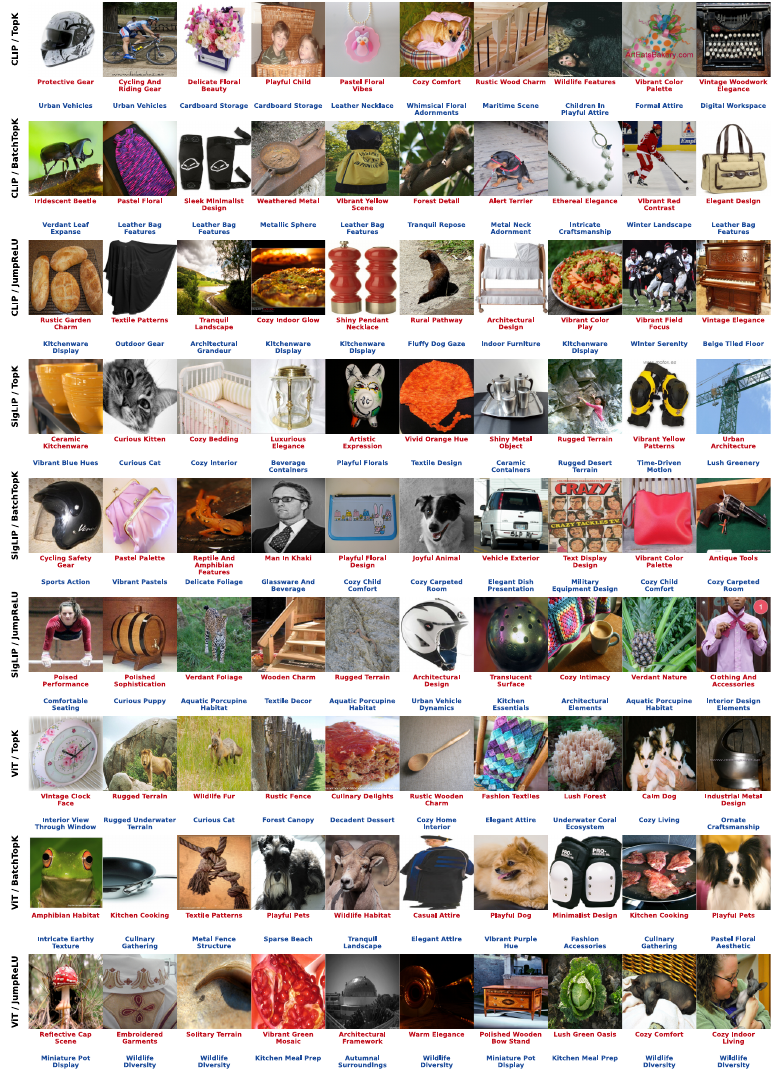}
        \caption{\textbf{Image--topic assignments between DualSAE and traditional SAEs} (one row
per encoder/activation, $K{=}100$). Labels show each image's dominant topic under DualSAE
(\textcolor{red}{{red}}) and the activation-matched baseline (\textcolor{blue}{{blue}});
topic names are LLM summaries of top topic words. DualSAE yields concrete, object-specific topics, while the baseline entangles objects with context or abstract style.}
        \label{fig:app_img2topic}
    \end{figure*}

    \textbf{Qualitative results.} Figure~\ref{fig:app_img2topic} qualitatively compares image--topic assignments from DualSAE and activation-matched traditional SAEs across multiple encoder and sparsification settings. A consistent pattern is that DualSAE produces more concrete topics that closely describe the dominant visual content of each image, whereas baseline SAE topics are often driven by incidental context, broad scene semantics, or abstract appearance. For example, images containing a helmet, beetle, bread, and frog are assigned by DualSAE to topics such as \emph{Protective Gear}, \emph{Iridescent Beetle}, \emph{Rustic Garden Charm}, and \emph{Amphibian Habitat}, while the corresponding baseline topics include less directly related concepts such as \emph{Urban Vehicles}, \emph{Verdant Leaf Expanse}, \emph{Kitchenware Display}, and \emph{Intricate Earthy Texture}. Similar behavior appears across CLIP and ViT encoders and different activation functions, suggesting that the improvement is not restricted to a particular SAE configuration. Overall, these examples indicate that DualSAE concentrates the Sparse Branch on salient, semantically coherent visual concepts, leading to topic assignments that are more directly grounded in image content.

\subsection{SAE for hypothesis generation}
\label{sec:hypo_sae}

\textbf{Description.}
We assess whether DualSAE's Sparse Branch yields better \textit{natural-language
hypotheses} than conventional SAEs. Following the HypotheSAEs
pipeline~\citep{movva2025sparse}, adapted to vision, we (i) select, for every class, the
$K$ dictionary features most discriminative of that class, (ii) ask a VLM to describe each
selected feature as a short natural-language hypothesis (``why the classifier predicts this
class''), and (iii) score how faithfully those hypotheses explain the class on held-out
images. We report two complementary metrics. \textit{AUROC} measures the predictive power of
a hypothesis set: a VLM annotates whether each held-out image satisfies each hypothesis, and
we fit a logistic model that predicts class membership from these binary annotations.
\textit{\#Significant} counts how many of the $K$ hypotheses remain significantly associated
with the class after Bonferroni correction (family-wise $\alpha{=}0.1$) — \ie, how many
individually meaningful explanations the model surfaces. For both metrics, higher is better.
We isolate the effect of the two-branch decomposition by comparing each
DualSAE only against its \textit{activation-matched} conventional baseline and report
$\Delta = \text{DualSAE} - \text{baseline}$; positive entries in Table~\ref{tab:hypogen}
denote improvements attributable to DualSAE.

\textbf{Settings.}
Datasets, encoders, and SAE configurations are identical to those used in the other experiments (ImageNet-1K and
OpenImagesV7; CLIP ViT-L/14, SigLIP2, supervised ViT-L/16; dictionaries of 8192 atoms at a
matched sparsity budget $k{=}64$). For each class we select $K{=}16$ latents; hypotheses are
proposed by InternVL2.5-8B from each latent's 8 top-activating and 8 non-activating training
images, and the same VLM produces the binary faithfulness annotations for every method.
Faithfulness is evaluated one-vs-rest on 100 randomly sampled classes per dataset (for the
multi-label OpenImages, positives are images containing the class, subsampled to at most 200
per class), and all values are macro-averaged over classes and averaged over three random
seeds; we never average across encoder, activation, or dataset, reporting every configuration
separately. Because the optimal latent-selection rule depends on the sparsity mechanism —
$\ell_1$-regularised selection systematically under-selects informative latents from the
dense JumpReLU code, depressing absolute faithfulness for \emph{both} models — we sweep three
selection rules (LASSO, correlation, separation score) and report each configuration using its
best-performing rule, always applying the same rule to DualSAE and its baseline so the
comparison remains matched.

\textbf{Significant hypotheses.}
DualSAE surfaces more significant hypotheses almost universally: $\Delta$\#Significant is
positive in 17 of the 18 encoder/activation/dataset configurations, the sole exception being
TopK/ViT on ImageNet-1K ($-0.05$, effectively parity). Gains are largest for BatchTopK/SigLIP on
ImageNet-1K ($+0.52$), TopK/SigLIP on OpenImages-V7 ($+0.39$), and JumpReLU/ViT on ImageNet-1K
($+0.33$), and are consistent across both datasets. This indicates that the Sparse Branch yields features whose natural-language descriptions are individually predictive of the
class, rather than diffusing class evidence across many weakly associated directions.

\textbf{Predictive power.}
AUROC differences are small in the near-saturated TopK/BatchTopK regimes (baseline AUROC
$\approx 0.93$--$0.99$), where DualSAE is at parity or slightly ahead in most configurations
(every deficit is $\leq 0.005$). The clearest AUROC gains appear for JumpReLU, the hardest
setting for both models: DualSAE improves AUROC in all six JumpReLU configurations, by up to
$+0.024$ (JumpReLU/ViT on ImageNet-1K).

\begin{table}[t]
\centering

\scriptsize
\caption{\textbf{Hypothesis-generation results} for SAE variants across ImageNet-1K and OpenImagesV7.
We report only the performance differences between the DualSAE and baseline models.
$\Delta$AUROC measures the change in predictive quality of generated hypotheses, while
$\Delta$\#Significant measures the change in the average number of statistically significant hypotheses.
DualSAE yields more significant hypotheses in 17 of the 18 configurations and higher AUROC in all six JumpReLU configurations; every AUROC deficit is at most 0.005. IN = ImageNet-1K, OI = OpenImagesV7, SL2 = SigLIP2, TK = TopK, BTK = BatchTopK, JR = JumpReLU.}
\label{tab:hypogen}
\begingroup
\setlength{\tabcolsep}{3pt}
\resizebox{\linewidth}{!}{%
\begin{tabular}{llcccc}
\toprule
\multirow{2}{*}{\textbf{Enc.}} &
\multirow{2}{*}{\textbf{SAE}} &
\multicolumn{2}{c}{\textbf{IN}} &
\multicolumn{2}{c}{\textbf{OI}} \\
\cmidrule(lr){3-4} \cmidrule(lr){5-6}
& & $\Delta$AUROC & $\Delta$\#Significant
  & $\Delta$AUROC & $\Delta$\#Significant \\
\midrule

\multirow{3}{*}{CLIP}
& TK        & \textbf{+0.001} & \textbf{+0.11} & \textbf{+0.002} & \textbf{+0.31} \\
& BTK & -0.003          & \textbf{+0.12} & -0.001          & \textbf{+0.07} \\
& JR  & \textbf{+0.011} & \textbf{+0.31} & \textbf{+0.010} & \textbf{+0.15} \\
\midrule

\multirow{3}{*}{SL2}
& TK        & \textbf{+0.001} & \textbf{+0.01} & +0.000          & \textbf{+0.39} \\
& BTK & \textbf{+0.005} & \textbf{+0.52} & -0.005          & \textbf{+0.25} \\
& JR  & \textbf{+0.010} & \textbf{+0.22} & \textbf{+0.006} & \textbf{+0.15} \\
\midrule

\multirow{3}{*}{ViT}
& TK        & -0.002          & -0.05          & +0.000          & \textbf{+0.19} \\
& BTK & \textbf{+0.002} & \textbf{+0.26} & \textbf{+0.005} & \textbf{+0.19} \\
& JR  & \textbf{+0.024} & \textbf{+0.33} & \textbf{+0.012} & \textbf{+0.12} \\
\bottomrule
\end{tabular}%
}
\endgroup
\end{table}

\textbf{Qualitative results.} Figure~\ref{fig:app_hypo_SAE} provides a qualitative example of class-level hypothesis generation for the \emph{prayer rug} class. The baseline TopK SAE selects only partially relevant or polysemantic latents: one focuses on generic intricate rug patterns, while another responds primarily to large mosque structures rather than the target object itself. In contrast, Dual-TopK identifies more class-specific concepts, including \emph{red rugs with intricate patterns} and \emph{bold, colorful rugs with intricate patterns}, whose highest-activating images consistently exhibit these visual attributes. Together with Figure~\ref{fig:hypo_main} of the main paper, this example suggests that separating the semantic sparse representation from residual information enables DualSAE to produce more coherent and class-relevant latents, rather than mixing target concepts with correlated or unrelated visual context.

\begin{figure*}[t]
      \centering
      % prayer rug part only (the bull mastiff part below is cropped); same scale as the former full figure
      \includegraphics[trim=0 181.5 0 0,clip,width=0.72\linewidth]{Figures/hypo_SAE.pdf}
        \caption{\textbf{Class-level hypotheses for the \textit{prayer rug} class.} We show the top-2 selected latents for the baseline TopK SAE (red) and Dual-TopK (blue), each accompanied by its VLM-generated hypothesis and five highest-activating images. The TopK latents are only partially relevant (generic rug patterns) or off-topic (mosque structures), whereas the Dual-TopK latents capture class-relevant rug concepts.}
        \label{fig:app_hypo_SAE}
    \end{figure*}

\subsection{LLM steering}
\label{app:llm_steering}

\begin{table*}[t]
\centering
\caption{Quantitative LLM steering on Pythia (SAEs trained on OpenWebText).
The unit-norm decoder direction of each selected latent is added to the
residual stream at the block of its SAE, with a strength calibrated to a
next-token KL of 0.2 nats.
AxBench score: harmonic mean of the phi-4 concept, instruction, and fluency
ratings; $P(\text{concept})$: probability of the target concept from a
BGE-based classifier.
The standard SAE has a lower mean than DualSAE in all 80 comparisons and, after Benjamini--Hochberg correction, is significantly worse in 33 and significantly better in none. Std.\ = standard SAE, Dual = DualSAE, TK = TopK, BTK = BatchTopK,
AMZ = Amazon reviews, AGN = AG News, GoE = GoEmotions.
Bold marks the higher mean in each setting.
Results are reported as mean $\pm$ standard deviation across three seeds.}
\label{tab:text_steering_quant}
\footnotesize
\setlength{\tabcolsep}{3.5pt}
\begin{tabular}{@{}llcccccccc@{}}
\toprule
& & \multicolumn{2}{c}{Block 6, TK} & \multicolumn{2}{c}{Block 12, TK} & \multicolumn{2}{c}{Block 18, TK} & \multicolumn{2}{c}{Block 6, BTK} \\
\cmidrule(lr){3-4}\cmidrule(lr){5-6}\cmidrule(lr){7-8}\cmidrule(lr){9-10}
Data & Concept & Dual & Standard & Dual & Standard & Dual & Standard & Dual & Standard \\
\midrule
\multicolumn{10}{@{}l}{\textit{AxBench score (0--2) $\uparrow$}} \\
\multirow{2}{*}{AMZ} & Positive & \stdbf{0.194}{0.085} & \stdval{0.118}{0.019} & \stdbf{0.249}{0.031} & \stdval{0.139}{0.019} & \stdbf{0.218}{0.013} & \stdval{0.136}{0.050} & \stdbf{0.228}{0.031} & \stdval{0.107}{0.022} \\
 & Negative & \stdbf{0.062}{0.023} & \stdval{0.041}{0.034} & \stdbf{0.091}{0.026} & \stdval{0.053}{0.042} & \stdbf{0.072}{0.010} & \stdval{0.037}{0.016} & \stdbf{0.050}{0.015} & \stdval{0.028}{0.016} \\
\cmidrule(lr){1-10}
\multirow{4}{*}{AGN} & World & \stdbf{0.219}{0.037} & \stdval{0.083}{0.025} & \stdbf{0.327}{0.032} & \stdval{0.088}{0.028} & \stdbf{0.176}{0.063} & \stdval{0.085}{0.065} & \stdbf{0.236}{0.035} & \stdval{0.158}{0.027} \\
 & Sports & \stdbf{0.170}{0.037} & \stdval{0.040}{0.011} & \stdbf{0.278}{0.118} & \stdval{0.167}{0.123} & \stdbf{0.279}{0.073} & \stdval{0.114}{0.117} & \stdbf{0.303}{0.013} & \stdval{0.185}{0.097} \\
 & Business & \stdbf{0.191}{0.011} & \stdval{0.065}{0.030} & \stdbf{0.311}{0.021} & \stdval{0.128}{0.026} & \stdbf{0.257}{0.073} & \stdval{0.080}{0.032} & \stdbf{0.196}{0.085} & \stdval{0.088}{0.022} \\
 & Sci/Tech & \stdbf{0.290}{0.088} & \stdval{0.040}{0.013} & \stdbf{0.446}{0.029} & \stdval{0.121}{0.074} & \stdbf{0.340}{0.098} & \stdval{0.099}{0.054} & \stdbf{0.226}{0.030} & \stdval{0.155}{0.064} \\
\cmidrule(lr){1-10}
\multirow{4}{*}{GoE} & Joy & \stdbf{0.408}{0.134} & \stdval{0.175}{0.048} & \stdbf{0.533}{0.054} & \stdval{0.340}{0.214} & \stdbf{0.471}{0.105} & \stdval{0.184}{0.064} & \stdbf{0.373}{0.135} & \stdval{0.178}{0.081} \\
 & Anger & \stdbf{0.158}{0.044} & \stdval{0.050}{0.011} & \stdbf{0.280}{0.018} & \stdval{0.201}{0.111} & \stdbf{0.199}{0.054} & \stdval{0.053}{0.014} & \stdbf{0.118}{0.020} & \stdval{0.079}{0.039} \\
 & Sadness & \stdbf{0.350}{0.037} & \stdval{0.059}{0.006} & \stdbf{0.415}{0.081} & \stdval{0.101}{0.005} & \stdbf{0.266}{0.042} & \stdval{0.074}{0.013} & \stdbf{0.227}{0.058} & \stdval{0.093}{0.017} \\
 & Fear & \stdbf{0.353}{0.012} & \stdval{0.163}{0.035} & \stdbf{0.415}{0.006} & \stdval{0.195}{0.083} & \stdbf{0.295}{0.038} & \stdval{0.121}{0.031} & \stdbf{0.280}{0.017} & \stdval{0.151}{0.038} \\
\midrule
\multicolumn{10}{@{}l}{\textit{$P(\text{concept})$ (0--1) $\uparrow$}} \\
\multirow{2}{*}{AMZ} & Positive & \stdbf{0.840}{0.052} & \stdval{0.781}{0.014} & \stdbf{0.924}{0.009} & \stdval{0.765}{0.041} & \stdbf{0.882}{0.052} & \stdval{0.748}{0.033} & \stdbf{0.885}{0.040} & \stdval{0.795}{0.044} \\
 & Negative & \stdbf{0.401}{0.087} & \stdval{0.312}{0.037} & \stdbf{0.580}{0.063} & \stdval{0.397}{0.125} & \stdbf{0.544}{0.049} & \stdval{0.302}{0.026} & \stdbf{0.399}{0.040} & \stdval{0.385}{0.076} \\
\cmidrule(lr){1-10}
\multirow{4}{*}{AGN} & World & \stdbf{0.763}{0.057} & \stdval{0.682}{0.012} & \stdbf{0.848}{0.005} & \stdval{0.674}{0.052} & \stdbf{0.781}{0.006} & \stdval{0.667}{0.095} & \stdbf{0.790}{0.015} & \stdval{0.743}{0.015} \\
 & Sports & \stdbf{0.319}{0.046} & \stdval{0.127}{0.028} & \stdbf{0.453}{0.096} & \stdval{0.277}{0.161} & \stdbf{0.411}{0.105} & \stdval{0.205}{0.138} & \stdbf{0.466}{0.025} & \stdval{0.321}{0.123} \\
 & Business & \stdbf{0.475}{0.045} & \stdval{0.257}{0.034} & \stdbf{0.676}{0.006} & \stdval{0.405}{0.046} & \stdbf{0.557}{0.057} & \stdval{0.314}{0.057} & \stdbf{0.454}{0.005} & \stdval{0.345}{0.044} \\
 & Sci/Tech & \stdbf{0.673}{0.061} & \stdval{0.419}{0.010} & \stdbf{0.802}{0.037} & \stdval{0.521}{0.071} & \stdbf{0.669}{0.049} & \stdval{0.500}{0.050} & \stdbf{0.588}{0.055} & \stdval{0.542}{0.063} \\
\cmidrule(lr){1-10}
\multirow{4}{*}{GoE} & Joy & \stdbf{0.407}{0.091} & \stdval{0.216}{0.027} & \stdbf{0.435}{0.020} & \stdval{0.311}{0.100} & \stdbf{0.410}{0.071} & \stdval{0.231}{0.055} & \stdbf{0.371}{0.098} & \stdval{0.220}{0.046} \\
 & Anger & \stdbf{0.178}{0.046} & \stdval{0.071}{0.008} & \stdbf{0.296}{0.025} & \stdval{0.189}{0.111} & \stdbf{0.200}{0.045} & \stdval{0.079}{0.010} & \stdbf{0.144}{0.015} & \stdval{0.121}{0.027} \\
 & Sadness & \stdbf{0.426}{0.020} & \stdval{0.180}{0.035} & \stdbf{0.503}{0.048} & \stdval{0.203}{0.011} & \stdbf{0.344}{0.020} & \stdval{0.176}{0.017} & \stdbf{0.329}{0.006} & \stdval{0.229}{0.014} \\
 & Fear & \stdbf{0.236}{0.011} & \stdval{0.107}{0.018} & \stdbf{0.322}{0.040} & \stdval{0.126}{0.038} & \stdbf{0.208}{0.006} & \stdval{0.078}{0.004} & \stdbf{0.197}{0.016} & \stdval{0.122}{0.019} \\
\bottomrule
\end{tabular}
\end{table*}

\textbf{Quantitative steering evaluation.}
We also evaluate steering quantitatively following AxBench~\citep{wu2025axbench},
using SAEs trained on OpenWebText at blocks 6, 12, and 18 of Pythia with the
configuration of Supplementary~\ref{app:llm} (three seeds each; here, DualSAE uses a 1{,}250-step
warm-up and the TopK baseline has untied encoder and decoder weights).
We consider 10 concepts: positive (5-star) and negative (1-star) Amazon
reviews~\citep{hou2024bridging}; the World, Sports, Business, and Sci/Tech
classes of AG News~\citep{zhang2015character}; and joy, anger, sadness, and
fear from GoEmotions~\citep{demszky2020goemotions}.
For each SAE and concept, we select the three latents with the highest AUROC
for separating the concept's documents from those of the other classes of its
dataset (activations mean-pooled per document; up to 2{,}000 training
documents per class), excluding latents that fire on the first token in over
50\% of contexts or on over 30\% of tokens.
For DualSAE, latents come from the Sparse Branch.
\textit{Steering.}
We add $\alpha u$ to the residual stream at every position except the first,
where $u$ is the latent's unit-norm decoder direction, and calibrate $\alpha$
per latent so that the next-token KL divergence from the clean model on
held-out OpenWebText contexts is 0.2 nats; all SAEs thus perturb the model
equally.
This budget was fixed before any SAE was evaluated, as the smallest at which a
difference-in-means concept direction induces the concept (mean concept
rating $\geq 0.5$) while random directions do not ($\leq 0.2$).
We sample two 48-token continuations (temperature 0.7, top-$p$ 0.9) for each
of 60 neutral prompts without concept words, with the same random seeds for
all SAEs.
\textit{Metrics and statistics.}
The AxBench score is the harmonic mean of the concept, instruction, and
fluency ratings (each 0--2) from a phi-4 judge~\citep{abdin2024phi4} with the
official AxBench templates (0 if any rating is 0).
$P(\text{concept})$ is the target-concept probability of a logistic-regression
classifier on BGE-large embeddings~\citep{xiao2024cpack} trained on the
concept datasets.
Scores are averaged over prompts, continuations, and the three latents of a
concept, then reported as mean $\pm$ standard deviation across three seeds.
We compare each DualSAE with the standard SAE of the same activation by a
two-sided Welch $t$-test over seeds, with Benjamini--Hochberg
correction~\citep{benjamini1995controlling} over all 149 tests ($q<0.10$).
As shown in Table~\ref{tab:text_steering_quant}, standard SAE has a lower mean than
DualSAE in all 80 comparisons (4 settings $\times$ 10 concepts
$\times$ 2 metrics).
After correction, standard SAE is significantly worse in 33 and significantly
better in none; 3 are equivalent within margins of 0.10 (AxBench score) and
0.05 ($P(\text{concept})$), 38 are inconclusive, and 6 are excluded by the
outlier rule.

\textbf{Qualitative examples.} Figures~\ref{fig:text_steering_sadness}--\ref{fig:text_steering_anger} show qualitative steering examples for the \textit{sadness}, \textit{fear}, \textit{sports}, and \textit{anger} concepts, complementing Figure~\ref{fig:text_qualitative_combined}b of the main paper.

\begin{figure*}[t]
\centering
\small

\textbf{\large Steering toward ``Sadness''}
\hspace{0.3em}

\vspace{0.6em}

\begin{tcolorbox}[
    colback=gray!3,
    colframe=gray!25,
    boxrule=0.6pt,
    arc=2mm,
    left=3mm,
    right=3mm,
    top=2mm,
    bottom=2mm
]
\textbf{Prompt.}
\texttt{Yesterday I found an old photograph}
\end{tcolorbox}

\vspace{0.4em}

\begin{tcolorbox}[
    colback=white,
    colframe=gray!25,
    boxrule=0.6pt,
    arc=2mm,
    left=3mm,
    right=3mm,
    top=2mm,
    bottom=2mm
]

\textbf{\large Unsteered}

\vspace{0.35em}

\texttt{
Yesterday I found an old photograph of an elderly man.
He was in a wheelchair and I was taking pictures.
I found his name on the back of the photograph and when I tried to read it
I found out that he was my father.
}

\end{tcolorbox}

\vspace{0.4em}

\begin{tcolorbox}[
    colback=white,
    colframe=gray!25,
    boxrule=0.6pt,
    arc=2mm,
    left=3mm,
    right=3mm,
    top=2mm,
    bottom=2mm
]

\textbf{\large BatchTopK}

\vspace{0.35em}

\texttt{
Yesterday I found an old photograph of the church on the Internet
and I have decided to put it up.
}

\vspace{0.35em}

\texttt{Here is the original picture:}

\vspace{0.35em}

\texttt{
In the photograph there is a large cross and a cross-shaped figure,
I think it's a Roman cross.
}

\end{tcolorbox}

\vspace{0.4em}

\begin{tcolorbox}[
    colback=white,
    colframe=gray!25,
    boxrule=0.6pt,
    arc=2mm,
    left=3mm,
    right=3mm,
    top=2mm,
    bottom=2mm
]

\textbf{\large Dual-BatchTopK}

\vspace{0.35em}

\texttt{
Yesterday I found an old photograph of my friend's daughter and
I felt a little bit }
\steerhl{sad}
\texttt{.
I knew that my daughter was }
\steerhl{sad}
\texttt{ that her dad was taking away her and she was very upset.
I felt that my heart was very }
\steerhl{sad}
\texttt{.
I wanted to cry and I}
\texttt{\ldots}

\end{tcolorbox}

\caption{
\textbf{Qualitative steering example for the ``sadness'' concept.}
Given the same initial prompt, the unsteered generation remains largely neutral, while steering with BatchTopK SAE does not reliably induce the target concept and instead shifts the continuation toward unrelated content. In contrast, steering with Dual-BatchTopK SAE produces a continuation containing multiple semantically consistent \dualhlcap{expressions of sadness}, including \textit{sad}, \textit{upset}, and \textit{cry}, demonstrating stronger concept-specific control.
}
\label{fig:text_steering_sadness}

\end{figure*}

\begin{figure*}[t]
\centering
\small

\textbf{\large Steering toward ``Fear''}
\hspace{0.3em}

\vspace{0.6em}

\begin{tcolorbox}[
    colback=gray!3,
    colframe=gray!25,
    boxrule=0.6pt,
    arc=2mm,
    left=3mm,
    right=3mm,
    top=2mm,
    bottom=2mm
]
\textbf{Prompt.}
\texttt{The book on the table was about}
\end{tcolorbox}

\vspace{0.4em}

\begin{tcolorbox}[
    colback=white,
    colframe=gray!25,
    boxrule=0.6pt,
    arc=2mm,
    left=3mm,
    right=3mm,
    top=2mm,
    bottom=2mm
]

\textbf{\large Unsteered}

\vspace{0.35em}

\texttt{
The book on the table was about to be torn apart by a man who,
after an initial discussion, then proceeded to take his own life.
The man's suicide was described as an act of self-destruction.
}

\vspace{0.35em}

\texttt{The book was written in the 1930s}

\end{tcolorbox}

\vspace{0.4em}

\begin{tcolorbox}[
    colback=white,
    colframe=gray!25,
    boxrule=0.6pt,
    arc=2mm,
    left=3mm,
    right=3mm,
    top=2mm,
    bottom=2mm
]

\textbf{\large JumpReLU}

\vspace{0.35em}

\texttt{
The book on the table was about the city of Tokyo, and it was a good book,
but the book on the table was about the city of New York.
I'm not so sure I'm in the same place as the book on the table.
}

\end{tcolorbox}

\vspace{0.4em}

\begin{tcolorbox}[
    colback=white,
    colframe=gray!25,
    boxrule=0.6pt,
    arc=2mm,
    left=3mm,
    right=3mm,
    top=2mm,
    bottom=2mm
]

\textbf{\large Dual-JumpReLU}

\vspace{0.35em}

\texttt{
The book on the table was about a ``}
\steerhl{big bad}
\texttt{'' that would be in the future, and so he was }
\steerhl{afraid}
\texttt{ to use the word ``terrorist'' so much.
}

\vspace{0.35em}

\texttt{
But the last word is on the tip of my tongue.
I'm }
\steerhl{worried}
\texttt{ about}
\texttt{\ldots}

\end{tcolorbox}

\caption{
\textbf{Qualitative steering example for the ``fear'' concept.}
Given the same prompt, the unsteered continuation develops an unrelated narrative, while JumpReLU SAE steering does not clearly induce the target emotion and instead produces a largely neutral continuation. In contrast, steering with Dual-JumpReLU SAE introduces multiple \dualhlcap{fear-related expressions}, including \textit{afraid} and \textit{worried}, producing a continuation more consistently aligned with the intended concept.
}
\label{fig:text_steering_fear}

\end{figure*}

\begin{figure*}[t]
\centering
\small

\textbf{\large Steering toward ``Sports''}
\hspace{0.3em}

\vspace{0.6em}

\begin{tcolorbox}[
    colback=gray!3,
    colframe=gray!25,
    boxrule=0.6pt,
    arc=2mm,
    left=3mm,
    right=3mm,
    top=2mm,
    bottom=2mm
]
\textbf{Prompt.}
\texttt{He picked up the pen and wrote}
\end{tcolorbox}

\vspace{0.4em}

\begin{tcolorbox}[
    colback=white,
    colframe=gray!25,
    boxrule=0.6pt,
    arc=2mm,
    left=3mm,
    right=3mm,
    top=2mm,
    bottom=2mm
]

\textbf{\large Unsteered}

\vspace{0.35em}

\texttt{
He picked up the pen and wrote ``I am a writer.''
}

\end{tcolorbox}

\vspace{0.4em}

\begin{tcolorbox}[
    colback=white,
    colframe=gray!25,
    boxrule=0.6pt,
    arc=2mm,
    left=3mm,
    right=3mm,
    top=2mm,
    bottom=2mm
]

\textbf{\large GatedSAE}
\textcolor{gray!70}{[0, 0.02]}

\vspace{0.35em}

\texttt{
He picked up the pen and wrote,
``A.B.''
}

\end{tcolorbox}

\vspace{0.4em}

\begin{tcolorbox}[
    colback=white,
    colframe=gray!25,
    boxrule=0.6pt,
    arc=2mm,
    left=3mm,
    right=3mm,
    top=2mm,
    bottom=2mm
]

\textbf{\large Dual-GatedSAE}
\textcolor{gray!70}{[2, 1.00]}

\vspace{0.35em}

\texttt{
He picked up the pen and wrote a few lines.
}

\vspace{0.35em}

\texttt{
The next day, the }
\steerhl{Yankees}
\texttt{ won the }
\steerhl{series}
\texttt{, and he hit .269 with four }
\steerhl{homers}
\texttt{ and 13 }
\steerhl{RBIs}
\texttt{ in a loss in the Yankees' home opener.
He was recalled to the }
\steerhl{Red Sox}
\texttt{ in the}
\texttt{\ldots}

\end{tcolorbox}

\caption{
\textbf{Qualitative steering example for the ``sports'' concept.}
Given the same prompt, the unsteered and GatedSAE continuations do not introduce sports-related content. In contrast, Dual-GatedSAE steering yields a fluent continuation containing multiple \dualhlcap{sport-related concepts}, including the \textit{Yankees}, \textit{series}, \textit{homers}, \textit{RBIs}, and \textit{Red Sox}, demonstrating more effective concept-specific steering while preserving generation quality.
}
\label{fig:text_steering_sports}

\end{figure*}

\begin{figure*}[t]
\centering
\small

\textbf{\large Steering toward ``Anger''}
\hspace{0.3em}

\vspace{0.6em}

\begin{tcolorbox}[
    colback=gray!3,
    colframe=gray!25,
    boxrule=0.6pt,
    arc=2mm,
    left=3mm,
    right=3mm,
    top=2mm,
    bottom=2mm
]
\textbf{Prompt.}
\texttt{The bakery is next to}
\end{tcolorbox}

\vspace{0.4em}

\begin{tcolorbox}[
    colback=white,
    colframe=gray!25,
    boxrule=0.6pt,
    arc=2mm,
    left=3mm,
    right=3mm,
    top=2mm,
    bottom=2mm
]

\textbf{\large Unsteered}

\vspace{0.35em}

\texttt{
The bakery is next to the school, and I have a feeling it will be nice.
The staff is also very nice and helpful.
}

\vspace{0.35em}

\texttt{
My husband and I went to the bakery for breakfast and lunch.
It was so nice. The staff was very friendly and
}

\end{tcolorbox}

\vspace{0.4em}

\begin{tcolorbox}[
    colback=white,
    colframe=gray!25,
    boxrule=0.6pt,
    arc=2mm,
    left=3mm,
    right=3mm,
    top=2mm,
    bottom=2mm
]

\textbf{\large MatryoshkaSAE}

\vspace{0.35em}

\texttt{
The bakery is next to the mall and the mall is next to the bakery.
}

\end{tcolorbox}

\vspace{0.4em}

\begin{tcolorbox}[
    colback=white,
    colframe=gray!25,
    boxrule=0.6pt,
    arc=2mm,
    left=3mm,
    right=3mm,
    top=2mm,
    bottom=2mm
]

\textbf{\large Dual-MatryoshkaSAE}
\textcolor{gray!70}{[2, 0.94]}

\vspace{0.35em}

\texttt{
The bakery is next to the bus station.
}

\vspace{0.35em}

\texttt{I'm so }
\steerhl{sick}
\texttt{ of people in a rush to get out of a }
\steerhl{terrible}
\texttt{ place and I }
\steerhl{hate}
\texttt{ that I'm in a world where I can't even make it to the airport in time.}

\vspace{0.35em}

\texttt{The airport is}
\texttt{\ldots}

\end{tcolorbox}

\caption{
\textbf{Qualitative steering example for the ``anger'' concept.}
Given the same prompt, the unsteered generation remains positive and neutral, while MatryoshkaSAE steering produces a repetitive continuation without expressing the target emotion. In contrast, Dual-MatryoshkaSAE steering shifts the generation toward \dualhlcap{anger-related language}, including \textit{sick}, \textit{terrible}, and \textit{hate}, while maintaining a coherent continuation.
}
\label{fig:text_steering_anger}

\end{figure*}

\end{document}